\documentclass[lettersize,journal]{IEEEtran}
\usepackage{array}
\usepackage{textcomp}
\usepackage{stfloats}
\usepackage{url}
\usepackage{verbatim}
\usepackage[numbers]{natbib}
\usepackage{macros}

\begin{document}

\title{Informative Path Planning with Gaussian Processes:\\Error Bounds, Algorithms, and Benchmarks}

\author{Kalvik~Jakkala,
        Jennifer~Wakulicz,
        Donald~G.~Dansereau,
        Srinivas~Akella,
        Lantao~Liu,
        and~Jason~M.~O'Kane
\thanks{Corresponding author: Kalvik Jakkala.}
\thanks{K. Jakkala and J. M. O'Kane are with the Department of Computer Science and Engineering, Texas A\&M University, College Station, TX 77843 USA (e-mail: itskalvik@gmail.com).}%
\thanks{J. Wakulicz and D. G. Dansereau are with the Australian Centre for Robotics, School of Aerospace, Mechanical and Mechatronic Engineering, University of Sydney, NSW 2006, Australia.}%
\thanks{S. Akella is with the University of North Carolina at Charlotte, Charlotte, NC 28223 USA.}%
\thanks{L. Liu is with the Luddy School of Informatics, Computing, and Engineering, Indiana University, Bloomington, IN 47408 USA.}%
}

\maketitle

\begin{abstract}
Accurate environmental monitoring relies on Informative Path Planning (IPP) to deploy autonomous systems that maximize information gain. While Gaussian Processes (GPs) provide a robust mathematical foundation for modeling the environment, contemporary IPP approaches remain hindered by heuristic operational constraints and the lack of simultaneous theoretical guarantees for worst-case estimation error and approximation quality. Rather than introducing yet another heuristic, this paper presents a rigorous theoretical analysis of existing error bounds and algorithms to further develop the mathematical foundation for IPP.

By evaluating probabilistic and deterministic bounds on reconstruction error, we establish that predictive variance is the theoretically ideal optimization objective for GP-based IPP. Furthermore, we analyze GreedyCover, a recent approach for optimizing predictive variance, and demonstrate that its near-optimal approximation guarantee matches the fundamental hardness-of-approximation limit for the problem. Comprehensive benchmarking confirms that this approach delivers superior estimation accuracy and computational efficiency compared to prominent information-theoretic objectives and optimization methods used in IPP. Finally, this theoretical analysis establishes a rigorous, near-optimal foundation for discrete offline IPP under bounded uncertainty constraints, offering clear and principled guidance for future work in continuous-space and online formulations.
\end{abstract}


\section{Introduction}

\subsection*{Motivation: Environmental Monitoring}
In an era defined by rapid climate change, ecological degradation, and shifting global ecosystems, accurate environmental monitoring has become critical for scientific understanding and informed decision-making. Characterizing complex, large-scale natural phenomena---such as ocean salinity, atmospheric pollutants, soil moisture, or glacial bathymetry---requires high-resolution data that is often dangerous, time-consuming, or prohibitively expensive to collect manually. To address this, an overarching framework of Robotic Information Gathering (RIG) has emerged, focusing on deploying autonomous systems to maximize information gain regarding these environmental data fields~\cite{ChenLK24, JakkalaOA26}. At its core, RIG shares fundamental similarities with active learning; both domains seek to select the most informative data points to query in order to maximize model accuracy with minimal data collection effort.

\begin{figure}[!ht]
\centering
\includegraphics[width=\linewidth]{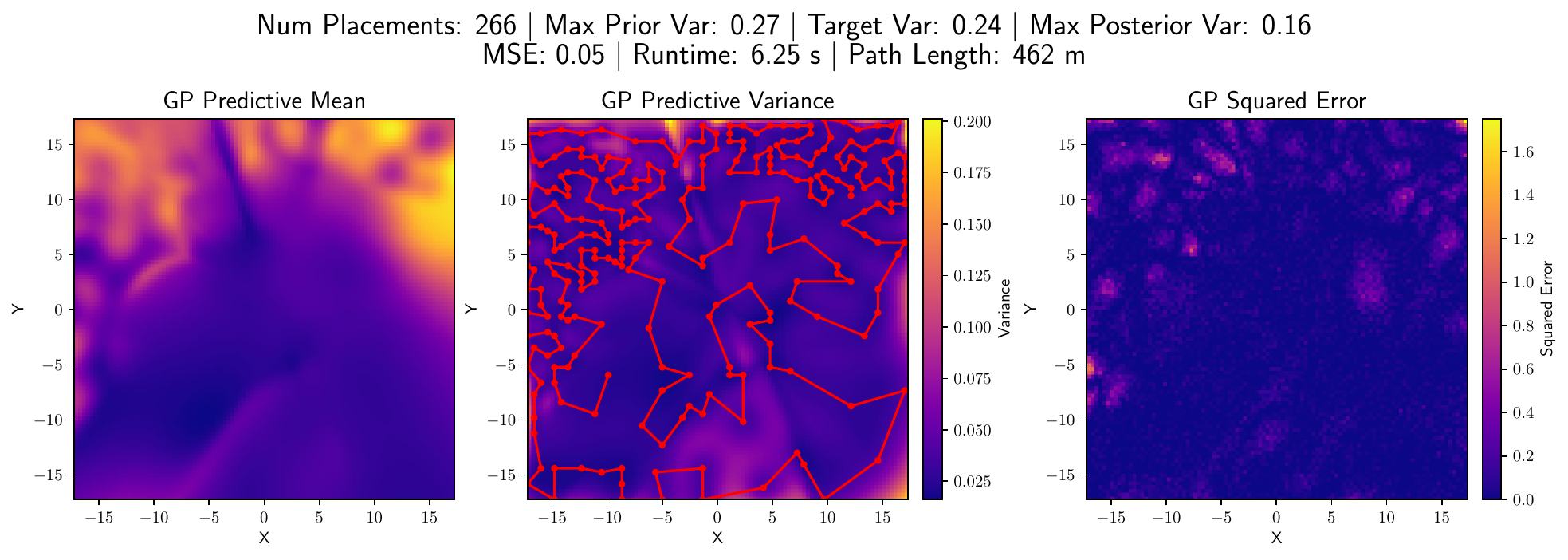}
\caption{Informative path generated by \textsc{GreedyCover} on the SRTM elevation dataset ($45^\circ\mathrm{N},\,123^\circ\mathrm{W}$). By leveraging a non-stationary kernel, the method adapts sampling density to spatial variations while guaranteeing bounded maximum posterior variance. Red: informative path and sensing locations.}
\label{fig:cover}
\vspace{-4mm}
\end{figure}

\subsection*{Shortcomings of Geometric Coverage}
Traditionally, environmental monitoring tasks have relied on geometric planners, such as boustrophedon (lawnmower) sweeps~\cite{ChosetP98} or Voronoi-based coverage methods~\cite{BreitenmoserSMSR10}. While these strategies successfully guarantee spatial visitation, they are inherently inefficient. By entirely ignoring the probabilistic data correlations within the environment, geometric sweeps frequently oversample highly predictable and redundant regions, wasting valuable time and energy.

\subsection*{Sensor Placement and Informative Path Planning}
Overcoming the limitations of geometric coverage requires addressing two interconnected problems: Sensor Placement (SP) and Informative Path Planning (IPP). While SP focuses on identifying optimal locations for static sensors, IPP routes mobile robots through informative regions while respecting operational constraints, such as motion constraints and limited onboard compute. To model data correlations, the vast majority of these approaches rely on Gaussian Processes (GPs)~\cite{RasmussenW05} because of their high sample efficiency and principled uncertainty quantification.

Within this GP framework, IPP encompasses two primary operational paradigms:
\begin{itemize}
    \item \textbf{Offline IPP:} The GP hyperparameters describing the environment are assumed to be known \textit{a priori}. This serves as a critical theoretical foundation to study the core problem structure.
    \item \textbf{Online IPP:} The GP hyperparameters are unknown beforehand and must be learned dynamically by the robot during the mission.
\end{itemize}

\subsection*{Research Gap: Heuristics, Compute Costs, and Guarantees}
Despite the mathematical foundation provided by GPs, a critical gap remains regarding the strength of theoretical guarantees offered by IPP algorithms. A persistent trade-off exists between discrete and continuous planning methods. While optimizing submodular objectives like mutual information enables discrete greedy planners to achieve strict approximation factor guarantees, these approaches suffer from severe computational bottlenecks due to expensive matrix operations. 

Conversely, planners operating in continuous spaces---such as Rapidly-exploring Random Trees (RRTs)~\cite{HollingerS14} or Ergodic Control~\cite{MillerSMM16}---can offer asymptotic convergence, but they inherently lack finite-time approximation guarantees. Indeed most IPP approaches on continuous domains provide anytime algorithms that provide better solutions the longer they run.

Furthermore, a fundamental limitation persists across both paradigms: the reliance on heuristic operational constraints. Most contemporary approaches require explicitly setting a predefined cardinality limit on sensing locations or an ad hoc travel distance budget. Ideally, a planner would instead allow users to specify a maximum allowable uncertainty or error, autonomously determining the necessary sensing budget and travel distance required to satisfy that bound. 

Crucially, because existing continuous methods—including ergodic exploration and sampling-based planners—typically plan over predetermined operational budgets (such as fixed time horizons or path lengths) and optimize spatial proxy distributions, they lack finite-time approximation guarantees on the resulting GP predictive variance and worst-case field estimation error.

\subsection*{Key Contributions}

To bridge the gap between heuristic planning and verifiable mapping quality, this work establishes a rigorous theoretical foundation for Informative Path Planning. Our core contributions are as follows:
\begin{itemize}
    \item \textbf{Squared Error Bounds and Objective Formulation:} 
    By leveraging existing probabilistic and deterministic error bounds, we contextualize the role of GP predictive variance in bounding true field reconstruction error—showing that it both lower-bounds expected mean squared error and, scaled by the RKHS norm, upper-bounds deterministic pointwise squared error. This formally establishes predictive variance as the mathematically ideal objective for offline GP-based IPP and provides a grounded blueprint for online formulations.
    
    \item \textbf{Theoretically Grounded Discrete Offline IPP:} 
    We analyze a recent optimization algorithm, GreedyCover~\cite{JakkalaAOA26}, for predictive variance and establish that its near-optimal approximation guarantee matches the fundamental hardness-of-approximation limit for the problem. Consequently, this establishes a theoretically grounded, complexity-optimal foundation for discrete offline GP-based IPP under bounded uncertainty constraints.

    \item \textbf{Comprehensive Benchmarking and Analysis:} We evaluate prominent information-theoretic metrics and optimization methods, empirically demonstrating that GreedyCover delivers strong error reduction with substantial computational efficiency. These results establish the method as a robust solution for offline GP-based IPP under bounded uncertainty constraints while highlighting critical implications for tackling continuous-space and online formulations.
\end{itemize}

\section{Related Work}

\subsection{Foundational Problems: RIG, SP, and IPP}
The overarching framework of Robotic Information Gathering (RIG) focuses on maximizing information gain regarding an unknown environmental field \cite{krauseSG08, RasmussenW05}. Within this domain, two foundational and heavily overlapping challenges are the Sensor Placement (SP) problem and the Informative Path Planning (IPP) problem. 

The SP problem seeks to identify an optimal set of fixed locations for sensors to maximize the informativeness of the collected data, with applications ranging from soil sampling to glacial monitoring \cite{ShewryW87, WuZ92, krauseSG08}. Building directly upon SP concepts, IPP addresses the routing aspect, aiming to plan optimal sensing paths for mobile robots while adhering to operational constraints such as travel budgets and vehicle kinematics \cite{HitzGGPS17, MaLHS18, OttKB24}. Both problems share analogous formulations with broader fields, such as active learning for machine learning~\cite{RenXCHLGCW21} and next-best-view planning in robotics~\cite{ZengWZL20}.

To effectively model the data correlations of the target environment, the vast majority of modern RIG approaches rely on Gaussian Processes (GPs) \cite{RasmussenW05}. GPs are ubiquitous in this domain due to their exceptional data efficiency and superior capacity to provide mathematically rigorous uncertainty estimation for unobserved locations \cite{krauseSG08, RasmussenW05}.

\subsection{Information-Theoretic Metrics}
A common strategy for solving SP and IPP involves selecting sampling locations that maximize specific information-theoretic objectives \cite{krauseSG08, CaseltonZ84}. These metrics define the fundamental trade-off between the quality of the environmental reconstruction and the computational cost required for planning.

\begin{itemize}
    \item \textbf{Variance and Entropy:} Selecting locations with the highest variance, a metric closely related to entropy, is an intuitive and computationally fast approach~\cite{krauseSG08}. However, variance-based metrics often yield less informative, suboptimal sensing locations, frequently clustering sensors along the boundaries of the search space instead of providing comprehensive coverage \cite{krauseSG08}.
    
    \item \textbf{Optimal Experimental Design Metrics:} Classical criteria from the field of optimal experimental design---such as B-, D-, and E-optimality \cite{OttKB24}---have been heavily utilized, but they suffer from similar limitations regarding sensing locations.
    
    \item \textbf{Predictive Variance:} Despite the limitations of naive variance minimization, mapping with respect to an explicit threshold for worst-case predictive variance is a robust, computationally efficient approach for bounding estimation uncertainty \cite{DuttaWTS23, SuryanT20}.
    
    \item \textbf{Mutual Information (MI):} Caselton and Zidek \cite{CaseltonZ84} proposed the application of MI for quantifying the reduction in uncertainty across all unobserved locations, naturally encouraging highly informative placements \cite{krauseSG08, LinCWT19, JakkalaOA26}. MI consistently outperforms variance-based and optimal experimental design criteria in terms of informativeness. Nevertheless, this superior performance incurs a severe computational cost; for an environment discretized into $n$ locations, computing MI requires $\mathcal{O}(n^3)$ operations, rendering it a massive bottleneck for real-time IPP applications where the objective must be evaluated repeatedly.
\end{itemize}

\subsection{Optimization Methods}
Given an information-theoretic metric, various optimization and planning algorithms have been proposed to compute paths that optimize the selected objective. 

\begin{itemize}
    \item \textbf{Graph-Based, Convex, and Greedy Approaches:} One foundational approach discretizes the environment and formulates the SP or IPP task as an integer program or a convex optimization problem \cite{JoshiB09, DuttaWS22, OttKB24}. While mathematically elegant, this confines potential sampling locations to a finite, predefined subset and scales poorly with high-resolution discretizations. Leveraging the submodularity of metrics like MI, discrete-space greedy algorithms iteratively select the most informative candidate locations, offering high efficiency but remaining tied to discrete grids \cite{ShewryW87, WuZ92, krauseSG08, BinneyKS10, KrauseG11, BinneyKS13}.
    
    \item \textbf{Continuous-Space and Sampling-Based Planners:} To overcome the restrictions of discrete methods, sampling-based planners, such as Rapidly-exploring Random Trees (RRTs), have been employed to maximize MI in continuous domains \cite{HollingerS14}. Alternatively, derivative-free black-box optimizers---including genetic algorithms \cite{HitzGGPS17, PopovicVHCSSN20} and Bayesian optimization \cite{FrancisOMR19, LinCWT19}---have been used. However, these methods scale poorly with dimensionality and demand numerous objective function evaluations, quickly becoming intractable when the objective is computationally expensive.
            
    \item \textbf{Gradient-Based and Ergodic Control Methods:} While prone to getting stuck in local minima, gradient-based approaches have emerged to enable fast, continuous-space optimization by utilizing differentiability of MI or by leveraging Sparse Gaussian Processes (SGPs) \cite{AsgharivaskasiKA22, JakkalaA24, JakkalaA25, JakkalaA25b}. For settings with known utility functions, continuous-space ergodic control algorithms are leveraged to efficiently guide exploration over the data field \cite{MillerSMM16, RaoBCJWC23, RaoCW25}.
    
    \item \textbf{POMDPs and Learning-Based Approaches:} To address the adaptive nature of IPP, some methods frame the problem as a Partially Observable Markov Decision Process (POMDP), utilizing Monte Carlo tree search \cite{OttBK23, OttKB24}. Furthermore, Deep Reinforcement Learning (DRL) and Imitation Learning (IL) are increasingly used to train reactive planning policies \cite{ManjannaD17, RuckinJP23}, although generalizing to novel environments remains an open challenge \cite{RuckinJP23}. 
    
    \item \textbf{Coverage Planners:} Baseline geometric planners---such as boustrophedon (lawnmower) sweeps \cite{ChosetP98, KarapetyanMLLOR18, AgarwalA22}, Voronoi-based coverage methods \cite{BreitenmoserSMSR10}, and art-gallery algorithms \cite{BanosL01}---ensure spatial visitation but typically oversample predictable regions by entirely ignoring probabilistic data correlations \cite{SchwagerVPRT17, BoumanOKCKLAB22}.
\end{itemize}


\subsection{Theoretical Guarantees}
A critical axis of comparison among IPP algorithms is their theoretical guarantees concerning both optimization performance and error bounds \cite{KrauseG11, DuttaWTS23, JakkalaAOA26}.

Although the SP problem is NP-complete~\cite{krauseSG08}, the submodular nature of mutual information (MI) allows discrete greedy algorithms to achieve a $(1 - 1/e)$-approximation guarantee~\cite{krauseSG08}. However, these methods require repeated MI evaluations involving costly matrix operations, creating a persistent trade-off between theoretical optimality and computational tractability.

Conversely, continuous-domain planners, such as RRTs, sacrifice finite-time approximation guarantees in exchange for continuous-space placements with asymptotic optimality guarantees \cite{HollingerS14, JadidiMD19}. Similarly, ergodic methods and SGP-based formulations offer rigorous asymptotic convergence guarantees for continuous space mapping \cite{MillerSMM16, JakkalaA25}.

Recent advances have specifically targeted the lack of formal estimation quality guarantees \cite{SuryanT20, DuttaWTS23}. Worst-case upper bounds on reconstruction error have been developed within Reproducing Kernel Hilbert Space frameworks for the noise-free~\cite{KarvonenST21} and bounded observation noise~\cite{Wakulicz2022} cases. While Wakulicz et al.~\cite{Wakulicz2022} leverage these bounds to guide informative path planning, their planner provides no near-optimal approximation guarantees on the resulting sensing set. Meanwhile, Jakkala et al.~\cite{JakkalaAOA26} leveraged the property that predictive variance lower bounds the GP Mean Squared Error (MSE) to establish near-optimal approximation guarantees for predictive variance optimization under bounded uncertainty.

\textbf{Research Gap:} While the GP MSE lower bound helps establish the required sensing budget under bounded uncertainty constraints, three key limitations remain unaddressed: (i) existing worst-case upper bounds have not been studied within this formulation to provide near-optimal approximation guarantees on predictive variance; (ii) hardness-of-approximation results for the GP predictive variance are lacking, leaving the theoretical tightness of the approximation guarantees unproven; and (iii) previous empirical evaluations have been restricted to just two baselines, obscuring performance relative to prominent existing methods. This work addresses all three limitations.
\section{Preliminaries: Gaussian Processes}
\label{sec:prelim_gp}
We model the monitored field as an unknown function $f: \mathcal{X} \to \mathbb{R}$ over a domain $\mathcal{X} \subset \mathbb{R}^d$ and use Gaussian processes~(GPs)~\cite{RasmussenW05} to capture correlations and uncertainty.

\textbf{GP Prior:}
A Gaussian process is a stochastic process such that any finite collection of function values is jointly Gaussian. Equivalently, a GP induces a distribution over functions, making it an expressive Bayesian prior for regression.
We place a GP prior on the latent function $f$:

$$f(x) \sim \mathcal{GP}\big(m(x), k(x,x')\big),$$

\noindent where $m(x)$ is the mean function (often taken to be $0$ after normalization) and kernel $k(x,x')$ is a covariance function. The kernel specifies prior correlations:

$$k(x,x') = \mathrm{cov}\!\left(f(x), f(x')\right).$$

\textbf{Noisy Observations and Posterior Prediction:}
Given $n$ noisy samples $\mathcal{D}=\{(x_i,y_i)\}_{i=1}^n$ with
$y_i = f(x_i) + \varepsilon_i$ and i.i.d.\ Gaussian noise $\varepsilon_i \sim \mathcal{N}(0,\sigma_n^2)$, define the training inputs $X=[x_1,\dots,x_n]$ and outputs $y=[y_1,\dots,y_n]^\top$. For a query location $x$, GP regression yields a Gaussian predictive distribution
$p(f(x) \vert{} X,y)=\mathcal{N}\big(\mu_{\mathrm{post}}(x \vert{} X),\sigma_{\mathrm{post}}^2(x \vert{} X)\big)$, with

$$\begin{aligned} 
\mu_{\mathrm{post}}(x \vert{} X) &= K_{x n}\big(K_{nn}+\sigma_n^2 I\big)^{-1} y, \\
\sigma_{\mathrm{post}}^2(x \vert{} X) &= k(x,x) - K_{x n}\big(K_{nn}+\sigma_n^2 I\big)^{-1}K_{n x}. 
\end{aligned}$$

Here, $K_{nn}$ is the $n\times n$ covariance matrix with $(K_{nn})_{ij}=k(x_i,x_j)$, $K_{x n}$ is the $1\times n$ vector of cross-covariances with $(K_{x n})_i = k(x,x_i)$, $K_{n x}=K_{x n}^\top$, and $k(x,x)$ is the prior variance at $x$.
The posterior (or predictive) variance, $\sigma_{\mathrm{post}}^2(x \mid X)$, quantifies intrinsic uncertainty based solely on input locations and Gaussian process hyperparameters, making it particularly well-suited for planning.

\textbf{Hyperparameter Optimization:}
The covariance function $k(x,x')$ and the noise model typically depend on a set of hyperparameters $\theta$ (e.g., lengthscales, signal variance, and the noise variance). To properly fit the GP to the observed data, these hyperparameters are optimized by maximizing the log marginal likelihood:

$$\begin{aligned}
\log p(y \vert{} X, \theta) = & -\frac{1}{2} y^\top \big(K_{nn} + \sigma_n^2 I\big)^{-1} y \\
& \quad - \frac{1}{2} \log \vert{}K_{nn} + \sigma_n^2 I\vert{} - \frac{n}{2} \log 2\pi.
\end{aligned}$$

\noindent This objective naturally balances data fit (the first term, which penalizes large residuals) against model complexity (the second term, which penalizes highly flexible covariance matrices). This allows the GP to adapt to the scale and smoothness of the target function without overfitting.
\section{Problem: Informative Path Planning Under Bounded Uncertainty Constraints}

\subsection{Informative Path Planning (IPP)}
The fundamental goal of Informative Path Planning (IPP) is to determine an optimal data collection strategy that maximizes the information gathered about an environmental data field (e.g., salinity, temperature, bathymetry) over a domain $\mathcal{X} \subset \mathbb{R}^d$. Let the true data field be denoted by $f: \mathcal{X} \to \mathbb{R}$.

\subsection{System Setup}
For planning and performance evaluation, we define a finite set of \emph{evaluation locations} $\mathcal{V} = \{x_1, \dots, x_N\} \subset \mathcal{X}$ where accurate estimates are required. A robot can only sample from a restricted set of \emph{candidate sensing locations} $\mathcal{C} = \{c_1, \dots, c_M\} \subseteq \mathcal{X}$. A feasible path is represented by waypoints $\mathcal{P} = (p_1, \dots, p_T)$, where each $p_t \in \mathcal{C}$. In typical IPP scenarios, $T \ll M \le N$. Furthermore, we assume that neither $\mathcal{V}$ nor $\mathcal{C}$ is labeled, and that both are discretized over the continuous domain (e.g., using uniform grids).

\subsection{Gaussian Process (GP) Environmental Modeling}
Directly minimizing the data field prediction error is infeasible because the ground truth $f(x)$ is unknown, and the realized error inherently depends on the specific measurement values observed along a path. To obtain a tractable surrogate objective, we model the unknown field $f$ as a Gaussian Process (GP). Under a GP, the prediction at any point is given by the posterior mean $\mu_{\mathrm{post}}(x \vert{} \mathcal{P})$.

\subsection{Problem Statement}
To prevent regions of unacceptably high uncertainty, we require the posterior variance, $\sigma^2_{\mathrm{post}}(x \vert{} \mathcal{P})$, to remain strictly bounded across the discretized environment. We formulate the problem as finding the minimum-cost path $\mathcal{P}^\star$ such that the posterior variance at every evaluation point $x_i \in \mathcal{V}$ satisfies a specified target threshold $\sigma^2_{\mathrm{tar}}$:

\begin{equation}
\label{eq:problem_statement}
\begin{aligned}
    \mathcal{P}^\star \in \operatorname*{argmin}_{\mathcal{P} \in \mathcal{C}^*} \quad & \mathrm{cost}(\mathcal{P}) \\
    \text{s.t.} \quad & \sigma^2_{\mathrm{post}}(x_i \vert{} \mathcal{P}) \le \sigma^2_{\mathrm{tar}}, \quad \forall x_i \in \mathcal{V}\,,
\end{aligned}
\end{equation}
\noindent where $\mathrm{cost}(\mathcal{P})$ denotes a path execution cost (e.g., total path length or cardinality $\vert{}\mathcal{P}\vert{}$), and $\mathcal{C}^*$ represents the set of all finite waypoint sequences formed from $\mathcal{C}$.
\section{Optimization Objective: Predictive Variance}

To quantify the performance of our data field estimation, we must first ground our objective in the theoretical bounds of Gaussian Process (GP) regression.

\subsection{Probabilistic MSE Lower Bound}

\medskip

\begin{restatable}{theorem}{lowerbound}
\label{thm:mse_lower_bound}
MSE Lower Bound ~\cite{WagbergZSS17}

When the true hyperparameters are known, the mean squared error (MSE) at any point $x \in \mathcal{X}$ from the domain is lower bounded:

$$\begin{aligned} 
\text{MSE} &= \mathbb{E}_{f\sim\mathcal{GP}(\mu_{\mathrm{post}}, \sigma_{\mathrm{post}}^2)}\left[(f(x) - \mu_{\mathrm{post}}(x \vert{} \mathcal{P}))^2\right] \\ 
&\ge k(x, x) - K_{x n}\big(K_{nn}+\sigma_n^2 I\big)^{-1}K_{nx} \\ 
&= \sigma^2_{\mathrm{post}}(x \vert{} \mathcal{P}) 
\end{aligned}$$

\end{restatable}

This theorem establishes that the GP predictive variance, $\sigma^2_{\mathrm{post}}(x \vert{} \mathcal{P})$, inherently acts as a mathematical lower bound on the predictive MSE in the offline formulation where GP hyperparameters are known a priori.

\begin{figure}[!ht]
\centering
\includegraphics[width=\linewidth]{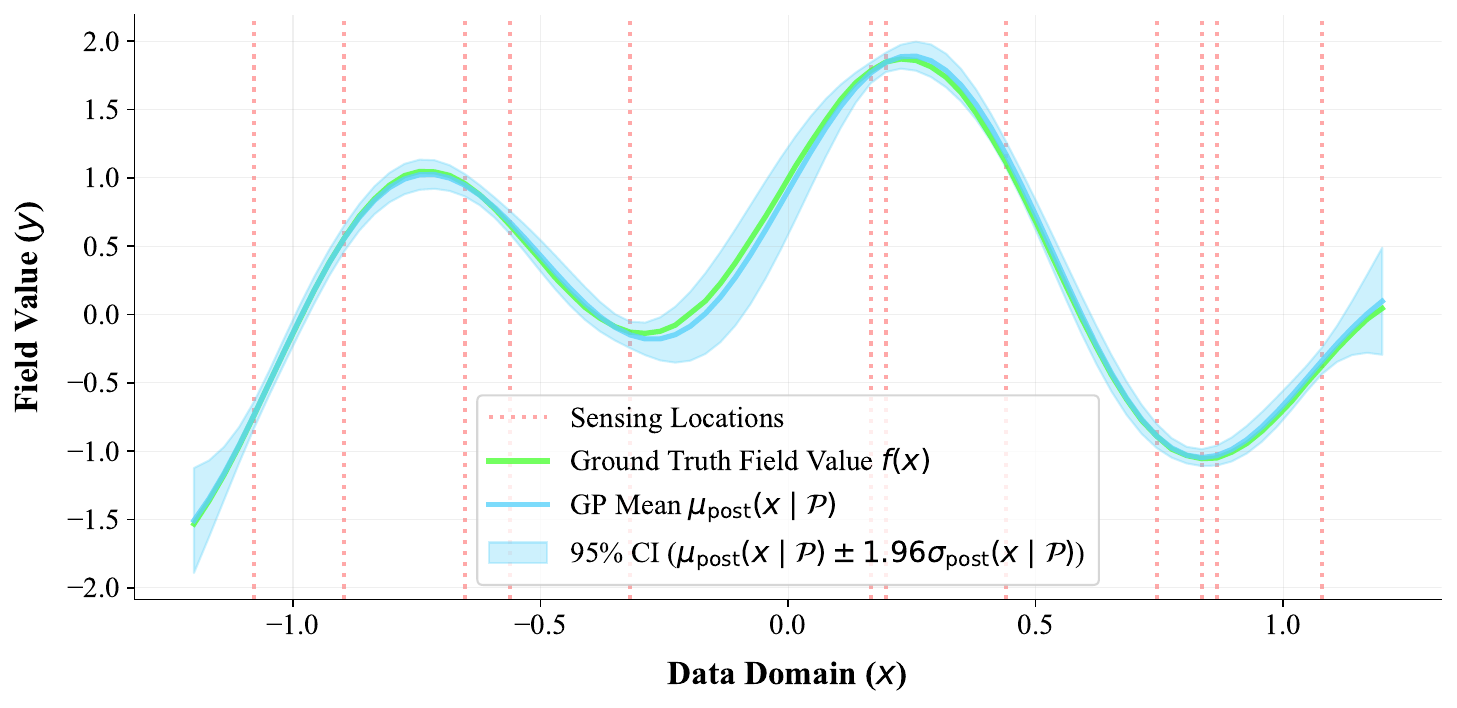}
\caption{GP regression on an illustrative non-stationary 1D data field. Observations (dashed vertical lines) are sampled near the outer intervals, leaving an unobserved gap in the middle. The estimated GP predictive mean $\mu_{\text{post}}(x \mid \mathcal{P})$ closely fits the ground-truth function $f(x)$, while the $95\%$ confidence interval ($\pm 1.96\sigma_{\text{post}}$) expands across the unsampled center to reflect increased predictive uncertainty.}
\label{fig:data_field_gp_approx}
\vspace{-3mm}
\end{figure}

\begin{figure}[!th]
\centering
\includegraphics[width=\linewidth]{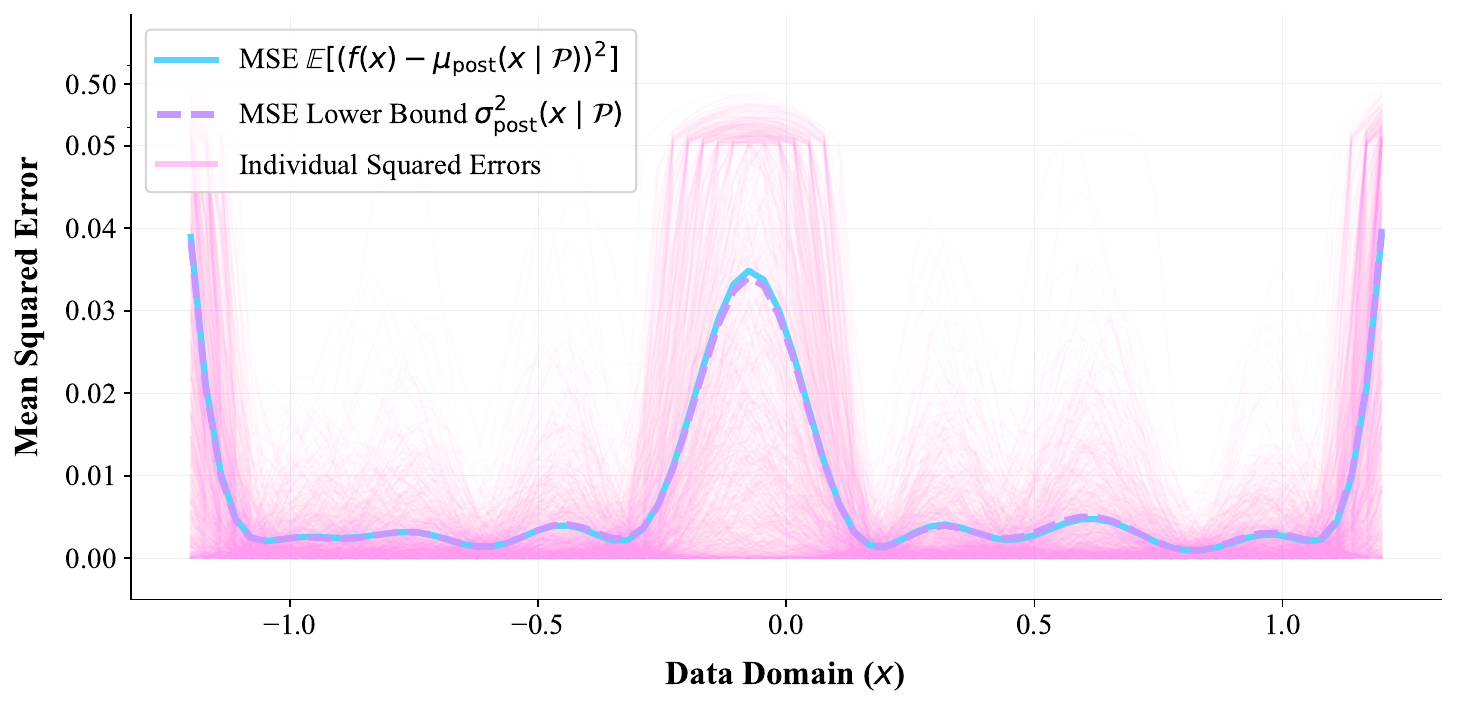}
\caption{Empirical validation of the GP MSE lower bound from Theorem~\ref{thm:mse_lower_bound}. The observed MSE and its theoretical lower bound are plotted alongside the pointwise squared errors from 1,000 realizations of the underlying stochastic process used to compute the MSE. $y$-axis shown on a logarithmic scale with a linear region near zero. The central peak reflects the spatial distribution of sensing locations (Figure.~\ref{fig:data_field_gp_approx}), which are placed symmetrically in the outer regions and leave an unobserved gap in the center.}
\label{fig:mse_lower_bound}
\vspace{-4mm}
\end{figure}

To validate the bound, we first train a GP with an Attentive kernel~\cite{ChenKL22} on 1D observations sampled uniformly from $\mathbb{F}(x) = \sin(6x) + \cos(2x)$ over $x \in [-1.5, 1.5]$, as shown in Figure~\ref{fig:data_field_gp_approx}. Using the posterior predictive distribution from this fitted model, Figure~\ref{fig:mse_lower_bound} empirically validates the theoretical MSE lower bound, demonstrating tight agreement between the predicted bound and empirical error statistics computed across repeated stochastic realizations.

In online applications where hyperparameters are unknown and must be estimated dynamically, the tightness and validity of this bound fluctuate:
\begin{itemize}
    \item When the estimated lengthscales are \textbf{shorter} than the true environmental lengthscales, the GP underestimates data correlation, causing the model to rapidly revert to the prior variance and rendering the resulting bound overly conservative and loose.
    \item If the estimated lengthscales are \textbf{larger} than reality, the model becomes overconfident, which artificially shrinks the predictive variance and distorts the bound's reliability.
\end{itemize}

While predictive variance serves as an ideal optimization objective for MSE under known hyperparameters, it acts only as a lower bound, leaving the true MSE uncharacterized. Furthermore, since MSE evaluates the expected error, bounding it does not prevent large errors in individual stochastic realizations of the function of interest. To establish strict, worst-case performance guarantees, we therefore transition to a deterministic framework.

\subsection{Deterministic Worst-Case Error Upper Bounds}

To overcome the limitations of probabilistic bounds, we analyze existing bounds from interpolation theory within Reproducing Kernel Hilbert Spaces (RKHS), starting with the worst-case absolute error.

Theorem~\ref{thm:ae_upper_bound} establishes a relationship between the worst-case absolute error and the standard deviation of the predictive distribution, which holds even in the online case where the GP hyperparameters are unknown.

\begin{restatable}{theorem}{aeupperbound}
\label{thm:ae_upper_bound}

Absolute Error Upper Bound \cite{Wakulicz2022}

Let $f \in \mathcal{H}_k$ where $\mathcal{H}_k$ is a Reproducing Kernel Hilbert Space defined by an arbitrary positive definite kernel $k$. Given a set of $n$ noisy observations $y = [y_1, \dots, y_n]^\top$ collected at sensing locations $\mathcal{P} = (p_1, \dots, p_n)$, assume the measurement noise is strictly bounded such that $\varepsilon_i^2  < \varepsilon_\star^2$ for all $n$ observations. The absolute error of the Gaussian Process predictive mean, $\mu_{\mathrm{post}}(x \vert{} \mathcal{P})$, at any point $x \in \mathcal{X}$ satisfies:
\[
|f(x) - \mu_{\mathrm{post}}(x \vert{} \mathcal{P})| \le ||f||_{\mathcal{H}_k} P_{\mathcal{P}}(x) + \sqrt{n} \varepsilon_\star \Lambda_k(x)
\]
where $\|f\|_{\mathcal{H}_k}^2$ is the RKHS complexity term, $P_{\mathcal{P}}(x) = \sqrt{\sigma^2_{\mathrm{post}}(x \vert{} \mathcal{P})}$ is the power function corresponding to the standard deviation of the predictive distribution, and $\Lambda_k(x) = ||K_{nn}^{-1} K_{n*}||_2$ represents the norm of the kernel matrix projection vector at the evaluation point.
\end{restatable}

Because the underlying data field is modeled as a deterministic function residing in an RKHS—where the RKHS norm $\Vert{}f\Vert{}_{\mathcal{H}_k}$ characterizes the function's complexity—minimizing this deterministic upper bound guarantees that the realized absolute estimation error remains strictly upper bounded.

\begin{figure}[!ht]
\centering
\includegraphics[width=\linewidth]{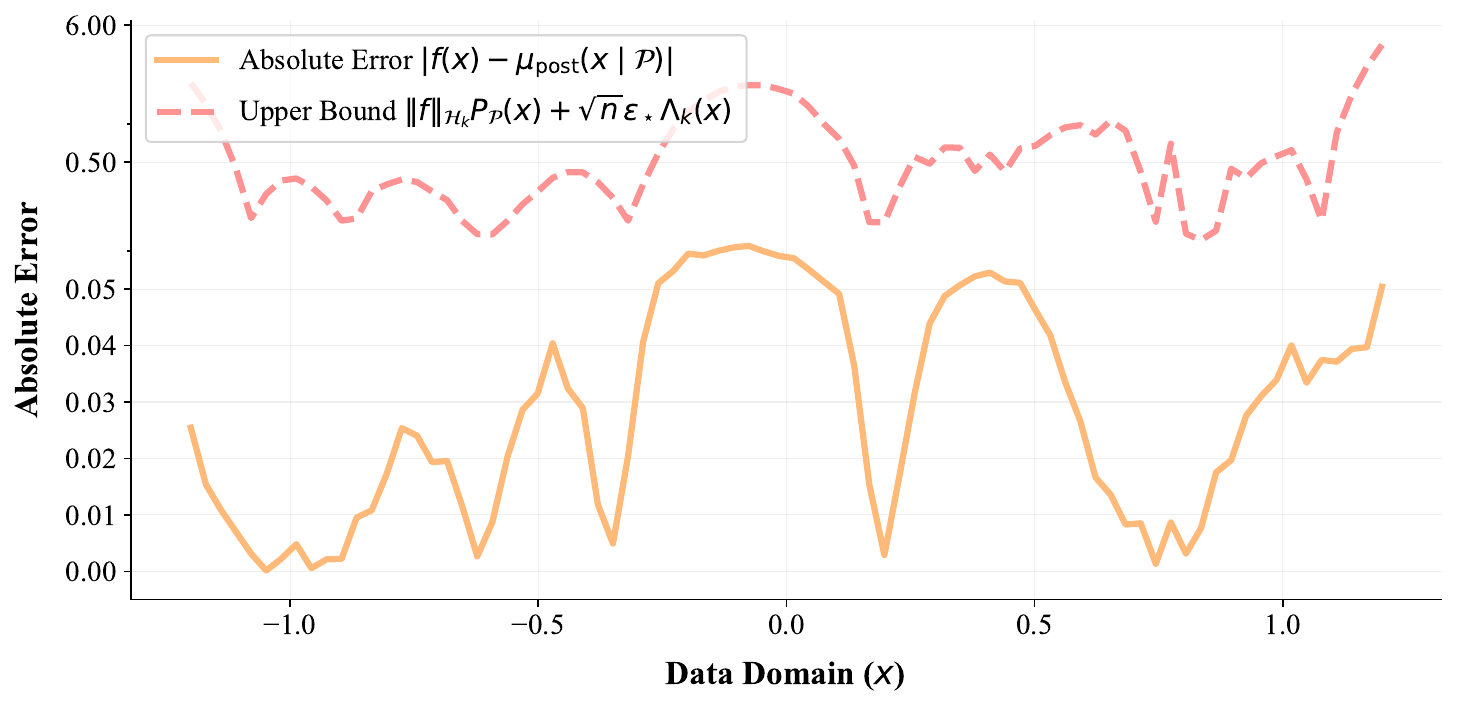}
\caption{Empirical validation of the deterministic absolute error upper bound from Theorem~\ref{thm:ae_upper_bound}. $y$-axis shown on a logarithmic scale with a linear region near zero.}
\label{fig:absolute_error_bound}
\vspace{-2mm}
\end{figure}

Using the GP model trained in Figure~\ref{fig:data_field_gp_approx}, Figure~\ref{fig:absolute_error_bound} empirically validates the theoretical bound (Theorem~\ref{thm:ae_upper_bound}), demonstrating that the upper bound strictly envelopes the observed absolute error.

Squaring the above bound (Theorem~\ref{thm:ae_upper_bound}) gives the direct connection between the GP predictive variance and pointwise squared error formalized in the following corollary.

\begin{restatable}{corollary}{seupperbound}
\label{thm:se_upper_bound}
Squared Error Upper Bound

Let $f \in \mathcal{H}_k$ be a target function within a Reproducing Kernel Hilbert Space defined by kernel $k$, and suppose we have $n$ noisy observations subject to bounded measurement noise $\varepsilon_i^2  < \varepsilon_\star^2$. The pointwise squared estimation error of the Gaussian Process predictive mean, $\mu_{\mathrm{post}}(x \vert{} \mathcal{P})$, at any point $x \in \mathcal{X}$ is strictly upper bounded by:

$$\begin{aligned} 
(f(x) - \mu_{\mathrm{post}}(x \vert{} \mathcal{P}))^2 
&\le 2 \Vert{}f\Vert{}_{\mathcal{H}_k}^2 \sigma^2_{\mathrm{post}}(x \vert{} \mathcal{P}) + 2n \varepsilon_\star^2 \Lambda_k^2(x) 
\end{aligned}$$

\proofnote
\end{restatable}

By isolating the predictive variance $\sigma^2_{\mathrm{post}}(x \vert{} \mathcal{P})$, Corollary~\ref{thm:se_upper_bound} demonstrates that minimizing predictive variance directly drives down the worst-case squared error. 

We adopt the current bound formulation to clearly isolate the contribution of predictive variance from the sensor noise. Although a tighter bound can be derived, it would introduce additional terms that couple these two sources of uncertainty, making their individual contributions less transparent.

\begin{figure}[!ht]
\centering
\includegraphics[width=\linewidth]{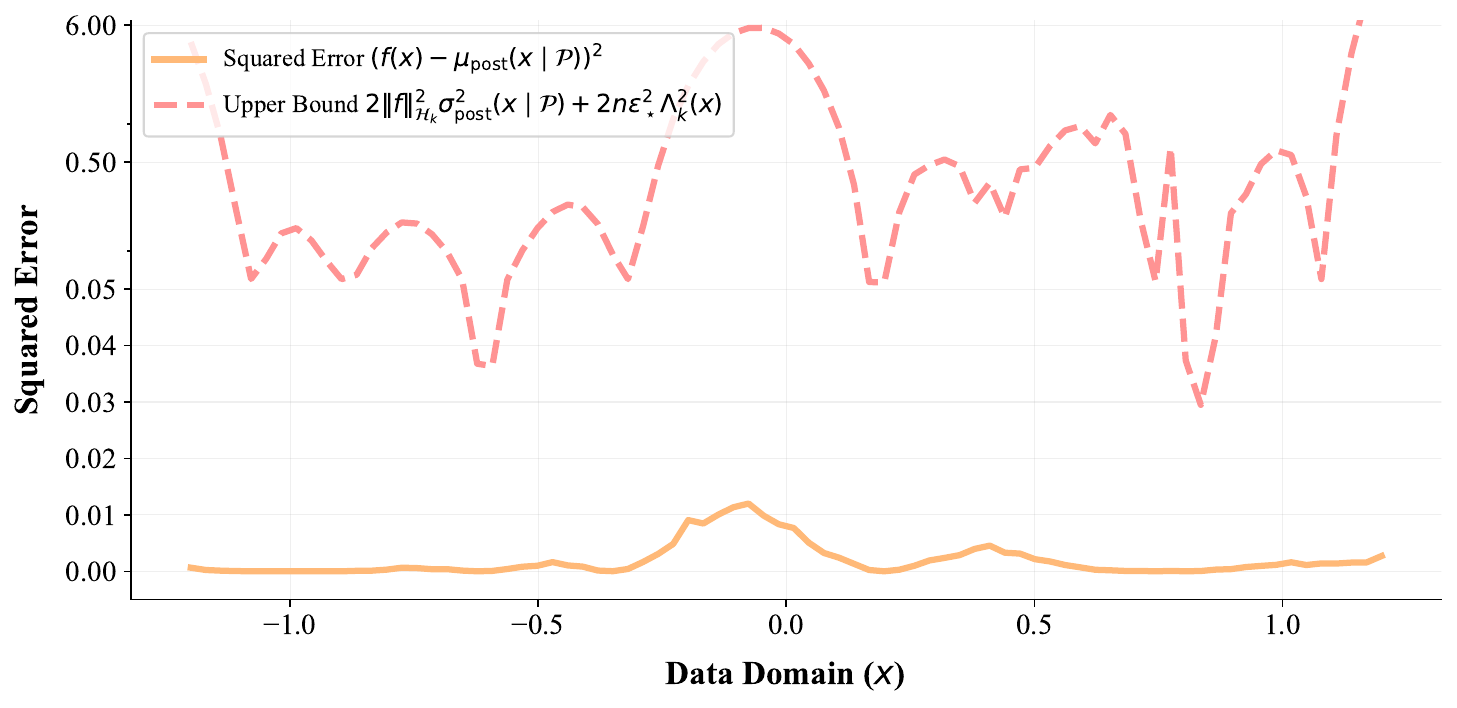}
\caption{Empirical validation of the deterministic squared error upper bound from Corollary~\ref{thm:se_upper_bound}. $y$-axis shown on a logarithmic scale with a linear region near zero.}
\label{fig:squared_error_bound}
\vspace{-4mm}
\end{figure}

\begin{figure}[!ht]
\centering
\includegraphics[width=\linewidth]{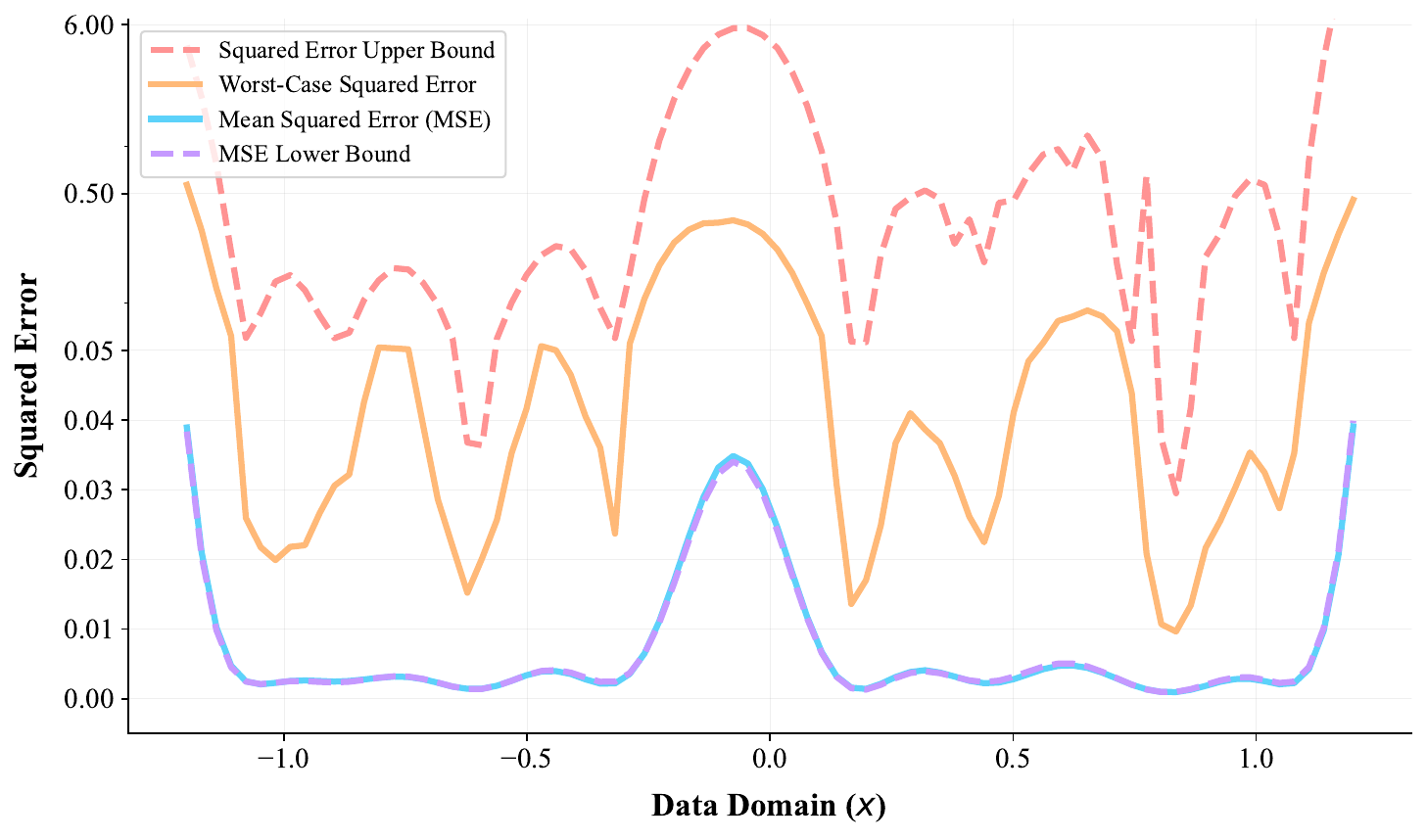}
\caption{GP error upper and lower bounds along with corresponding errors. $y$-axis shown on a logarithmic scale with a linear region near zero.}
\label{fig:combined_sq_error_bounds}
\vspace{-4mm}
\end{figure}

Using the GP model trained in Figure~\ref{fig:data_field_gp_approx}, the results in Figure~\ref{fig:squared_error_bound} empirically validate Corollary~\ref{thm:se_upper_bound} by demonstrating that the theoretical upper bound strictly envelopes the observed squared error across the domain. Finally, Figure~\ref{fig:combined_sq_error_bounds} summarizes the complete theoretical envelope, contrasting both the MSE lower bound and the squared error upper bound relative to the realized estimation errors.

In environments with negligible measurement noise, this relationship simplifies considerably, as formalized next. Theorem~\ref{thm:nf_se_upper_bound} establishes a clear relationship between the GP predictive variance $\sigma^2_{\mathrm{post}}(x \vert{} \mathcal{P})$ and the pointwise squared error. In this formulation, the RKHS norm $\Vert{}f\Vert{}_{\mathcal{H}_k}^2$ directly dictates the tightness of the upper bound.

\begin{restatable}{theorem}{nfseupperbound}
\label{thm:nf_se_upper_bound}
Noise-Free Squared Error Upper Bound~\cite{KarvonenST21}

Assume a noiseless sensing scenario where the measurement noise variance $\sigma_n^2 = 0$. The pointwise squared estimation error of the Gaussian Process predictive mean $\mu_{\mathrm{post}}(x \vert{} \mathcal{P})$ for a target function $f \in \mathcal{H}_k$ is strictly bounded by the product of the squared RKHS norm of $f$ and the GP predictive variance:
\begin{align*}
(f(x) - \mu_{\mathrm{post}}&(x \mid \mathcal{P}))^2 \le \|f\|_{\mathcal{H}_k}^2 \, \sigma^2_{\mathrm{post}}(x \mid \mathcal{P}), 
\end{align*}

\noindent where $\|f\|_{\mathcal{H}_k}^2$ is the RKHS complexity term and $\sigma^2_{\mathrm{post}}(x \mid \mathcal{P})$ is the predictive variance.

\end{restatable}

\subsection{Interpretation of the RKHS Norm}

To fully contextualize these error bounds, it is essential to understand the mathematical definition of the RKHS norm, $\|f\|_{\mathcal{H}_k}$, and its role within GP regression.

\textbf{Kernel Space and RKHS Connection:}
By the Moore--Aronszajn Theorem~\cite{RasmussenW05}, the positive-definite GP kernel $k(\cdot, \cdot)$ uniquely specifies a Reproducing Kernel Hilbert Space $\mathcal{H}_k$. The kernel function and its hyperparameters (e.g., lengthscale, signal variance) completely determine the inner product and geometry of $\mathcal{H}_k$, embedding structural assumptions regarding the continuity, differentiability, and allowable spatial variation of functions residing in the space.

\textbf{Spectral Perspective and Model Complexity:}
The norm is best understood through Mercer's Theorem~\cite{RasmussenW05}, which decomposes the kernel into an orthonormal basis of continuous eigenfunctions $\phi_j$ and associated non-negative eigenvalues $\lambda_j$:
\begin{equation}
k(x, x') = \sum_{j=1}^\infty \lambda_j \phi_j(x) \phi_j(x').
\end{equation}
Any function $f \in \mathcal{H}_k$ can be expanded in this eigenbasis as $f(x) = \sum_{j=1}^\infty w_j \phi_j(x)$, yielding the squared RKHS norm:
\begin{equation}
\|f\|_{\mathcal{H}_k}^2 = \sum_{j=1}^\infty \frac{w_j^2}{\lambda_j}.
\end{equation}
For standard stationary kernels (such as the Squared Exponential or Matérn families), the eigenvalues $\lambda_j$ decay rapidly toward zero as the index $j$ increases, with higher indices corresponding to higher-frequency spatial harmonics. Because $\lambda_j$ appears in the denominator, any spectral energy $w_j^2$ allocated to high frequencies inflates the norm. Consequently, $\Vert{}f\Vert{}_{\mathcal{H}_k}^2$ acts as a \emph{complexity metric} over the RKHS. 

\subsection{Implications for Informative Path Planning (IPP)}

Ultimately, the true RKHS norm is fully determined by the GP kernel and its hyperparameters. Since computing $\Vert{}f\Vert{}_{\mathcal{H}_k}^2$ directly is generally intractable, this relationship dictates how IPP optimization must be formulated:

\begin{itemize}
\item \textbf{Offline IPP:} In this paradigm, GP hyperparameters are assumed to be known \textit{a priori}. Consequently, the RKHS norm directly quantifies the intrinsic spatial roughness of the ground-truth data field—independent of any observation set. If the underlying field exhibits high curvature or sharp spatial gradients, the resulting large RKHS norm inflates the theoretical error bound regardless of the executed sensing path. Crucially, because the kernel structure and hyperparameters remain fixed, the RKHS norm acts as a fixed (albeit unknown) constant.

\begin{tcolorbox}[
    colback=NeonBlue!10,      
    colframe=NeonBlue!50,     
    arc=4pt,                  
    boxrule=1pt,              
    left=6pt, right=6pt, top=6pt, bottom=6pt,
    before skip=6pt, after skip=6pt        
]
Therefore, minimizing GP predictive variance to determine the solution sensing locations serves as the mathematically ideal objective function for offline GP-based IPP.
\end{tcolorbox}
\smallskip

\item \textbf{Online IPP:} When GP hyperparameters are estimated adaptively from streaming observations, the norm $\Vert{}f\Vert{}_{\mathcal{H}_k}^2$ measures complexity relative to the estimated RKHS rather than to the true field. Discrepancies between this estimated space and the true underlying field geometry can lead to model misspecification, potentially invalidating the theoretical error bounds.

\begin{tcolorbox}[
    colback=NeonBlue!10,      
    colframe=NeonBlue!50,     
    arc=4pt,                  
    boxrule=1pt,              
    left=6pt, right=6pt, top=6pt, bottom=6pt,
    before skip=6pt, after skip=2pt
]
Therefore, online IPP methods must balance data collection for predictive variance reduction with sampling to refine hyperparameter estimates, ensuring the model operates within the true RKHS and tightly bounds the squared estimation error.
\end{tcolorbox}

\end{itemize}
\section{Optimization Method: GreedyCover}

The preceding section established Gaussian process (GP) predictive variance as a theoretically ideal objective for Informative Path Planning (IPP). However, naively selecting sensing locations to minimize this metric not only yields suboptimal paths, but also fails to provide theoretical approximation guarantees. 

To address this challenge, we adopt the \textsc{GreedyCover} algorithm~\cite{JakkalaAOA26}, which presents a decoupled approach to IPP by separating sensor placement from path generation. First, it identifies a near-minimal set of sensing locations with provable approximation guarantees to strictly bound predictive variance below a user-specified threshold. Second, it routes the robot across the selected locations using an approximation algorithm for the Traveling Salesperson Problem (TSP)~\cite{Christofides22}. Unlike conventional IPP methods that impose heuristic travel distance or cardinality limits, \textsc{GreedyCover} directly certifies a formal bound on the maximum predictive variance.

While Sections~\ref{sec:theoretical_foundations}--\ref{sec:approximation_guarantees} review the \textsc{GreedyCover} algorithm~\cite{JakkalaAOA26}, our key contribution lies in analyzing its theoretical limits: under the standard assumption that $P \neq NP$, we prove that its approximation factor matches the fundamental hardness-of-approximation threshold, demonstrating that it attains the optimal polynomial-time bound (Section~\ref{sec:approximation_hardness}).

\subsection{Theoretical Foundations}
\label{sec:theoretical_foundations}

To achieve these approximation guarantees, the approach reframes predictive variance optimization as a Set Cover problem. It begins by establishing the mathematical conditions under which a single data field observation guarantees a target bound on posterior uncertainty. After extending these conditions to multiple observations, these theoretical foundations are used to generate a binary coverage matrix, effectively reframing the continuous optimization problem as a tractable submodular set cover problem that can be efficiently solved.

\begin{restatable}{theorem}{singlepoint}
\label{thm:single_point}
Minimum Required Prior Covariance \cite{JakkalaAOA26}

Let $f \sim \mathcal{GP}(0,k)$ with independent Gaussian observation noise variance $\sigma_n^2$, and consider a single noisy observation at candidate location $c \in \mathcal{C}$. 
Fix an evaluation location $x \in \mathcal{V}$. 
For any target posterior variance $\sigma^2_{\mathrm{tar}} \in (0,k(x,x))$, the condition
\[
\sigma^2_{\mathrm{post}}(x \mid c) \le \sigma^2_{\mathrm{tar}}
\]
holds \emph{if and only if}
\begin{equation}
\label{eq:cov_threshold_detailed}
|k(c,x)|
\;\ge\;
\sqrt{\bigl(k(x,x) - \sigma^2_{\mathrm{tar}}\bigr)\bigl(k(c,c) + \sigma_n^2\bigr)}.
\end{equation}
Thus, achieving a posterior variance of at most $\sigma^2_{\mathrm{tar}}$ at $x$ requires that the prior covariance between $c$ and $x$
exceed the threshold in~Equation~\ref{eq:cov_threshold_detailed}.
\end{restatable}

\begin{restatable}{theorem}{multipoint}
\label{thm:multi_point}
Variance Reduction with Multiple Locations \cite{JakkalaAOA26}

Under the setup of Theorem~\ref{thm:single_point}, consider $n$ candidate sensing locations $c_1,\dots,c_n \in \mathcal{C}$.
Fix an evaluation location $x \in \mathcal{V}$ and a target variance $\sigma^2_{\mathrm{tar}} \in (0,k(x,x))$. Suppose there exists
$j \in \{1,\dots,n\}$ such that
\[
|k(c_j,x)|
\;\ge\;
\sqrt{\bigl(k(x,x) - \sigma^2_{\mathrm{tar}}\bigr)\bigl(k(c_j,c_j) + \sigma_n^2\bigr)}.
\]
Then the posterior variance at $x$ conditioned on \emph{all} observations satisfies
\[
\sigma^2_{\mathrm{post}}\!\left(x \mid c_{1:n}\right) \le \sigma^2_{\mathrm{tar}}.
\]
\end{restatable}

\subsection{Greedy Optimization Algorithm}
\label{sec:optimization_algorithm}

To efficiently optimize the GP predictive variance, the method first constructs a spatial binary coverage map. Following Theorems~\ref{thm:single_point} and~\ref{thm:multi_point}, a candidate sensing location $c_j$ effectively covers an evaluation point $x_i$ if the prior covariance $k(c_j,x_i)$ strictly exceeds the minimum threshold in Equation~\ref{eq:cov_threshold_detailed}. This coverage relationship is formalized with a binary matrix $B \in \{0,1\}^{M\times N}$. For every candidate sensing location $c_j \in \mathcal{C}$ and evaluation point $x_i \in \mathcal{V}$, the corresponding matrix entry is defined as:
\[
B_{ji} = \mathbb{I}\!\left\{
|k(c_j,x_i)| \;\ge\;
\sqrt{\left(k(x_i,x_i)-\sigma^2_{\mathrm{tar}}\right)\;
\left(k(c_j,c_j)+\sigma_n^2\right)}
\right\}
\]

Here, $\mathbb{I}\{\cdot\}$ is an indicator function returning $1$ if the condition holds and $0$ otherwise. This binary representation offers significant practical advantages by enabling fast updates during planning via simple set operations. This also circumvents the computationally expensive process of repeatedly evaluating GP posterior variances across candidate subsets. Leveraging this binary matrix, the GreedyCover algorithm (Algorithm~\ref{alg:greedy-coverage}) iteratively selects sensing locations that maximize the marginal gain of newly covered points, efficiently optimizing the predictive variance. The algorithm operates with a worst-case time complexity of $\mathcal{O}(M^2N)$, when $M \le N$. Once optimized, the resulting set of sensing locations, $S$, is routed using a TSP solver (such as the Christofides algorithm~\cite{Christofides22}) to generate a low-cost route visiting all locations $\{c_j : j \in S\}$.

\begin{algorithm}[!ht]
\caption{GreedyCover: Estimation Uncertainty Guaranteed Placement~\cite{JakkalaAOA26}}
\label{alg:greedy-coverage}
\begin{algorithmic}[1]
\Require Coverage matrix $B \in \{0,1\}^{M \times N}$, evaluation set $\mathcal{V}=\{x_1,\dots,x_N\}$.
\Ensure Selected index set $S \subseteq \{1,\dots,M\}$.
\State $S \gets \emptyset$ \Comment{selected sensing-location indices}
\State $U \gets \emptyset$ \Comment{covered evaluation points}
\While{$|U| < N$}
    \For{each candidate $j \in \{1,\dots,M\}\setminus S$}
        \State $g(j) \gets \bigl|\{x_i \in \mathcal{V} : B_{ji}=1,\ x_i \notin U\}\bigr|$ 
        \State \Comment{marginal gain}
    \EndFor
    \State $j^\star \gets \arg\max_{j} g(j)$
    \If{$g(j^\star) = 0$}
        \State \textbf{break} \Comment{no further improvement possible}
    \EndIf
    \State $S \gets S \cup \{j^\star\}$
    \State $U \gets U \cup \{x_i \in \mathcal{V} : B_{j^\star i}=1\}$ \label{line:greedy_coverage_update}
\EndWhile
\State \Return $S$
\end{algorithmic}
\end{algorithm}
\vspace{-6mm}

\subsection{Monotone Submodular Objective}
\label{sec:monotone_submodular}

The objective implicitly optimized by the GreedyCover algorithm is a classic monotone submodular set function~\cite{NemhauserWF78, Wolsey82}. By maximizing coverage encoded in the binary matrix $B$, the algorithm effectively optimizes the underlying GP predictive variance. For any selected index set $S \subseteq \{1,\dots,M\}$, the \emph{objective function} is defined as:

\begin{equation}
\label{eq:coverage_objective}
F(S)
:=
\left|\bigcup_{j \in S}\{x_i \in \mathcal{V} : B_{ji}=1\}\right|\,.
\end{equation}

This function $F(S)$ counts the total number of evaluation points whose predictive variance is bounded below the target threshold by the selected sensing locations. Because $F(S)$ exhibits diminishing returns---the marginal benefit of adding a new sensing location decreases as more of the environment already meets the target variance---it is a monotone and submodular objective~\cite{JakkalaAOA26}.

\subsection{Near-Optimal Approximation Guarantees}
\label{sec:approximation_guarantees}

Finding the exact minimal sensing set $S$ to cover all evaluation points in $\mathcal{V}$ is computationally intractable in general, as the underlying combinatorial problem is NP-hard. However, because the coverage objective $F(S)$ in Equation~\eqref{eq:coverage_objective} is nonnegative, monotone, and submodular, \textsc{GreedyCover} provides two complementary, near-optimal polynomial-time guarantees depending on the termination criterion:

\textbf{1) Bounded-Uncertainty Regime (Cardinality Guarantee):} When \textsc{GreedyCover} runs until complete domain coverage ($F(S) = N$) to strictly satisfy $\sigma_{\mathrm{post}}^2(x_i \mid S) \le \sigma_{\mathrm{tar}}^2$ across all $x_i \in \mathcal{V}$, it solves a submodular Set Cover instance. As established by Wolsey~\cite{Wolsey82}, the greedy algorithm achieves an approximation ratio bounded by the harmonic number of the maximum single-element coverage. If $S_{\mathrm{greedy}}$ denotes the sensing set returned upon termination and $S_{\mathrm{opt}}$ denotes the minimum-cardinality optimal set achieving full coverage, the required number of sensing locations is strictly bounded by:

\smallskip
\begin{tcolorbox}[
    colback=NeonBlue!10,      
    colframe=NeonBlue!50,     
    arc=4pt,                  
    boxrule=1pt,              
    left=6pt,
    right=6pt,
    top=2pt,
    bottom=2pt,
    before skip=6pt,
    after skip=6pt        
]
\begin{equation}
|S_{\mathrm{greedy}}| \le \left( 1 + \ln\left( \max_{j} F(\{j\}) \right) \right) |S_{\mathrm{opt}}|.
\end{equation}
\end{tcolorbox}
\smallskip

\textbf{2) Fixed-Budget Anytime Regime (Placement Quality Guarantee):} Because \textsc{GreedyCover} iteratively selects the candidate location maximizing marginal coverage gain $g(j)$, its prefix sequence of placements simultaneously solves the submodular Maximum Coverage problem~\cite{NemhauserWF78, krauseSG08}. Specifically, if execution is terminated early after selecting any fixed budget of $T$ sensing locations (denoted $S_T$), the number of evaluation points satisfying the target variance threshold $\sigma_{\mathrm{tar}}^2$ is guaranteed to be within a $(1 - 1/e)$ factor of the optimal $T$-sensor placement $S_T^*$:

\smallskip
\begin{tcolorbox}[
    colback=NeonBlue!10,      
    colframe=NeonBlue!50,     
    arc=4pt,                  
    boxrule=1pt,              
    left=6pt,
    right=6pt,
    top=2pt,
    bottom=2pt,
    before skip=6pt,
    after skip=6pt        
]
\begin{equation}
F(S_T) \ge \left(1 - \frac{1}{e}\right) \max_{|S| \le T} F(S) = \left(1 - \frac{1}{e}\right) F(S_T^*).
\end{equation}
\end{tcolorbox}
\smallskip

Thus, \textsc{GreedyCover} simultaneously guarantees $(1 - 1/e)$-optimal spatial coverage at every intermediate budget $T$ and $(1 + \ln N)$-optimal sensing cardinality upon reaching complete domain coverage ($F(S) = N$).

Note that both approximation guarantees hold with respect to the discrete coverage instance encoded by the binary matrix $B$. Directly optimizing raw GP predictive variance over combinatorial subsets is intractable because multi-point variance reduction does not, in general, exhibit submodularity. By leveraging the single-observation sufficient coverage threshold (Theorems~\ref{thm:single_point} and~\ref{thm:multi_point}), this formulation maps continuous uncertainty constraints into a monotone submodular set function—circumventing the non-submodularity of raw predictive variance while certifying that the resulting sensing set satisfies the target uncertainty bound.

\subsection{Hardness of Approximation}
\label{sec:approximation_hardness}

The approximation guarantees achieved by \textsc{GreedyCover} are not merely artifacts of the greedy heuristic; they represent the fundamental computational limits of the problem. To establish this, we prove a bidirectional equivalence between the GP uncertainty bounded sensing formulation and combinatorial coverage problems. As shown in Section~\ref{sec:optimization_algorithm}, Theorems~\ref{thm:single_point} and~\ref{thm:multi_point} directly map any continuous GP uncertainty bounded planning problem into a discrete coverage instance encoded by the binary matrix $B \in \{0,1\}^{M \times N}$. 

To establish that the computational hardness of combinatorial coverage applies to our problem, we prove the converse reduction: any arbitrary binary incidence matrix $B$ can be realized within the GP uncertainty bounded sensing formulation by constructing an appropriate non-stationary covariance kernel. By establishing an approximation-preserving reduction, Lemma~\ref{lem:hardness_reduction} demonstrates that uncertainty bounded IPP inherits the fundamental hardness-of-approximation lower bounds of both Minimum Set Cover and Maximum Coverage.

\begin{restatable}{lemma}{hardnessreduction}
\label{lem:hardness_reduction}
Embedding Arbitrary Set Cover Instances

Let $(U, \mathcal{S})$ be an arbitrary instance of the minimum Set Cover problem with ground universe $U = \{u_1, \dots, u_N\}$ and subsets $\mathcal{S} = \{S_1, \dots, S_M\}$, described by the binary incidence matrix $B \in \{0,1\}^{M \times N}$, where $B_{ji} = 1$ if and only if $u_i \in S_j$. For any spatial dimension $d \ge 2$, there exist distinct candidate locations $\mathcal{C} \subset \mathbb{R}^d$, evaluation locations $\mathcal{V} \subset \mathbb{R}^d$, observation noise $\sigma_n^2 > 0$, a target threshold $\sigma^2_{\mathrm{tar}} > 0$, and a valid non-stationary positive semi-definite kernel function $k$ such that

\begin{equation*}
\begin{split}
|k(c_j, x_i)| &\ge \sqrt{\left(k(x_i,x_i) - \sigma^2_{\mathrm{tar}}\right)\left(k(c_j,c_j) + \sigma_n^2\right)} \\
&\quad \iff B_{ji} = 1.
\end{split}
\end{equation*}

\proofnote
\end{restatable}

Because the exact spatial kernel embedding in Lemma~\ref{lem:hardness_reduction} is independent of the termination criterion, it directly establishes matching polynomial-time inapproximability thresholds for both operational regimes:

\textbf{1) Bounded-Uncertainty Regime (Cardinality Hardness):} When requiring complete domain coverage ($F(S) = N$), finding the minimum-cardinality sensing set is mathematically equivalent to the Minimum Set Cover problem. Feige~\cite{Feige98} demonstrated that Set Cover admits no polynomial-time approximation ratio better than $(1 - o(1)) \ln N$ unless NP problems are solvable in slightly superpolynomial time, and Dinur and Steurer~\cite{DinurS14} subsequently tightened this bound under the standard complexity assumption that $P \neq NP$. Therefore, over general non-stationary kernels, any polynomial-time algorithm yielding a valid sensing subset $S_{\mathrm{approx}}$ is strictly bounded in the worst case by:

\smallskip
\begin{tcolorbox}[
    colback=NeonBlue!10,
    colframe=NeonBlue!50,
    arc=4pt,
    boxrule=1pt,
    left=6pt,
    right=6pt,
    top=2pt,
    bottom=2pt,
    before skip=6pt,
    after skip=6pt        
]
\begin{equation}
\sup_{\mathbf{I}} \frac{|S_{\mathrm{approx}}(\mathbf{I})|}{|S_{\mathrm{opt}}(\mathbf{I})|} \ge (1 - o(1)) \ln N,
\end{equation}
\end{tcolorbox}
\smallskip

\noindent where the supremum ($\sup_\mathbf{I}$) is taken over all valid problem instances $\mathbf{I}$, and $S_{\mathrm{opt}}(\mathbf{I})$ denotes the minimal valid sensing subset. While an algorithm may efficiently solve trivial or uniform geometric configurations, adversarial non-stationary instances mathematically force its worst-case cardinality suboptimality factor to scale logarithmically with $N$.

\textbf{2) Fixed-Budget Anytime Regime (Placement Quality Hardness):} When a strict cardinality budget $|S| \le T$ prevents full domain coverage, maximizing the number of evaluation points satisfying $\sigma_{\mathrm{post}}^2(x_i \vert{} S) \le \sigma_{\mathrm{tar}}^2$ is mathematically equivalent to the Maximum $T$-Coverage problem. In the same landmark study~\cite{Feige98}, Feige proved that no polynomial-time algorithm can approximate Maximum $T$-Coverage within a factor better than $(1 - 1/e + \epsilon)$ for any $\epsilon > 0$ unless $P = NP$. Consequently, for any polynomial-time algorithm returning a $T$-sensor placement $S_{\mathrm{approx}, T}$, its worst-case coverage ratio against the optimal $T$-sensor placement $S_T^*$ is upper-bounded:

\smallskip
\begin{tcolorbox}[
    colback=NeonBlue!10,
    colframe=NeonBlue!50,
    arc=4pt,
    boxrule=1pt,
    left=6pt,
    right=6pt,
    top=2pt,
    bottom=2pt,
    before skip=6pt,
    after skip=6pt        
]
\begin{equation}
\inf_{\mathbf{I}} \frac{F(S_{\mathrm{approx}, T}(\mathbf{I}))}{F(S_T^*(\mathbf{I}))} \le 1 - \frac{1}{e} + o(1).
\end{equation}
\end{tcolorbox}
\smallskip

\noindent By achieving the $(1 + \ln N)$ cardinality upper bound of Wolsey~\cite{Wolsey82} upon full coverage and the $(1 - 1/e)$ coverage lower bound of Nemhauser et al.~\cite{NemhauserWF78} at every intermediate budget $T$, \textsc{GreedyCover} asymptotically matches both theoretical hardness thresholds. Under the assumption that $P \neq NP$, \textsc{GreedyCover} is simultaneously complexity-optimal in polynomial time for both bounded-uncertainty termination and fixed-budget sensor placement.

\section{Experiments}

The preceding theoretical analysis established predictive variance as the ideal objective function for Gaussian Process (GP)-based Informative Path Planning (IPP). It also established that GreedyCover achieves the theoretical performance limit when optimizing this metric in discrete settings. To empirically validate these findings and explore their practical implications, this section presents two benchmarks:

\begin{enumerate}
    \item \textbf{Information-Theoretic Objectives Benchmark:} Evaluating the performance of using predictive variance as an optimization objective versus other prominent metrics.
    \item \textbf{Optimization Methods Benchmark:} Evaluating the computational efficiency and estimation error quality of GreedyCover against prominent optimization algorithms.
\end{enumerate}

\subsection{Experimental Setup and Methodology}

All benchmark evaluations were executed on a workstation equipped with an AMD EPYC CPU with 16 cores and 32 GB of RAM, utilizing CPU-only operations. We conducted experiments using the SRTM dataset~\cite{SRTM}, presenting results for one primary location $(45^\circ\mathrm{N},\,123^\circ\mathrm{W})$ in this section and three additional locations in the appendix. 

The evaluation focuses on offline single-robot IPP, which assumes the GP hyperparameters are known \textit{a priori}. To guarantee hyperparameter consistency across evaluations, we trained an oracle GP using an Attentive kernel~\cite{ChenKL22} on the SRTM data, and sampled a true function from this model to generate our ground-truth evaluation data. 

Most IPP methods consider heuristic distance or cardinality constraints on sensing locations rather than the more useful maximum predictive variance (i.e., uncertainty) constraint. Therefore, we used the GreedyCover algorithm to establish a universal cardinality constraint. Specifically, we configured GreedyCover to bound the maximum predictive variance at 85\% of the maximum prior variance, yielding a budget of 102 sensing locations. This sensing budget was subsequently enforced across all methods in both benchmarks to ensure a fair comparison.

The quality of the generated paths is assessed using the following metrics, computed by comparing the GP predictive mean to the sampled ground-truth test data:
\begin{itemize}
    \item \textbf{Mean Squared Error (MSE):} The average squared discrepancy between predictions and ground truth:
    \begin{equation}
        \text{MSE} = \frac{1}{|\mathcal{V}|} \sum_{x_i \in \mathcal{V}} (f(x_i) - \mu_{\mathrm{post}}(x_i \mid \mathcal{P}))^2\,.
    \end{equation}

    \item \textbf{Path Length:} The total Euclidean distance traveled along the planned path $\mathcal{P} = (p_1, \dots, p_T)$:
    \begin{equation}
        L(\mathcal{P}) = \sum_{t=1}^{T-1} \|p_{t+1} - p_t\|_2\,.
    \end{equation}
    
    \item \textbf{Runtime:} The total computation time required for an algorithm to yield the solution path.
\end{itemize}

Each benchmark experiment was repeated ten times, with the mean and standard deviation reported in the results.

\subsection{Benchmark I: Information-Theoretic Objectives}

\begin{figure*}[!ht]
   \centering
    \begin{subfigure}{0.44\textwidth}
        \includegraphics[width=\textwidth]{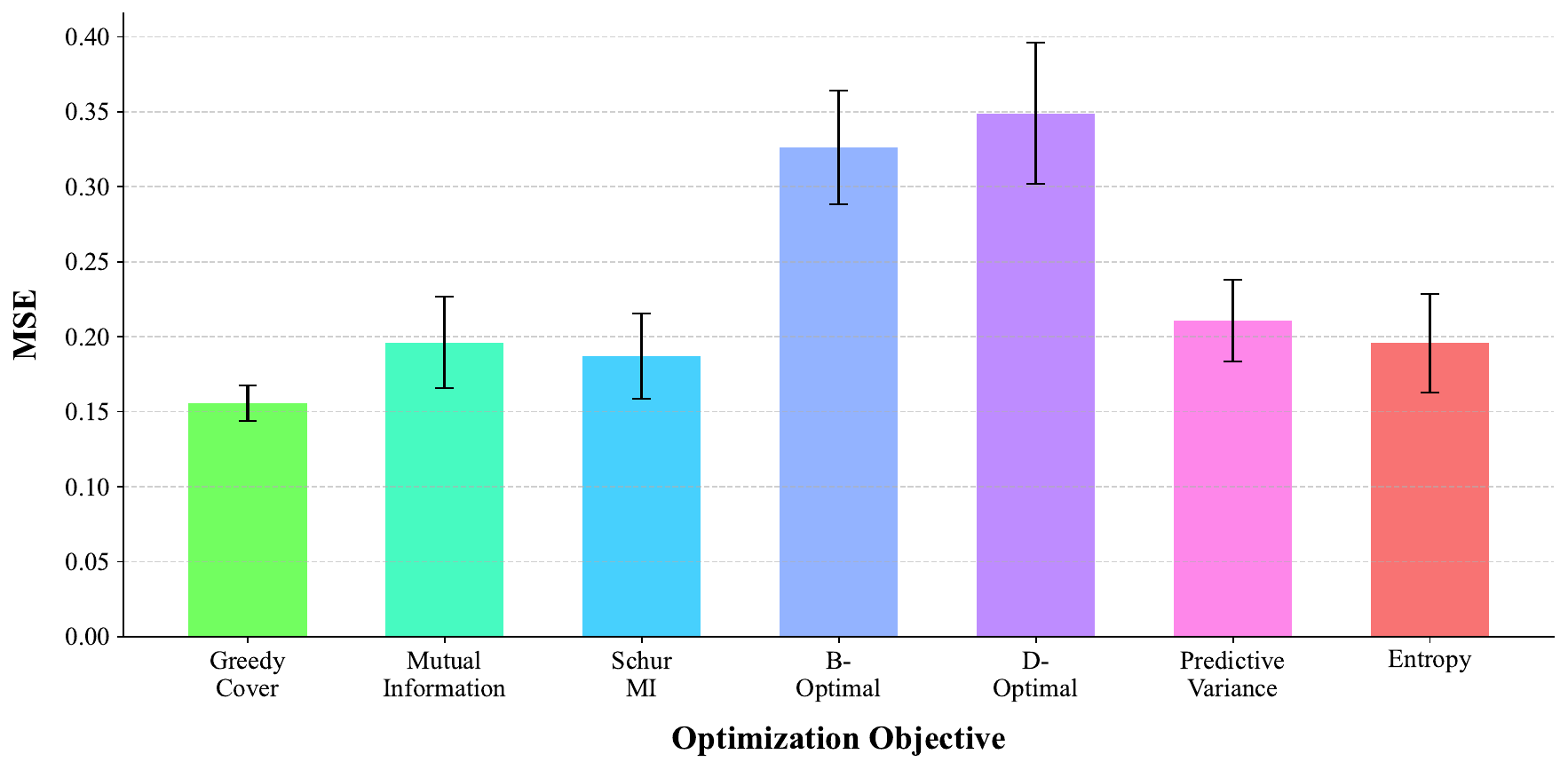}
    \end{subfigure}
    \hfill
    \begin{subfigure}{0.44\textwidth}
        \includegraphics[width=\textwidth]{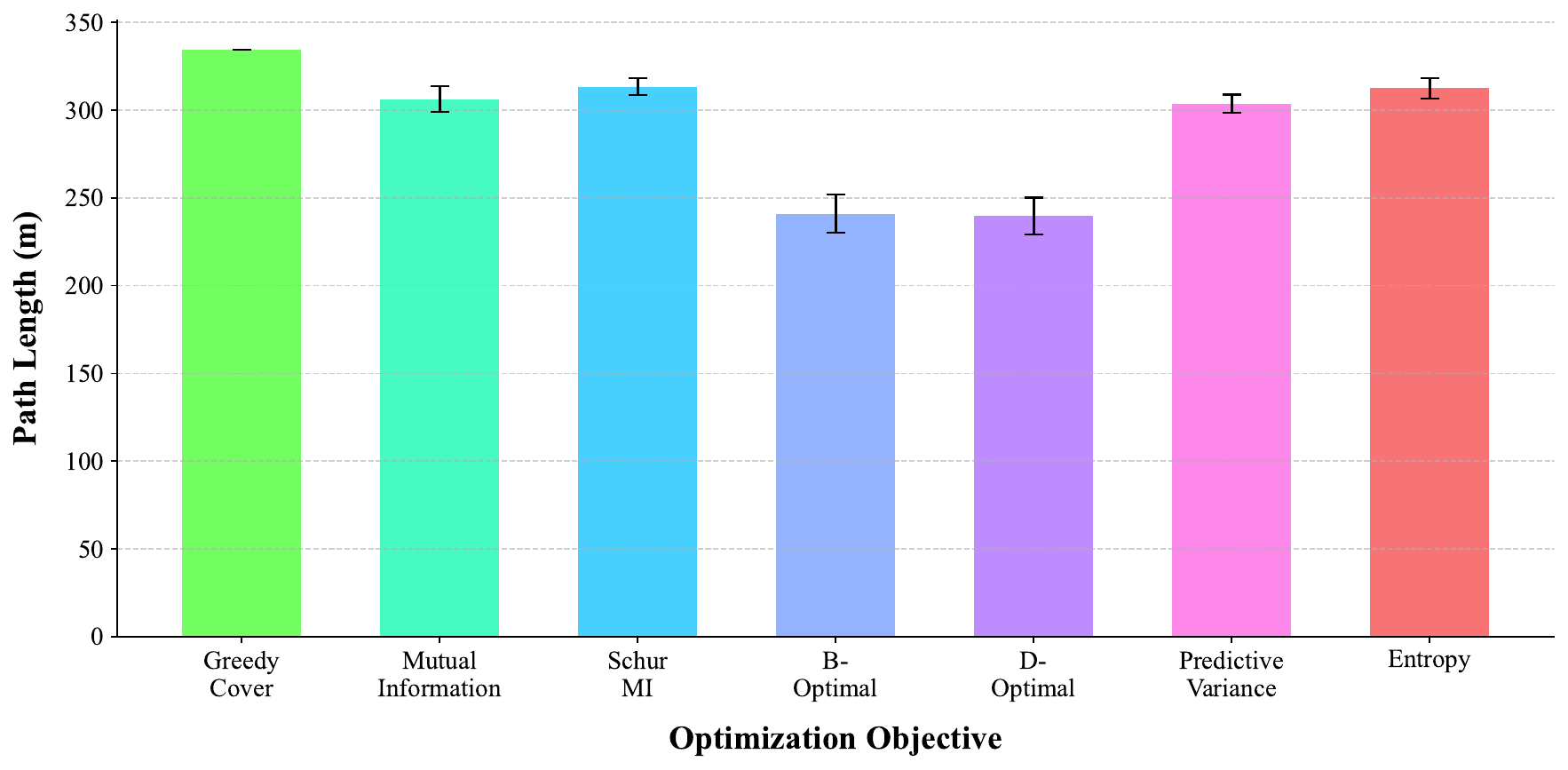}
    \end{subfigure}
    \caption{$(45^\circ\mathrm{N},\,123^\circ\mathrm{W})$ Estimation error (MSE) and path length across information-theoretic objectives. GreedyCover achieves lower overall error despite its discrete-space constraints, outperforming baseline objectives optimized in continuous space using CMA-ES.}  
    \label{fig:objective_metrics}
    \vspace{-4mm}
\end{figure*}

\begin{figure*}[!ht]
   \centering
    \begin{subfigure}{0.44\textwidth}
        \includegraphics[width=\textwidth]{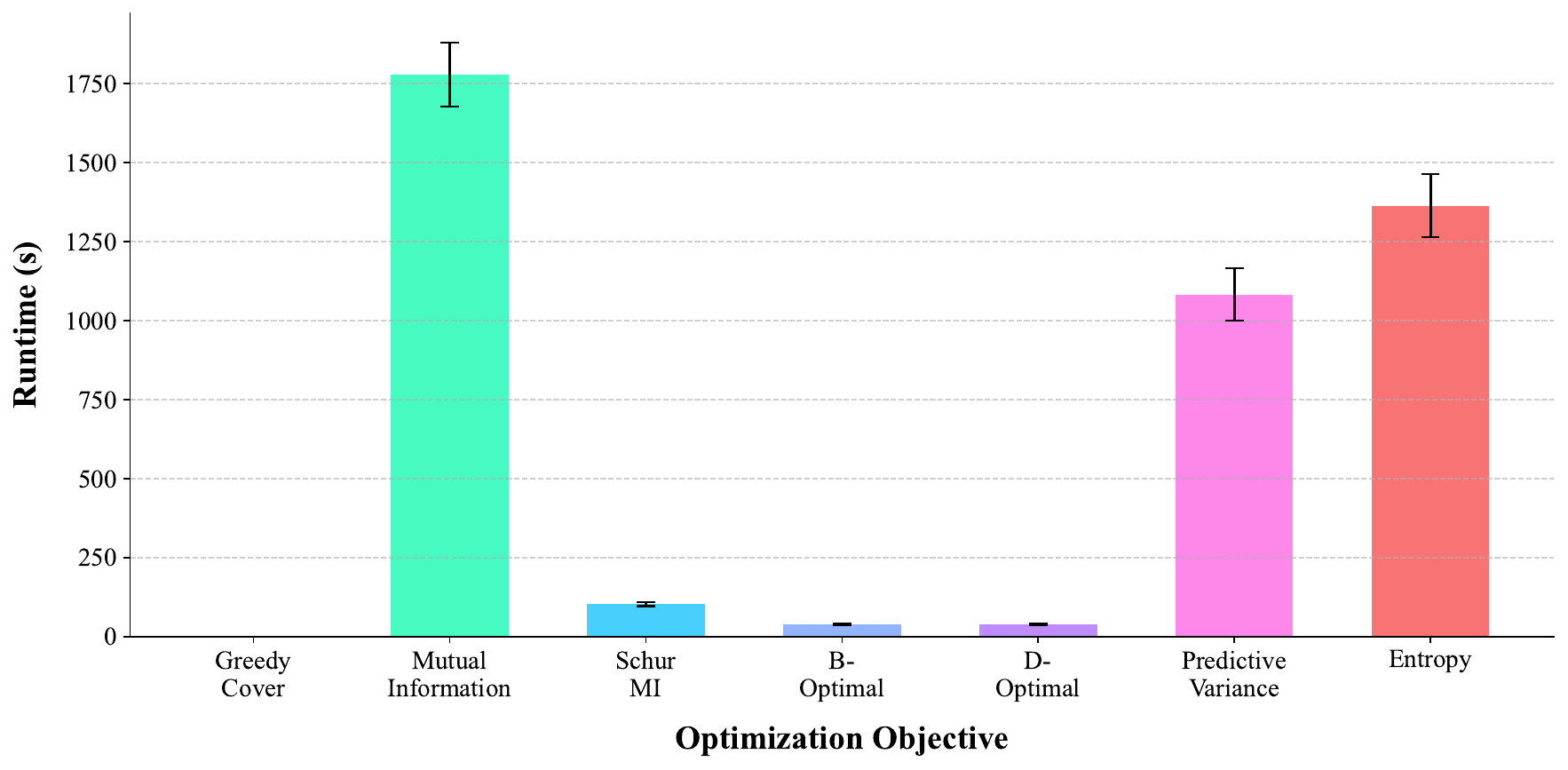}
    \end{subfigure}
    \hfill
    \begin{subfigure}{0.44\textwidth}
        \includegraphics[width=\textwidth]{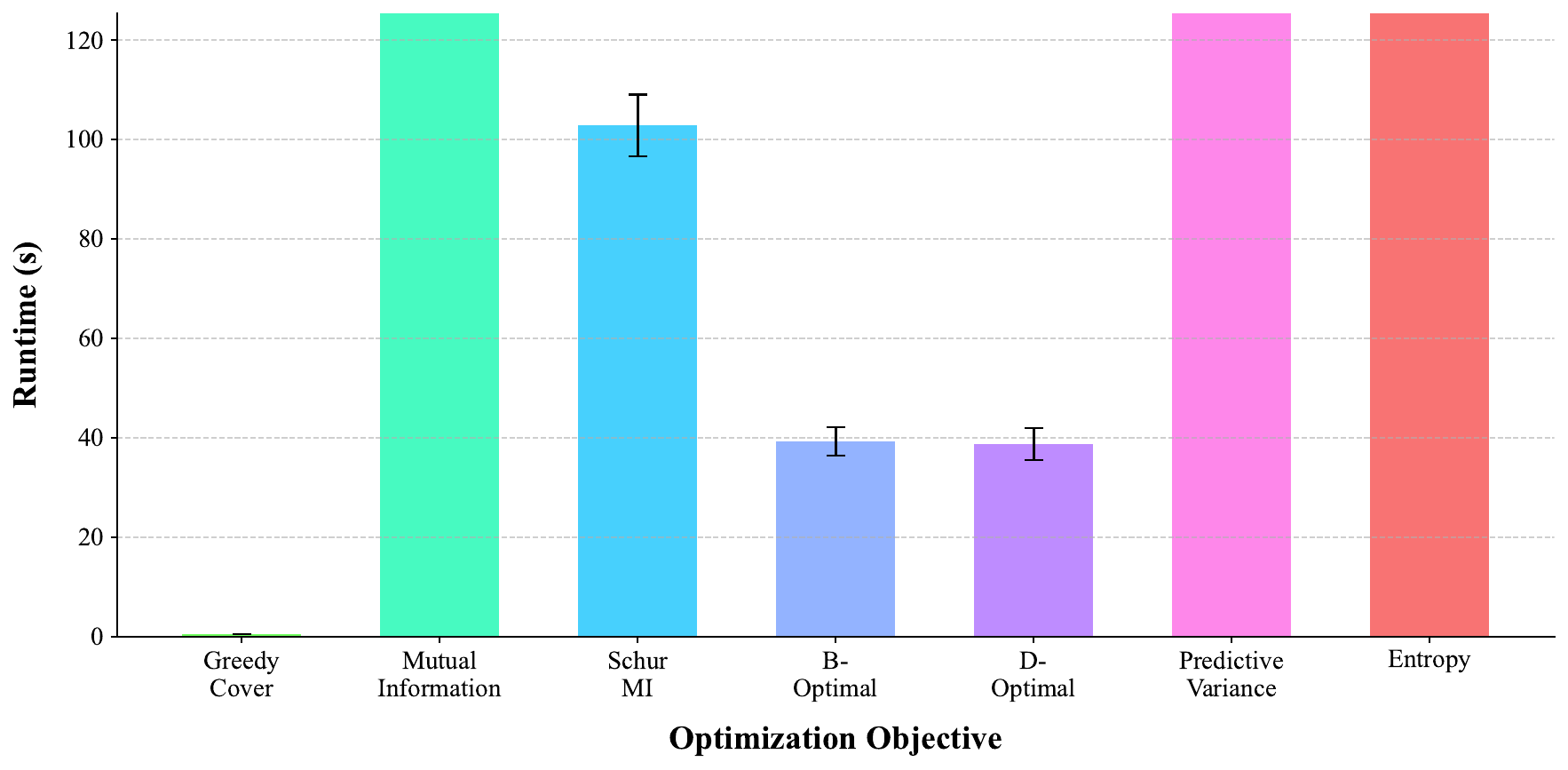}
    \end{subfigure}
    \caption{$(45^\circ\mathrm{N},\,123^\circ\mathrm{W})$ Runtime comparison of prominent information-theoretic objectives (right: zoomed version of the left). While continuous-space optimization severely inflates the search space and computational overhead for CMA-ES, GreedyCover leverages its discrete formulation to deliver solutions orders of magnitude faster.} 
    \label{fig:objective_runtime}
    \vspace{-4mm}
\end{figure*}

\smallskip
\begin{tcolorbox}[
    colback=NeonBlue!10,      
    colframe=NeonBlue!50,     
    arc=4pt,                  
    boxrule=1pt,              
    left=8pt,
    right=8pt,
    top=8pt,
    bottom=8pt,
    before skip=8pt,
    after skip=8pt        
]
    \textbf{Research Question:} How does GreedyCover's discrete predictive variance optimization compare against prominent information-theoretic metrics optimized with a black-box optimizer, such as CMA-ES?
    
    \vspace{0.5em}
    
    \textbf{Key Takeaway:} GreedyCover demonstrates a clear advantage. Despite operating over a discrete domain, it attains the lowest estimation error. Moreover, it incurs a notably lower computational cost by avoiding repetitive objective function evaluations across candidate solutions.
\end{tcolorbox}
\smallskip

The first benchmark empirically evaluates prominent information-theoretic objective functions detailed in Table~\ref{tab:objectives}. To assess these objectives without imposing discrete grid constraints, we employed the Covariance Matrix Adaptation Evolution Strategy (CMA-ES). As a derivative-free, black-box genetic algorithm, CMA-ES generally outperforms Bayesian Optimization when handling the complex, continuous-space utility functions typical of IPP~\cite{HitzGGPS17}. The results of this evaluation are shown in Figure~\ref{fig:objective_metrics} and Figure~\ref{fig:objective_runtime}.

\begin{figure*}[!ht]
   \centering
    \begin{subfigure}{0.44\textwidth}
        \includegraphics[width=\textwidth]{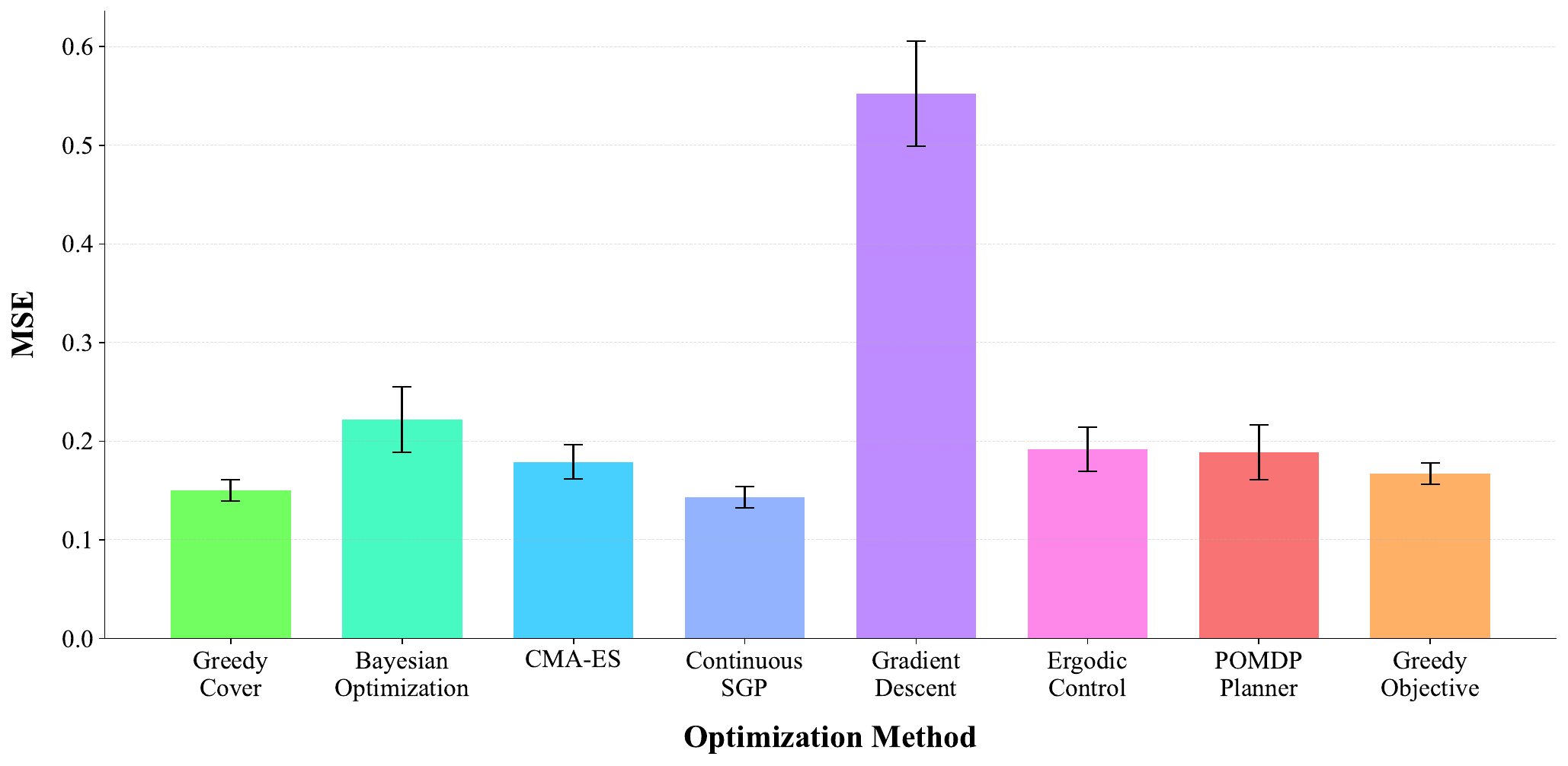}
    \end{subfigure}
    \hfill
    \begin{subfigure}{0.44\textwidth}
        \includegraphics[width=\textwidth]{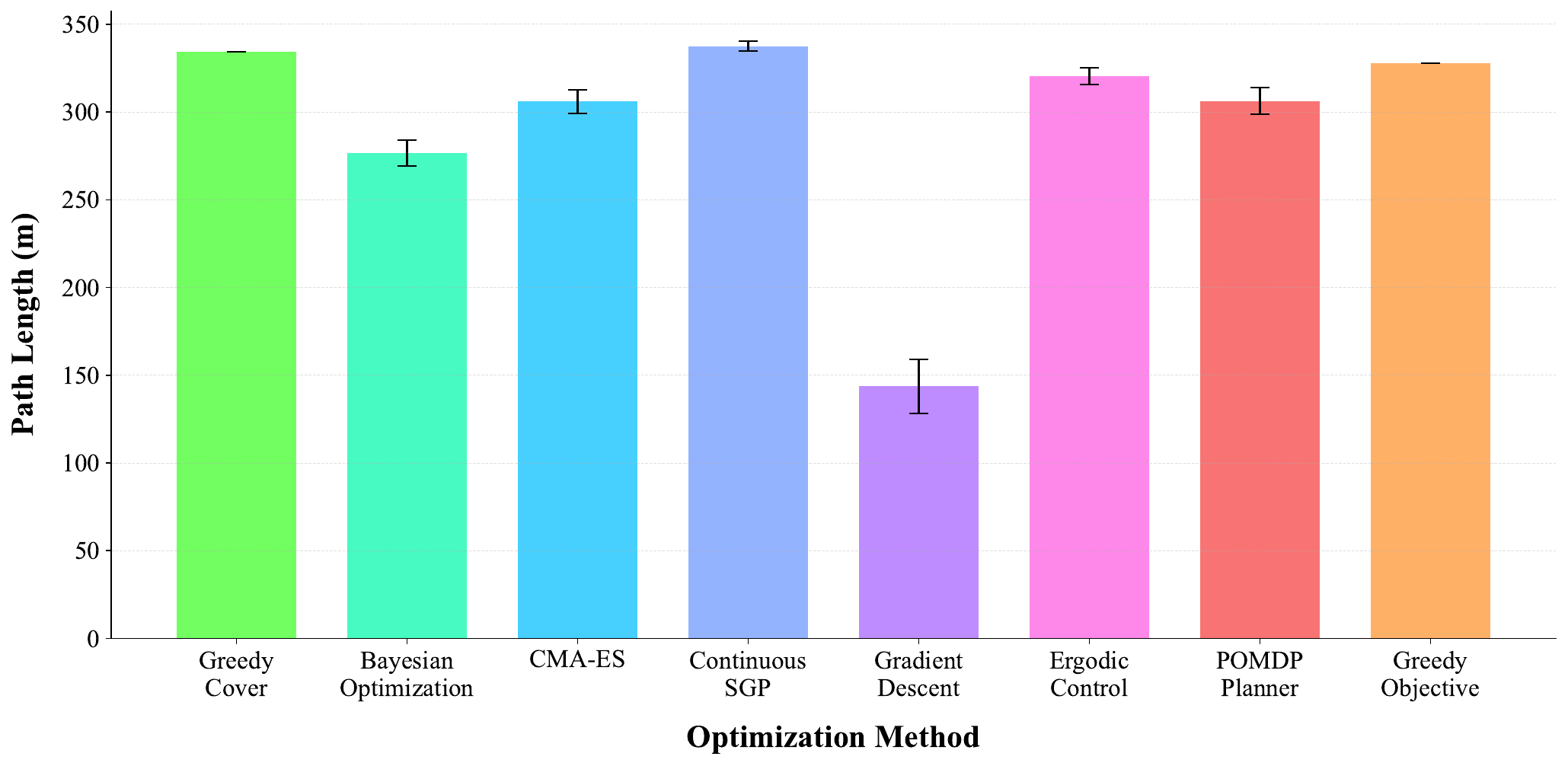}
    \end{subfigure}
    \caption{$(45^\circ\mathrm{N},\,123^\circ\mathrm{W})$ Estimation error (MSE) and path length across prominent optimization methods. GreedyCover matches the best-in-class accuracy of continuous planners while avoiding the susceptibility to local minima inherent in gradient-based approaches.}
    \label{fig:optimizers_metrics}
    \vspace{-4mm}
\end{figure*}

\begin{figure*}[!ht]
   \centering
    \begin{subfigure}{0.44\textwidth}
        \includegraphics[width=\textwidth]{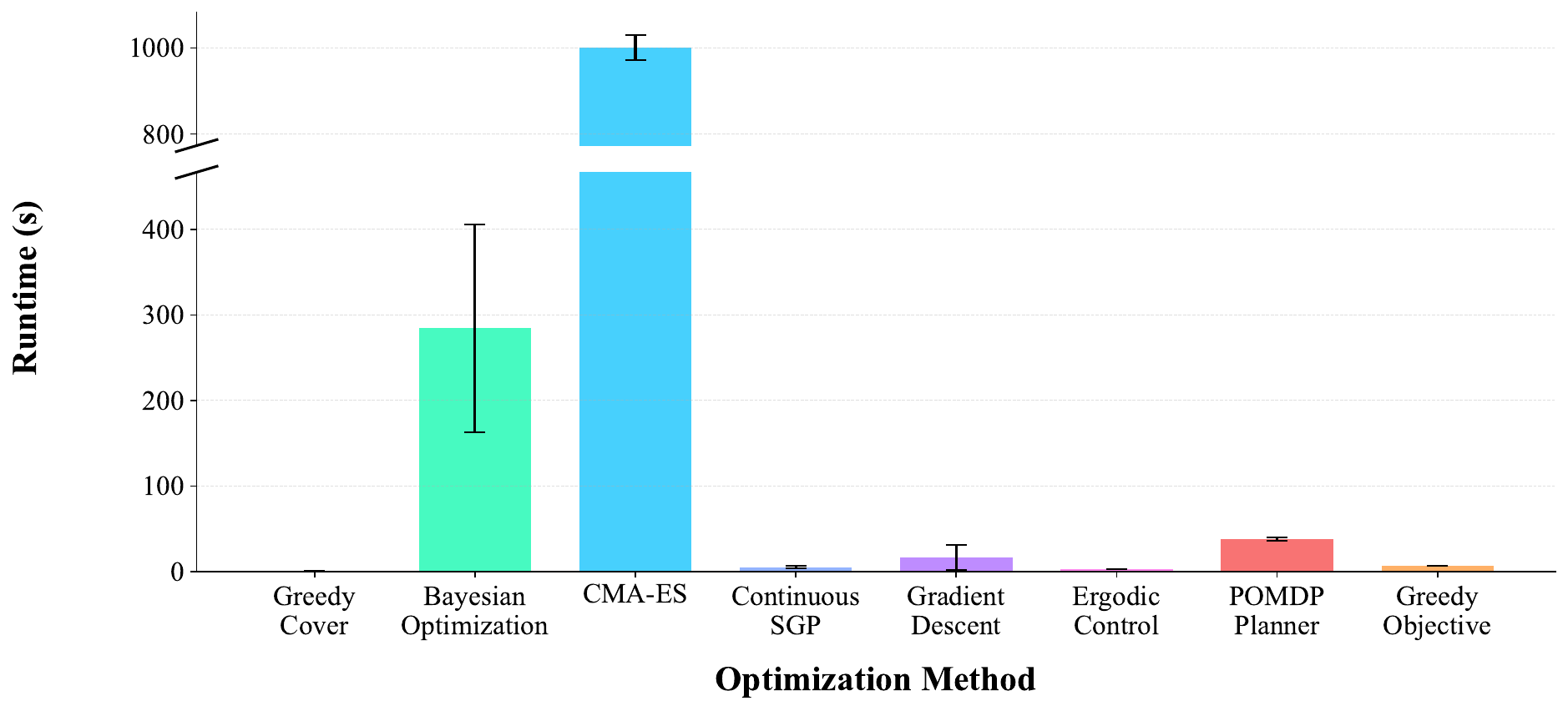}
    \end{subfigure}
    \hfill
    \begin{subfigure}{0.44\textwidth}
        \includegraphics[width=\textwidth]{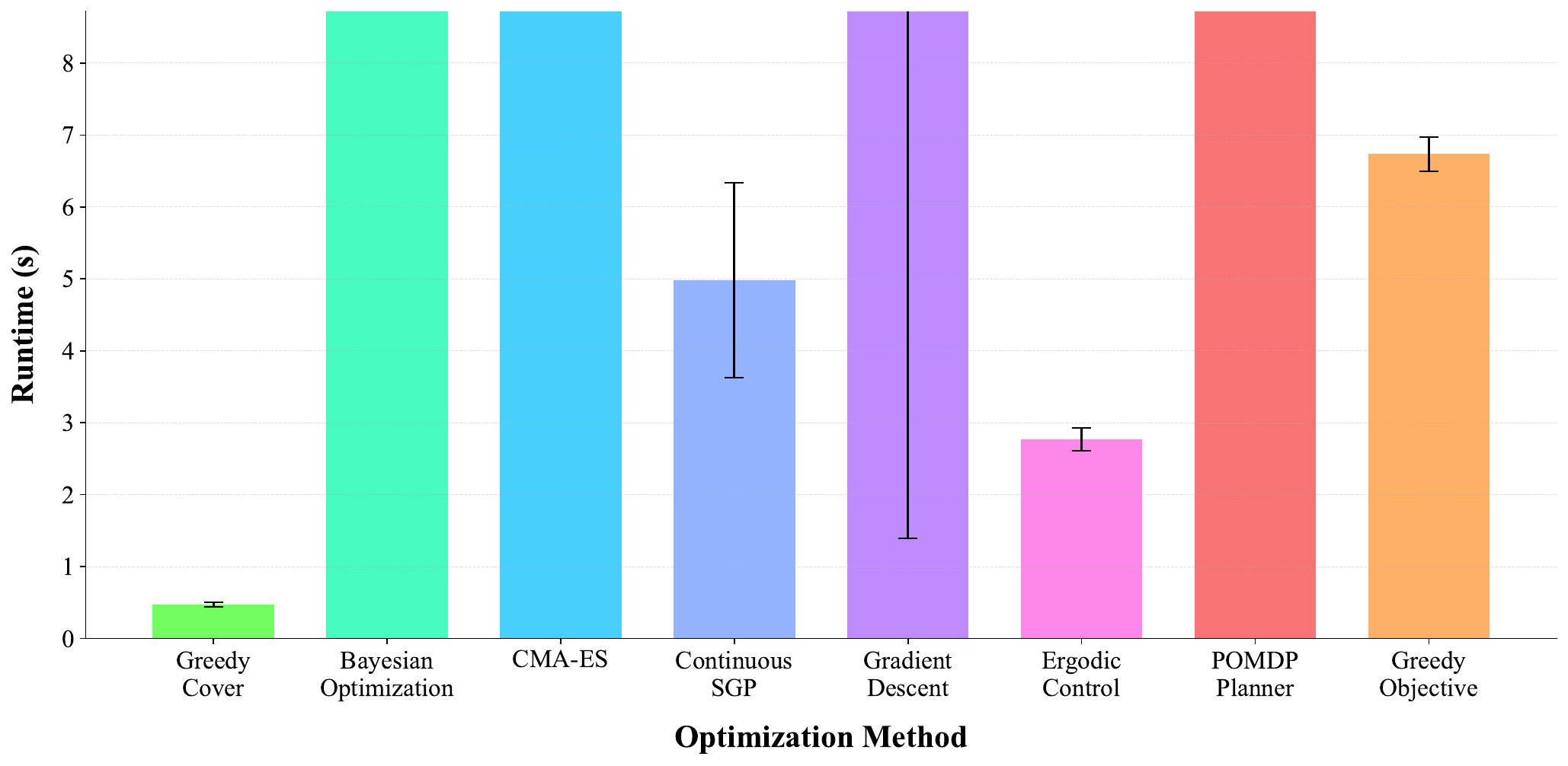}
    \end{subfigure}
    \caption{$(45^\circ\mathrm{N},\,123^\circ\mathrm{W})$ Runtime comparison of prominent optimization methods (right: zoomed version of the left). GreedyCover delivers unparalleled computational speed, sharply contrasting with competing baselines that are hindered by the need for numerous, expensive objective function evaluations.}  
    \label{fig:optimizers_runtime}
    \vspace{-4mm}
\end{figure*}

\begin{table}[!t]
\centering
\renewcommand{\arraystretch}{1.4} 

\rowcolors{2}{NeonBlue!15}{white} 

\begin{tabularx}{\columnwidth}{l X}
\rowcolor{NeonBlue!50} 
\textbf{Objective} & \textbf{Equation Formulation} \\ 
\textbf{Mutual Information}~\cite{krauseSG08} & $\log|K_{\mathcal{P}\mathcal{P}}| + \log|K_{\mathcal{V}\mathcal{V}}| - \log|K_{\mathrm{combined}}|$ \\
\textbf{Schur MI}~\cite{JakkalaOA26} & $\log|K_{\mathcal{P}\mathcal{P}}| - \log|K_{\mathcal{P}\mathcal{P}} - K_{\mathcal{P}\mathcal{V}} K_{\mathcal{V}\mathcal{V}}^{-1} K_{\mathcal{V}\mathcal{P}}|$ \\
\textbf{B-Optimal}~\cite{OttBK23} & $\text{Tr}(K_{\mathcal{P}\mathcal{P}}^{-1})$ \\ 
\textbf{D-Optimal}~\cite{krauseSG08} & $-\log|K_{\mathcal{P}\mathcal{P}}|$ \\ 
\textbf{Predictive Variance}~\cite{RasmussenW05} & $-\text{Tr}(\Sigma_{\mathrm{post}})$ \\ 
\textbf{Entropy}~\cite{RasmussenW05} & $-0.5 \log|\Sigma_{\mathrm{post}}|$ \\ 
\end{tabularx}
\caption{Information-theoretic objective functions used in the benchmark.}
\label{tab:objectives}
\vspace{-5mm}
\end{table}

\textbf{Analysis:} Consistent with our theoretical analysis establishing predictive variance as the optimal objective for minimizing worst-case estimation error in offline GP-based IPP, GreedyCover empirically achieves the greatest error reduction.

In contrast, the continuous optimization baselines using CMA-ES—regardless of the objective metric—fail to match the performance floor established by GreedyCover. This shortcoming stems from CMA-ES operating over the unconstrained continuous domain, which drastically expands the search space and exacerbates optimization difficulty. Furthermore, unlike GreedyCover, it inherently lacks theoretical guarantees for near-optimality. While running CMA-ES significantly beyond its current 500-iteration budget would likely yield improved solutions, doing so is computationally prohibitive given its runtime overhead. In contrast, although GreedyCover restricts sampling to a discretized domain, its provable near-optimality within that space accounts for its superior error reduction.

Path length is inversely correlated with MSE, as methods that achieve lower estimation error require greater travel distances. This trade-off suggests that each method finds effective solutions relative to its resulting path length. However, GreedyCover holds two distinct advantages: it is the only approach that provides provable finite-time near-optimality guarantees, and it directly determines the required sensing budget from a target variance threshold, whereas all baseline methods require the number of sensing locations to be specified a priori.

Moreover, GreedyCover is significantly faster than the baselines. This efficiency gain stems from the fact that CMA-ES requires numerous objective function evaluations across candidate solution sets—which, depending on the objective metric, can scale cubically with the number of sensing locations per evaluation. In contrast, GreedyCover only computes the covariances of individual candidate sensing locations and relies on efficient binary operations to achieve fast execution.

\subsection{Benchmark II: Optimization Methods}

\smallskip
\begin{tcolorbox}[
    colback=NeonBlue!10,      
    colframe=NeonBlue!50,     
    arc=4pt,                  
    boxrule=1pt,              
    left=8pt,
    right=8pt,
    top=8pt,
    bottom=8pt,
    before skip=8pt,
    after skip=8pt
]
    \textbf{Research Question:} How does GreedyCover compare to prominent optimization methods in balancing estimation error and runtime efficiency?
        
    \vspace{0.5em}
    
    \textbf{Key Takeaway:} GreedyCover consistently achieves the best balance between error reduction and runtime, while matching the individual best-in-class performance across both metrics. Moreover, GreedyCover is the only algorithm that achieves the theoretical limit of approximation for its problem class, providing provable near-optimality that guarantees high-quality solutions across varied environments and GP hyperparameters.
    
\end{tcolorbox}

\begin{table}[!t]
\centering
\renewcommand{\arraystretch}{1.4} 

\rowcolors{2}{NeonBlue!15}{white} 

\begin{tabularx}{\columnwidth}{l X}
\rowcolor{NeonBlue!50} 
\textbf{Optimization Method} & \textbf{Description} \\ 
\textbf{GreedyCover}~\cite{JakkalaAOA26} & A near-optimal discrete algorithm that utilizes a submodular set-cover formulation to strictly constrain predictive variance. \\ 
\textbf{Bayesian Optimization}~\cite{FrancisOMR19} & A continuous-space black-box Bayesian optimization method utilizing surrogate modeling for objective maximization. \\ 
\textbf{CMA-ES}~\cite{HitzGGPS17} & A derivative-free, continuous-space evolutionary algorithm based on covariance matrix adaptation. \\ 
\textbf{Continuous SGP}~\cite{JakkalaA25b} & A gradient-based continuous space path planner leveraging variational Sparse Gaussian Process approximations. \\ 
\textbf{Gradient Descent}~\cite{JakkalaOA26} & A continuous-space method using exact analytical gradients with the Adam optimizer. \\
\textbf{Ergodic Control}~\cite{MillerSMM16} & A continuous-space trajectory planner that optimizes paths to match the spatial distribution of an underlying information map. \\ 
\textbf{POMDP Planner}~\cite{RayasDCS22} & A Partially Observable Markov Decision Process formulation solved via Monte Carlo Tree Search with Upper Confidence Bounds. \\ 
\textbf{Greedy Objective}~\cite{krauseSG08} & A discrete algorithm that sequentially selects candidate locations to maximize marginal objective gains. \\
\end{tabularx}
\caption{Informative path planning optimization methods used in the benchmark.}
\label{tab:methods}
\vspace{-6mm}
\end{table}

The second benchmark compares prominent IPP optimization methods detailed in Table~\ref{tab:methods}. Except for Continuous SGP, all evaluated algorithms are configured to optimize predictive variance. Note that while the Greedy Objective optimizer greedily optimizes the raw predictive variance, GreedyCover leverages its submodular formulation. The comparative results are presented in Fig.~\ref{fig:optimizers_metrics} and Fig.~\ref{fig:optimizers_runtime}.

\textit{Note:} Baseline methods were implemented using established Python libraries where possible. While these offer practical runtime approximations, they may not capture the absolute optimal performance of each approach. Implementation details are available in the code appendix.

\textbf{Analysis:}
GreedyCover consistently achieves low estimation error while maintaining fast execution times. With the exception of Gradient Descent, all baseline methods exhibit comparable MSE and path length performance. In terms of runtime, however, GreedyCover outperforms all baselines, consistently converging in under one second across all experimental trials.

Moreover, Continuous SGP does not optimize predictive variance, thereby foregoing the theoretical error bounds associated with this objective. While Ergodic Control proves to be a viable continuous-space planner—achieving estimation errors comparable to GreedyCover with reasonable execution times—its practical deployment remains challenging. Specifically, there is currently no principled approach to determine the sensing budget required to satisfy bounded uncertainty constraints, nor does the method provide finite-time convergence guarantees.
\section{Key Findings and Conclusion}

Our theoretical analysis and empirical evaluations yield critical insights into Informative Path Planning (IPP), enabling us to formally establish rigorous bounds in offline Gaussian Process (GP)-based IPP and identify fundamental gaps in continuous-space and online paradigms.

\subsection{The Resolution of Discrete Offline IPP}

Theorem~\ref{thm:nf_se_upper_bound} establishes that the pointwise squared error is upper-bounded by the product of the RKHS norm $\Vert{}f\Vert{}_{\mathcal{H}_k}$ and the GP predictive variance $\sigma^2_{\mathrm{post}}$. In offline IPP, \textit{a priori} knowledge of the hyperparameters ensures that the RKHS norm is a fixed constant reflecting the complexity of the underlying data field. Consequently, minimizing the predictive variance across candidate sensing locations emerges as the mathematically ideal objective to reduce the worst-case estimation error.

GreedyCover (Algorithm~\ref{alg:greedy-coverage}) operationalizes this property by enforcing a maximum predictive variance threshold, thereby bounding pointwise squared error up to an unknown RKHS norm scaling factor. Formulated as a submodular Set Cover problem, GreedyCover attains near-optimal approximation guarantees that match the fundamental hardness-of-approximation limits for the problem class. Coupled with an efficient $\mathcal{O}(M^2N)$ complexity and thorough empirical validation, the algorithm provides a theoretically and practically ideal solution.

Nevertheless, a key limitation of \textsc{GreedyCover} is its decoupled approach, which separates sensor placement from path generation via TSP post-processing. While this decoupling guarantees polynomial-time optimality with respect to sensing set cardinality and fixed-budget coverage alongside a constant-factor approximation on the subsequent path length, minimizing the number of sensing locations does not necessarily minimize total travel distance; multiple valid sensing sets of identical cardinality may satisfy the uncertainty threshold, yet their spatial configurations can yield different path lengths. Extending these complexity-optimal error and approximation guarantees to a fully coupled formulation remains an important direction for future work.

\smallskip
\begin{tcolorbox}[
    colback=NeonBlue!10,      
    colframe=NeonBlue!50,     
    arc=4pt,                  
    boxrule=1pt,              
    left=8pt, right=8pt, top=8pt, bottom=8pt,
    before skip=8pt, after skip=8pt       
]
GreedyCover optimizes predictive variance—a theoretically ideal objective for IPP—and achieves the tightest possible approximation guarantee attainable in polynomial time, establishing it as a robust approach for discrete offline IPP under bounded uncertainty constraints.
\end{tcolorbox}
\smallskip

\subsection{The Frontier of Continuous-Space Planning}

Establishing similarly rigorous bounds for continuous-space IPP is the next logical frontier. Currently, GreedyCover empirically outperforms most continuous-space planners in both runtime and error reduction. Ergodic Control represents a promising continuous-space alternative, although it currently relies on heuristic path-length constraints and provides only asymptotic convergence guarantees.

\smallskip
\begin{tcolorbox}[
    colback=NeonBlue!10,      
    colframe=NeonBlue!50,     
    arc=4pt,                  
    boxrule=1pt,              
    left=8pt, right=8pt, top=8pt, bottom=8pt,
    before skip=8pt, after skip=8pt       
]
Continuous-space IPP remains an open problem. Bridging the gap with discrete formulations requires establishing finite-time approximation guarantees while transitioning from heuristic path-length budgets to principled, bounded uncertainty constraints.
\end{tcolorbox}
\smallskip

\subsection{Online IPP and Model Uncertainty}

Online IPP—where hyperparameters must be learned dynamically during deployment—also remains an open challenge. Although predictive variance provides a theoretically grounded objective, naively optimizing it using deterministic estimates of GP hyperparameters is insufficient. Existing online methods focus exclusively on spatial uncertainty arising from data sparsity, an assumption valid only in offline settings with fixed, known hyperparameters. In online IPP, planners must also account for epistemic model uncertainty by explicitly incorporating the uncertainty in the hyperparameters themselves.

\smallskip
\begin{tcolorbox}[
    colback=NeonBlue!10,      
    colframe=NeonBlue!50,     
    arc=4pt,                  
    boxrule=1pt,              
    left=8pt, right=8pt, top=8pt, bottom=8pt,
    before skip=8pt, after skip=8pt       
]
Online IPP must explicitly account for hyperparameter uncertainty, enabling planners to collect informative data that jointly reduces spatial uncertainty from data sparsity and epistemic model uncertainty.
\end{tcolorbox}
\smallskip

\bibliography{references}

\appendices
\appendix

\subsection{Proofs}
\label{sec:proofs}

\gdef\proofnote{}  

The proof of Corollary~\ref{thm:se_upper_bound} and Lemma~\ref{lem:hardness_reduction} are presented in this appendix.


\seupperbound*

\begin{proof}
Starting from the absolute deterministic prediction error bound established previously, we square both sides to map the bound to the squared estimation error objective:
\[
(f(x) - \mu_{\mathrm{post}}(x \vert{} \mathcal{P}))^2 \le \big( \Vert{}f\Vert{}_{\mathcal{H}_k} P_{\mathcal{P}}(x) + \sqrt{n} \varepsilon_\star \Lambda_k(x) \big)^2
\]
To decouple the variance term from the measurement noise term, we apply the standard algebraic identity $(a+b)^2 \le 2a^2 + 2b^2$:
\[
(f(x) - \mu_{\mathrm{post}}(x \vert{} \mathcal{P}))^2 \le 2 \big( \Vert{}f\Vert{}_{\mathcal{H}_k} P_{\mathcal{P}}(x) \big)^2 + 2 \big( \sqrt{n} \varepsilon_\star \Lambda_k(x) \big)^2
\]
Because the power function $P_{\mathcal{P}}(x)$ represents the worst-case interpolation error norm in a noise-free setting, its square is precisely equal to the Gaussian Process predictive variance $\sigma_{\mathrm{post}}^2(x \vert{} \mathcal{P})$. Distributing the squares and substituting $P_{\mathcal{P}}(x)^2 = \sigma_{\mathrm{post}}^2(x \vert{} \mathcal{P})$ yields the final squared bound:
\[
(f(x) - \mu_{\mathrm{post}}(x \vert{} \mathcal{P}))^2 \le 2 \Vert{}f\Vert{}_{\mathcal{H}_k}^2 \sigma^2_{\mathrm{post}}(x \vert{} \mathcal{P}) + 2n \varepsilon_\star^2 \Lambda_k^2(x)
\]
\end{proof}


\hardnessreduction*

\begin{proof}
This instance of the minimum Set Cover problem is uniquely encoded by the binary incidence matrix $B \in \{0,1\}^{M \times N}$, where $B_{ji} = 1$ if $u_i \in S_j$ and $0$ otherwise. We construct an exact spatial embedding in $\mathbb{R}^d$ for $d \ge 2$ through the following sequence of steps.

\emph{Spatial embedding.} Select $M$ candidate locations $\mathcal{C} = \{c_1, \dots, c_M\}$ and $N$ evaluation locations $\mathcal{V} = \{x_1, \dots, x_N\}$ in $\mathbb{R}^d$ such that all $M + N$ locations are distinct. Let $\delta > 0$ be the minimum Euclidean distance between any two distinct locations in $\mathcal{C} \cup \mathcal{V}$.

\emph{High-dimensional feature space.} We construct a latent feature map $\phi: \mathbb{R}^d \to \mathbb{R}^{M+N}$ to represent the Set Cover topology. Let $\{\mathbf{e}_1, \dots, \mathbf{e}_M\}$ form the standard orthonormal basis for the first $M$ dimensions, and $\{\mathbf{v}_1, \dots, \mathbf{v}_N\}$ form the basis for the remaining $N$ dimensions. We assign discrete feature vectors at the designated spatial locations. For each candidate sensor $c_j \in \mathcal{C}$, define $\phi(c_j) = \mathbf{e}_j$. For each evaluation location $x_i \in \mathcal{V}$, define

$$\phi(x_i) = \sum_{m=1}^M B_{mi} \mathbf{e}_m + \sqrt{M - d_i} \, \mathbf{v}_i\,,$$

\noindent where $d_i = \sum_{m=1}^M B_{mi} \ge 1$ is the degree of element $u_i$.

\emph{Continuous domain extension.} To ensure the Gaussian process is mathematically valid over the entire continuous domain $\mathbb{R}^d$, we extend the discrete feature map $\phi$ into a globally continuous function $\Phi: \mathbb{R}^d \to \mathbb{R}^{M+N}$. Let $b: \mathbb{R}^d \to [0,1]$ be a smooth, compactly supported radial bump function such that $b(\mathbf{0}) = 1$ and $b(z) = 0$ for all $\Vert{}z\Vert{} \ge \delta/2$. Define the continuous field

$$\Phi(z) = \sum_{j=1}^M \phi(c_j) b(z - c_j) + \sum_{i=1}^N \phi(x_i) b(z - x_i)\,.$$

\noindent Because the bump functions' supports are strictly disjoint, the summation evaluates to exactly $\Phi(c_j) = \phi(c_j)$ and $\Phi(x_i) = \phi(x_i)$ at the locations of the sensors and evaluation points, and yields the zero vector in empty space.

\emph{Kernel construction.} Define the non-stationary covariance kernel via the inner product $k(z, z') = \langle \Phi(z), \Phi(z') \rangle$. As the inner product of an explicit feature map $\Phi(z)$, $k(z, z') = \langle \Phi(z), \Phi(z') \rangle$ is symmetric and positive semi-definite by construction. Evaluating the covariance at our specific points yields a prior variance at candidates of $k(c_j, c_j) = \Vert{}\mathbf{e}_j\Vert{}^2 = 1$, and a prior variance at evaluation points of $k(x_i, x_i) = \Vert{}\sum_{m=1}^M B_{mi} \mathbf{e}_m\Vert{}^2 + \Vert{}\sqrt{M - d_i} \, \mathbf{v}_i\Vert{}^2 = d_i + (M - d_i) = M$. The cross-covariance evaluates exactly to the incidence matrix

$$k(c_j, x_i) = \left\langle \mathbf{e}_j, \sum_{m=1}^M B_{mi} \mathbf{e}_m + \sqrt{M - d_i} \, \mathbf{v}_i \right\rangle = B_{ji}\,.$$

\emph{Threshold calibration.} To map this back to the continuous informative path planning threshold, we select an observation noise $\sigma_n^2 > 0$ and target variance $\sigma^2_{\mathrm{tar}} \in (0, k(x_i, x_i))$ such that the threshold $\tau = \sqrt{(k(x_i, x_i) - \sigma^2_{\mathrm{tar}})(k(c_j, c_j) + \sigma_n^2)}$ perfectly acts as a binary classifier. Let $\sigma_n^2 = 0.25$ and $\sigma^2_{\mathrm{tar}} = M - 0.2$. Since $M \ge 1$, we satisfy $0 < \sigma^2_{\mathrm{tar}} < M$.

\emph{Equivalence to Set Cover incidence.} Substituting the prior variances yields

$$\tau = \sqrt{(M - (M - 0.2))(1 + 0.25)} = \sqrt{0.2 \times 1.25} = 0.5$$

According to the cross-covariance computed previously, $k(c_j, x_i) = 1$ if $B_{ji} = 1$, and $0$ if $B_{ji} = 0$. Therefore, we achieve the strict equivalence

$$\vert{}k(c_j, x_i)\vert{} \ge \tau \iff B_{ji} \ge 0.5 \iff B_{ji} = 1$$

This proves that the predictive variance bounds derived from this constructed non-stationary Gaussian process kernel exactly reproduce the arbitrary Set Cover incidence matrix $B$.
\end{proof}


\subsection{Additional Benchmark Results}

This section evaluates benchmark performance across three additional regions from the SRTM dataset. Specifically, regions at $17^\circ\mathrm{N},\,073^\circ\mathrm{W}$, $02^\circ\mathrm{N},\,021^\circ\mathrm{W}$, and $47^\circ\mathrm{N},\,124^\circ\mathrm{W}$ require 103, 119, and 117 sensing locations, respectively, to satisfy the 85\% uncertainty bound.


\begin{figure*}[!ht]
   \centering
    \begin{subfigure}{0.49\linewidth}
        \includegraphics[width=\linewidth]{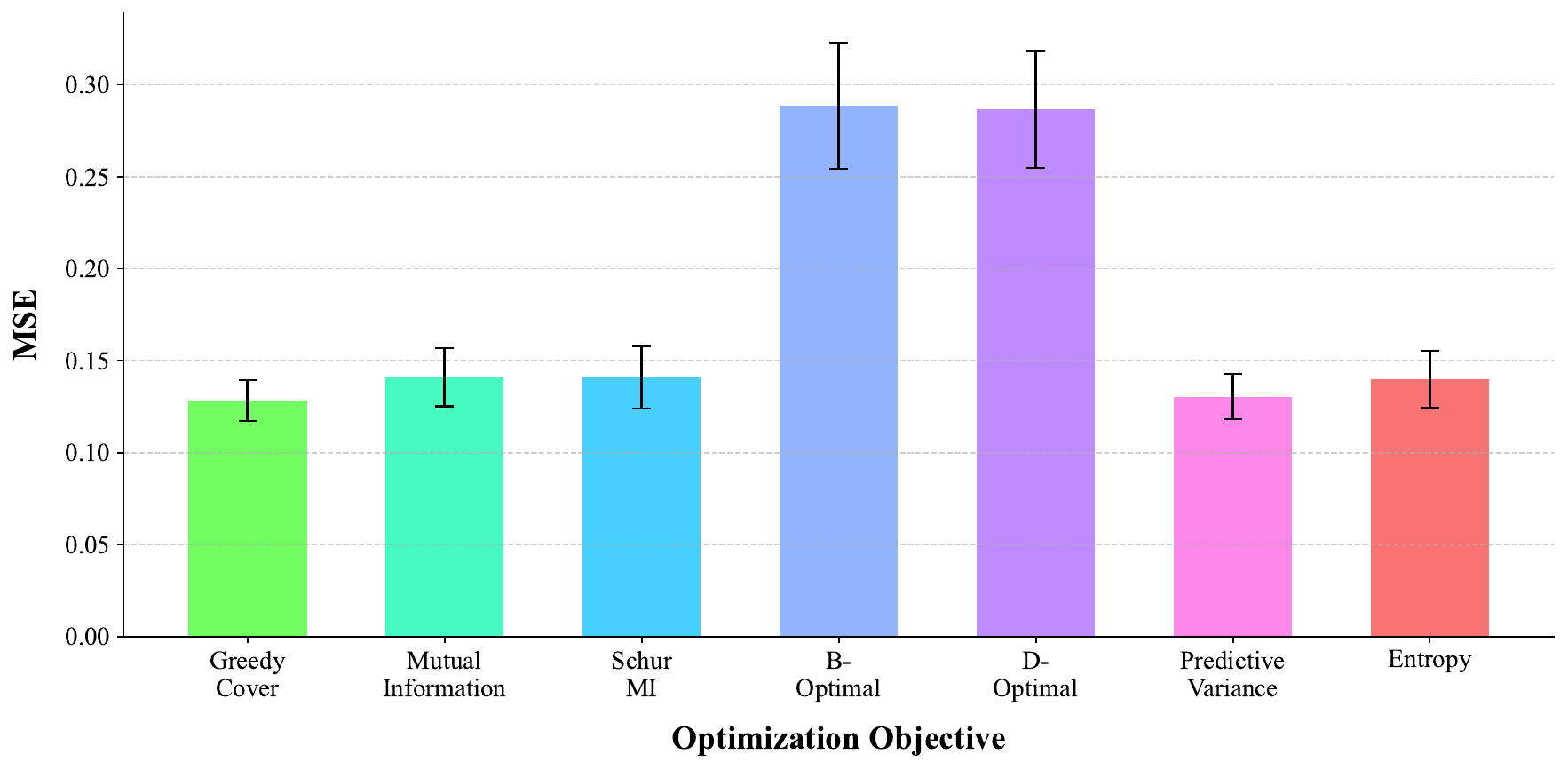}
    \end{subfigure}
    \hfill
    \begin{subfigure}{0.49\linewidth}
        \includegraphics[width=\linewidth]{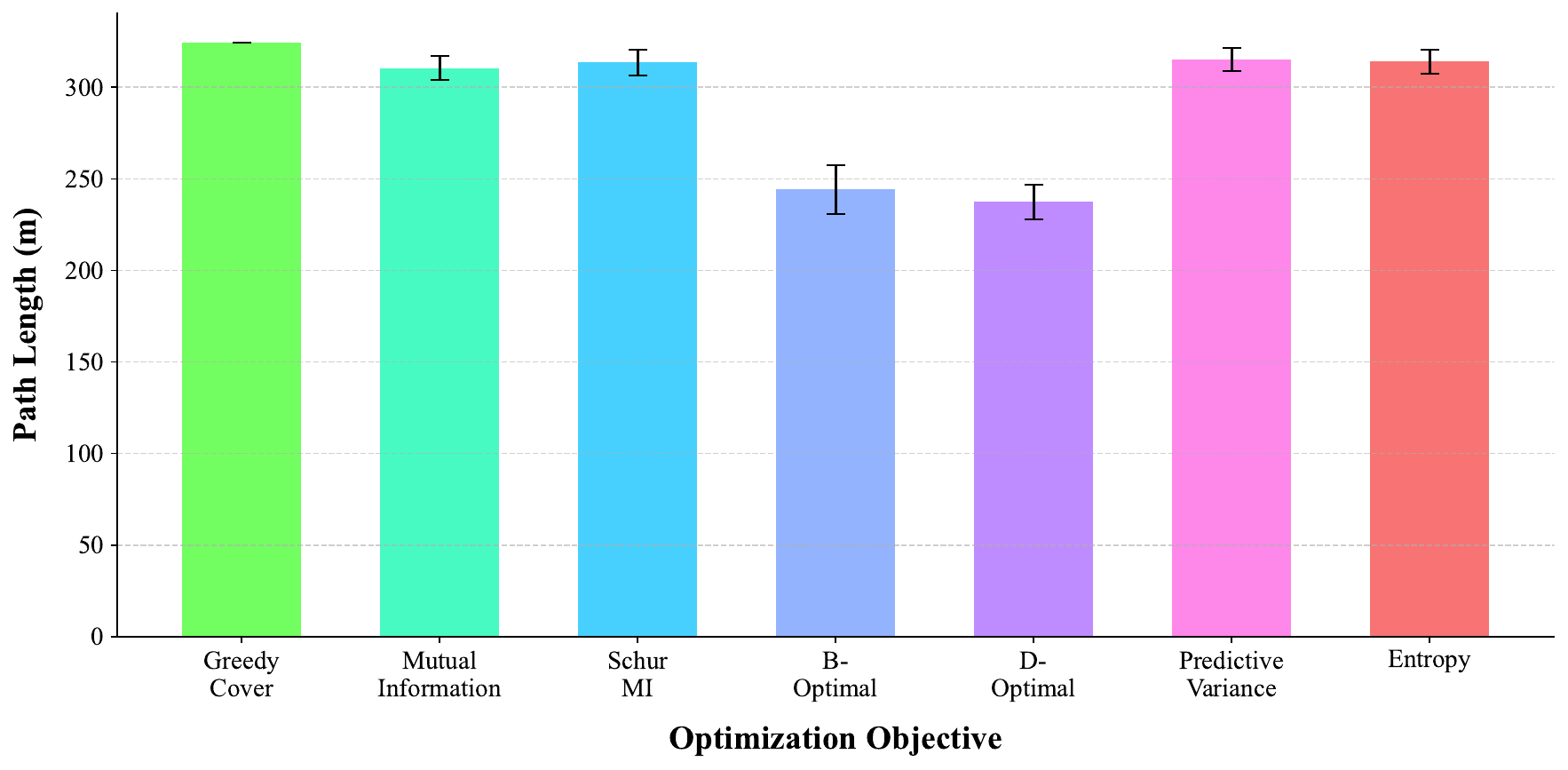}
    \end{subfigure}
    \caption{\((17^\circ\mathrm{N},\,073^\circ\mathrm{W})\) Estimation error (MSE and path length) across prominent information-theoretic objectives.}  
    \label{fig:objective_metrics_N17E073}
\end{figure*}

\begin{figure*}[!ht]
   \centering
    \begin{subfigure}{0.49\linewidth}
        \includegraphics[width=\linewidth]{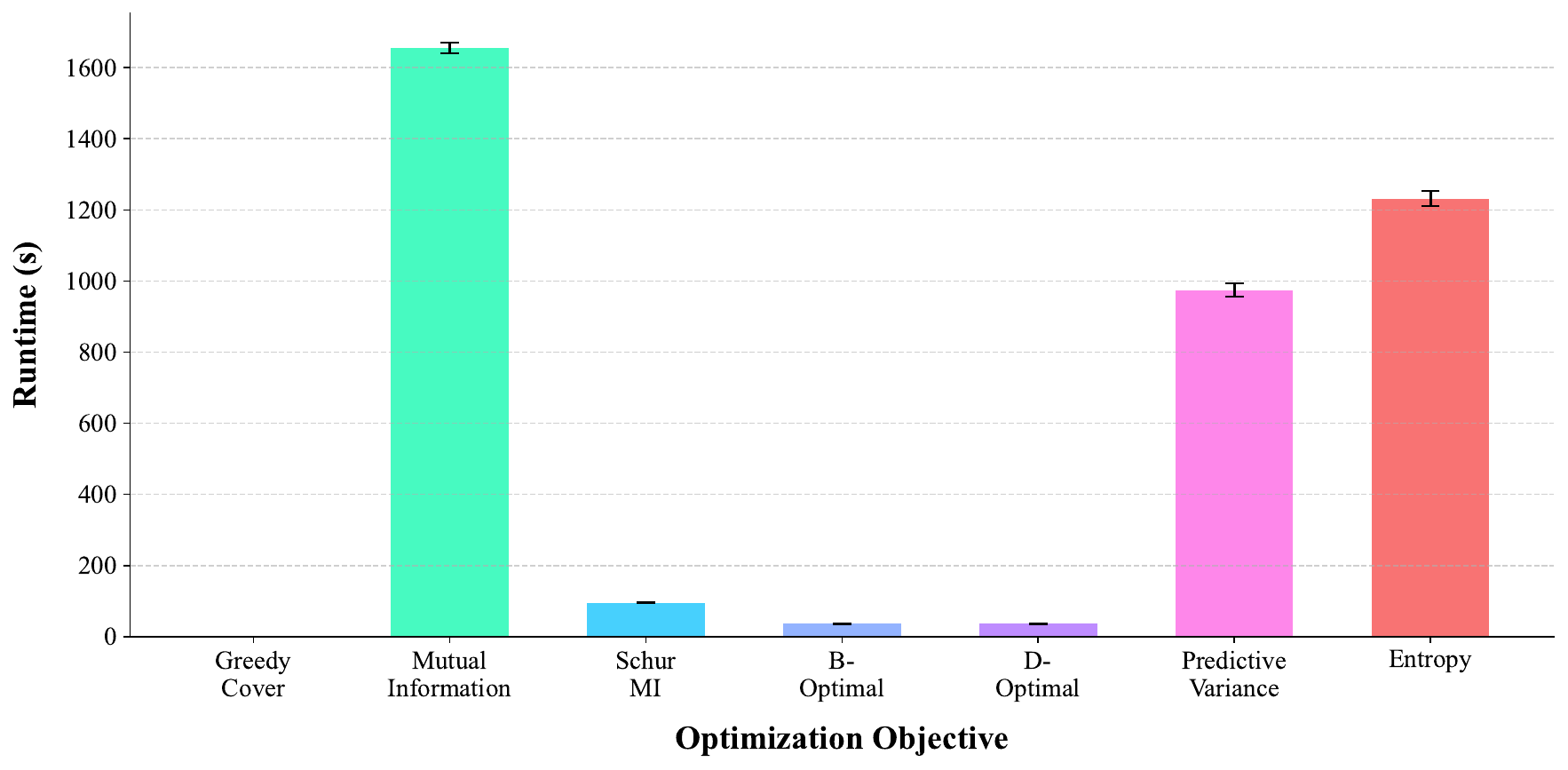}
    \end{subfigure}
    \hfill
    \begin{subfigure}{0.49\linewidth}
        \includegraphics[width=\linewidth]{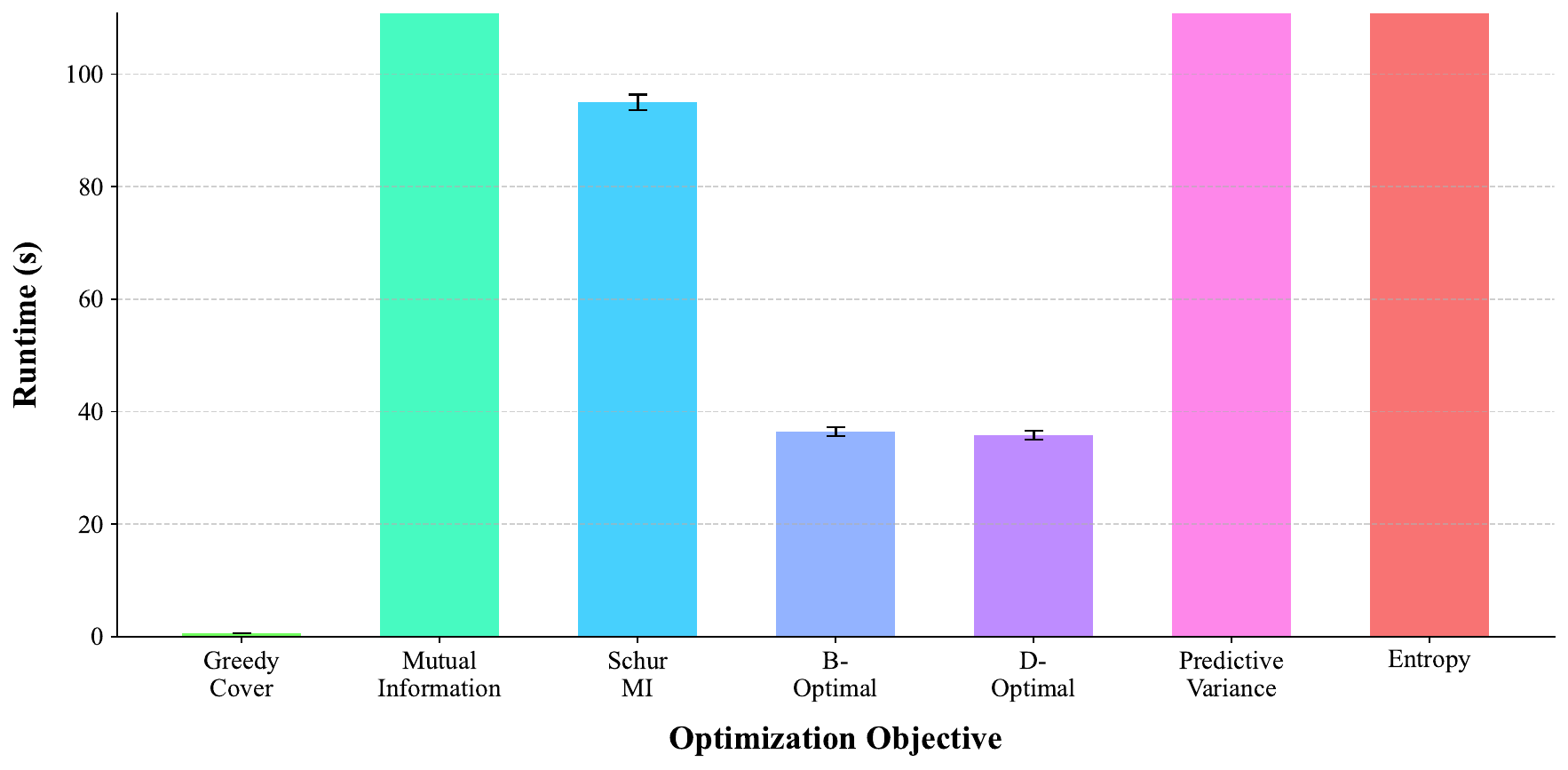}
    \end{subfigure}
    \caption{\((17^\circ\mathrm{N},\,073^\circ\mathrm{W})\) Runtime evaluation across information-theoretic objectives (right: zoomed version of the left).} 
    \label{fig:objective_runtime_N17E073}
\end{figure*}

\begin{figure*}[!ht]
   \centering
    \begin{subfigure}{0.49\linewidth}
        \includegraphics[width=\linewidth]{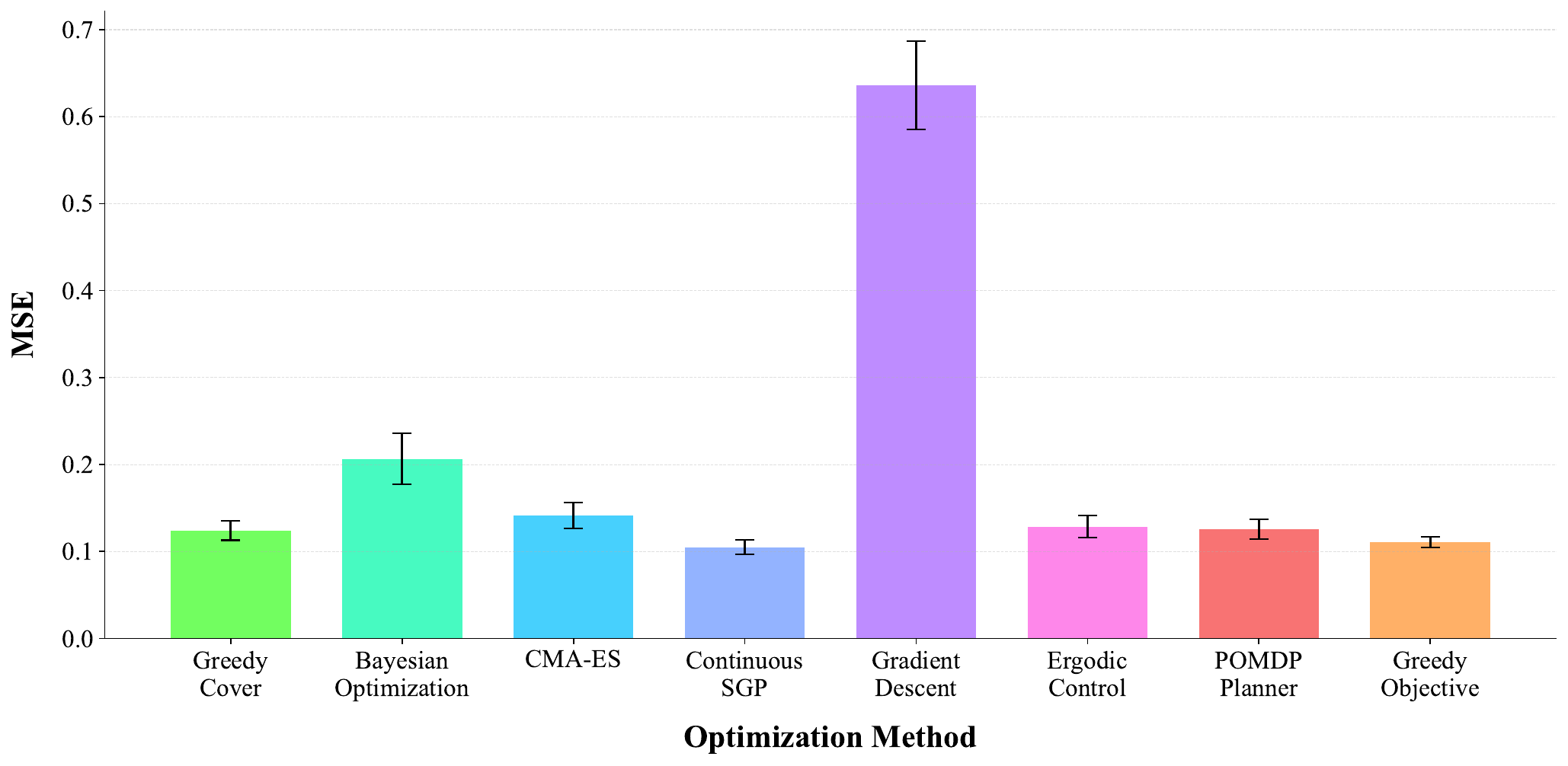}
    \end{subfigure}
    \hfill
    \begin{subfigure}{0.49\linewidth}
        \includegraphics[width=\linewidth]{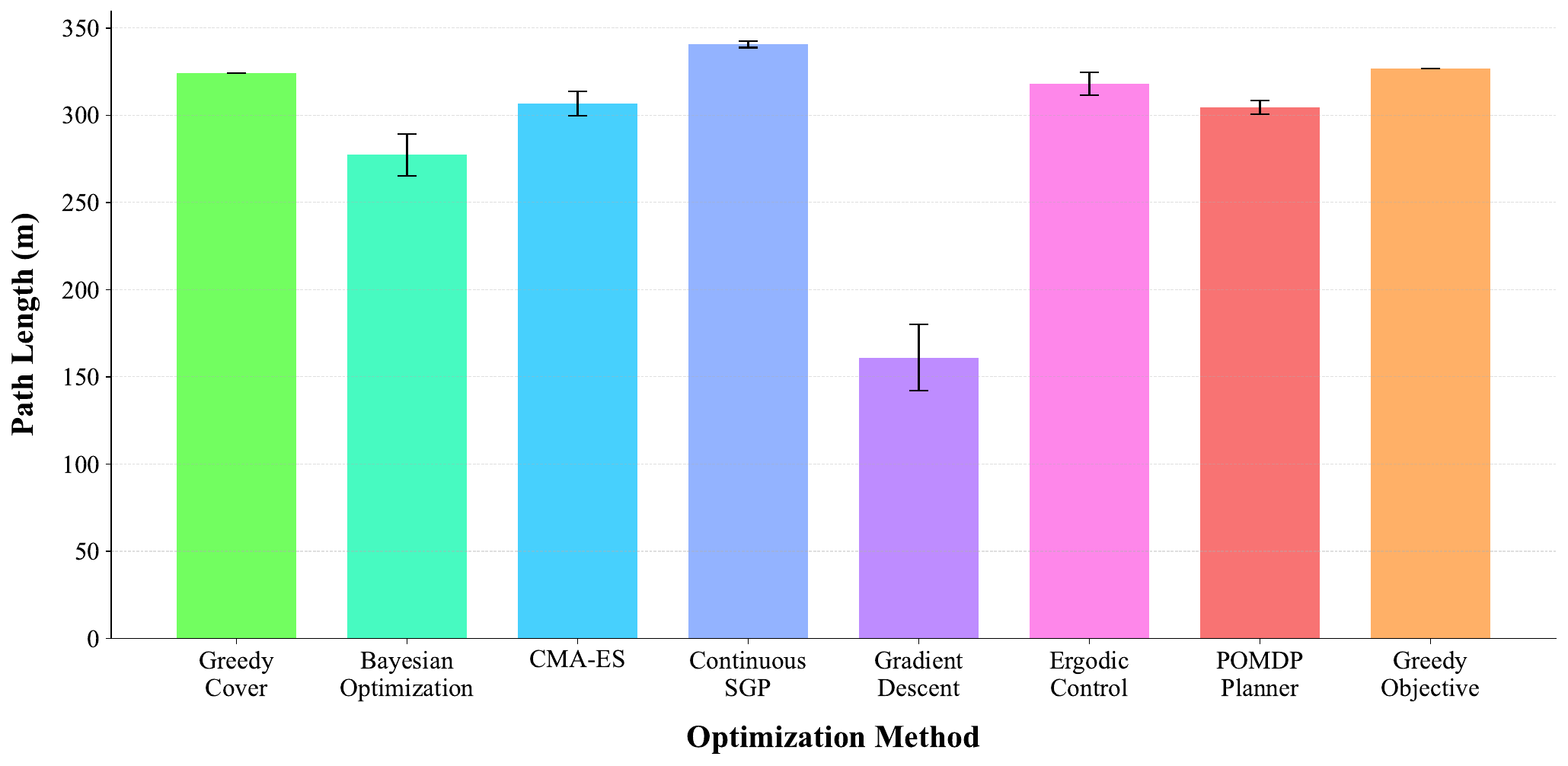}
    \end{subfigure}
    \caption{\((17^\circ\mathrm{N},\,073^\circ\mathrm{W})\) Estimation error (MSE and path length) across prominent optimization methods.}
    \label{fig:optimizers_metrics_N17E073}
\end{figure*}

\begin{figure*}[!ht]
   \centering
    \begin{subfigure}{0.49\linewidth}
        \includegraphics[width=\linewidth]{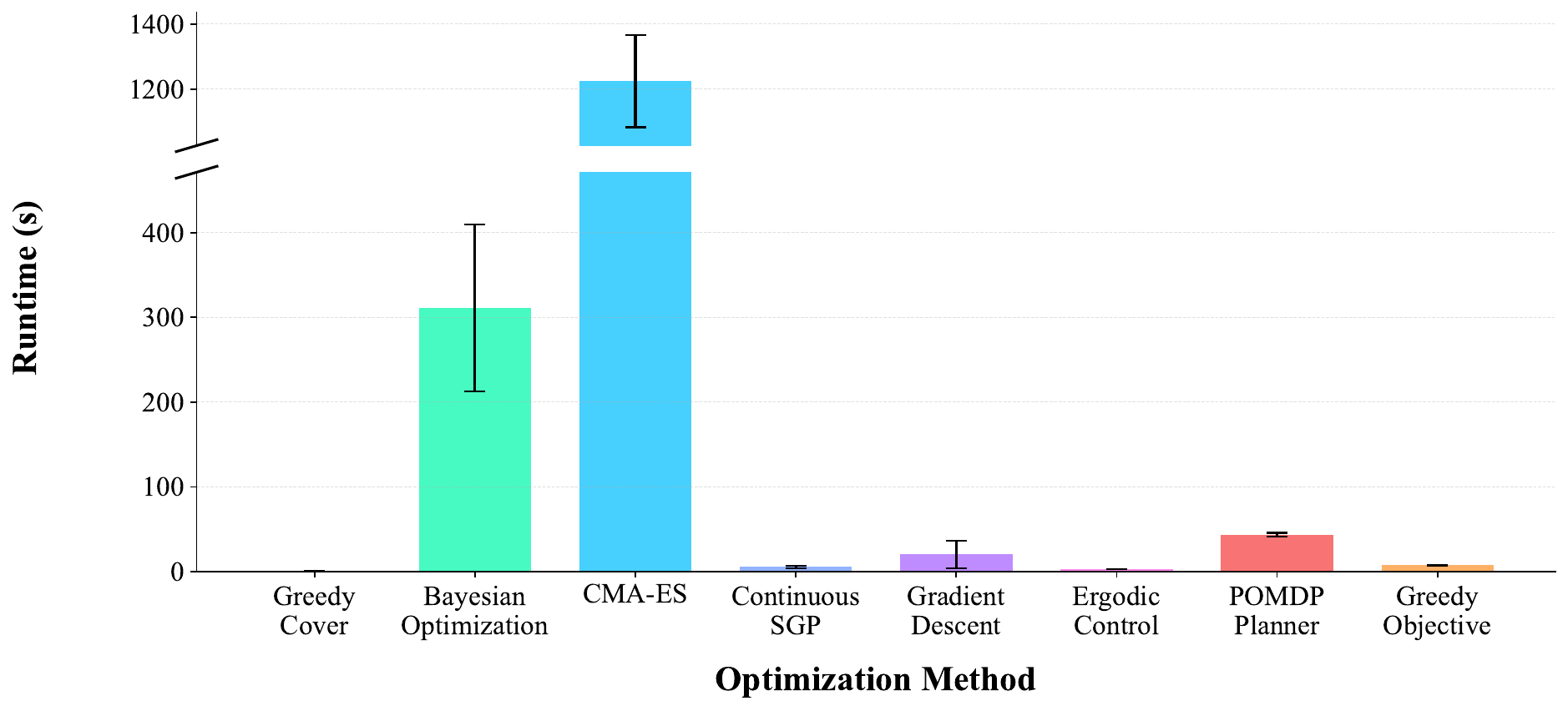}
    \end{subfigure}
    \hfill
    \begin{subfigure}{0.49\linewidth}
        \includegraphics[width=\linewidth]{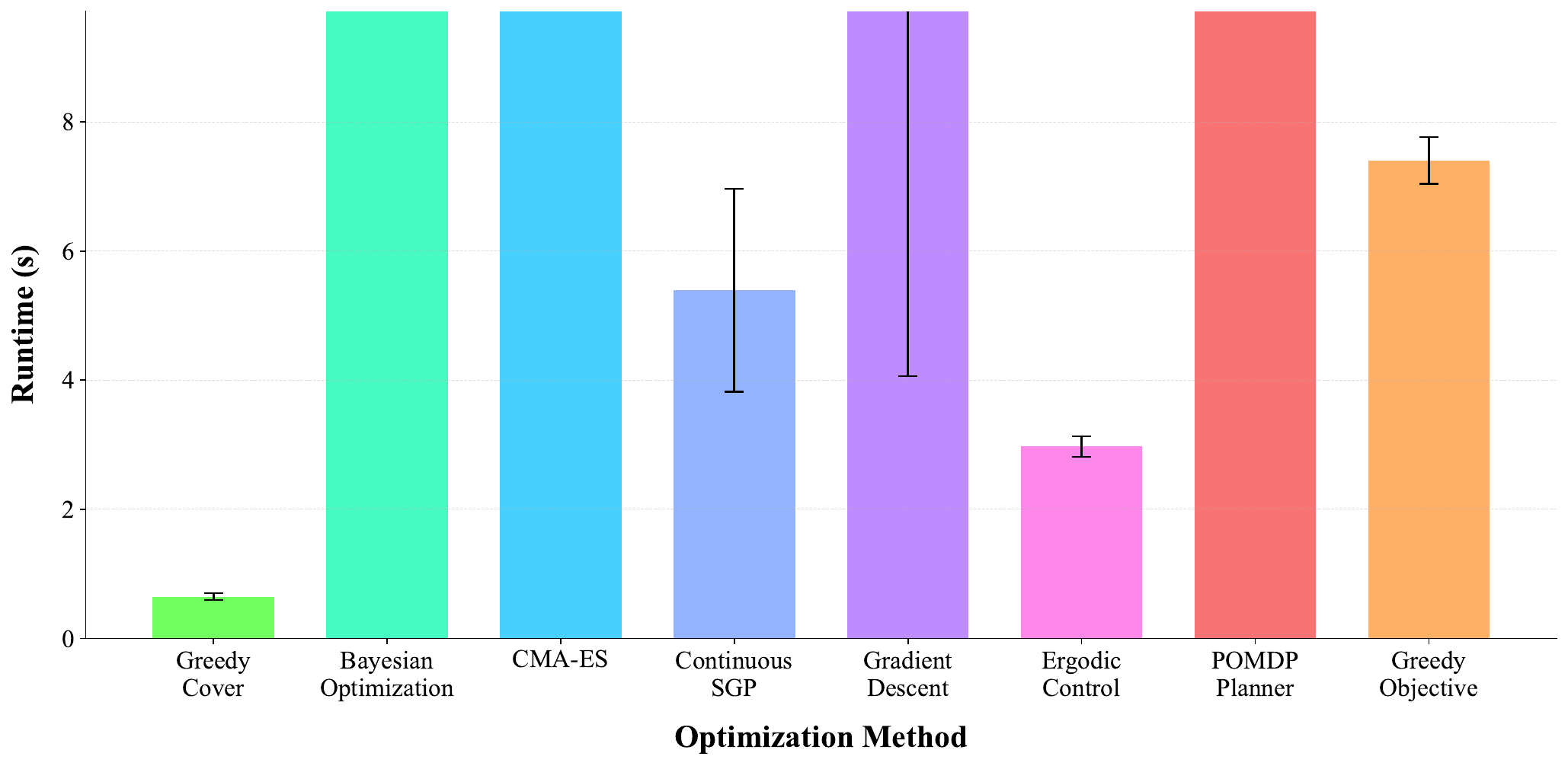}
    \end{subfigure}
    \caption{\((17^\circ\mathrm{N},\,073^\circ\mathrm{W})\) Runtime evaluation across prominent optimization methods (right: zoomed version of the left).}  
    \label{fig:optimizers_runtime_N17E073}
\end{figure*}


\begin{figure*}[!ht]
   \centering
    \begin{subfigure}{0.49\linewidth}
        \includegraphics[width=\linewidth]{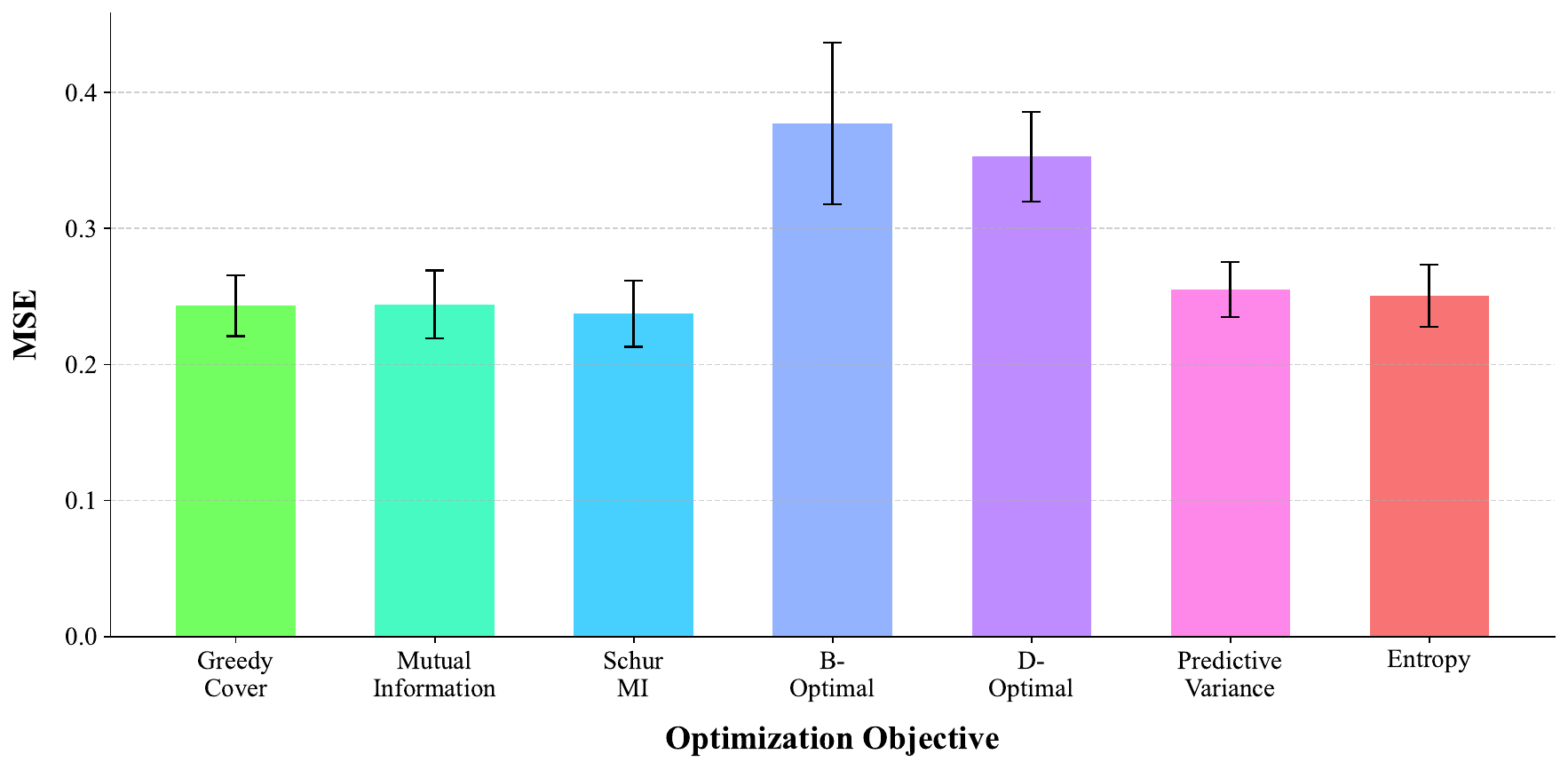}
    \end{subfigure}
    \hfill
    \begin{subfigure}{0.49\linewidth}
        \includegraphics[width=\linewidth]{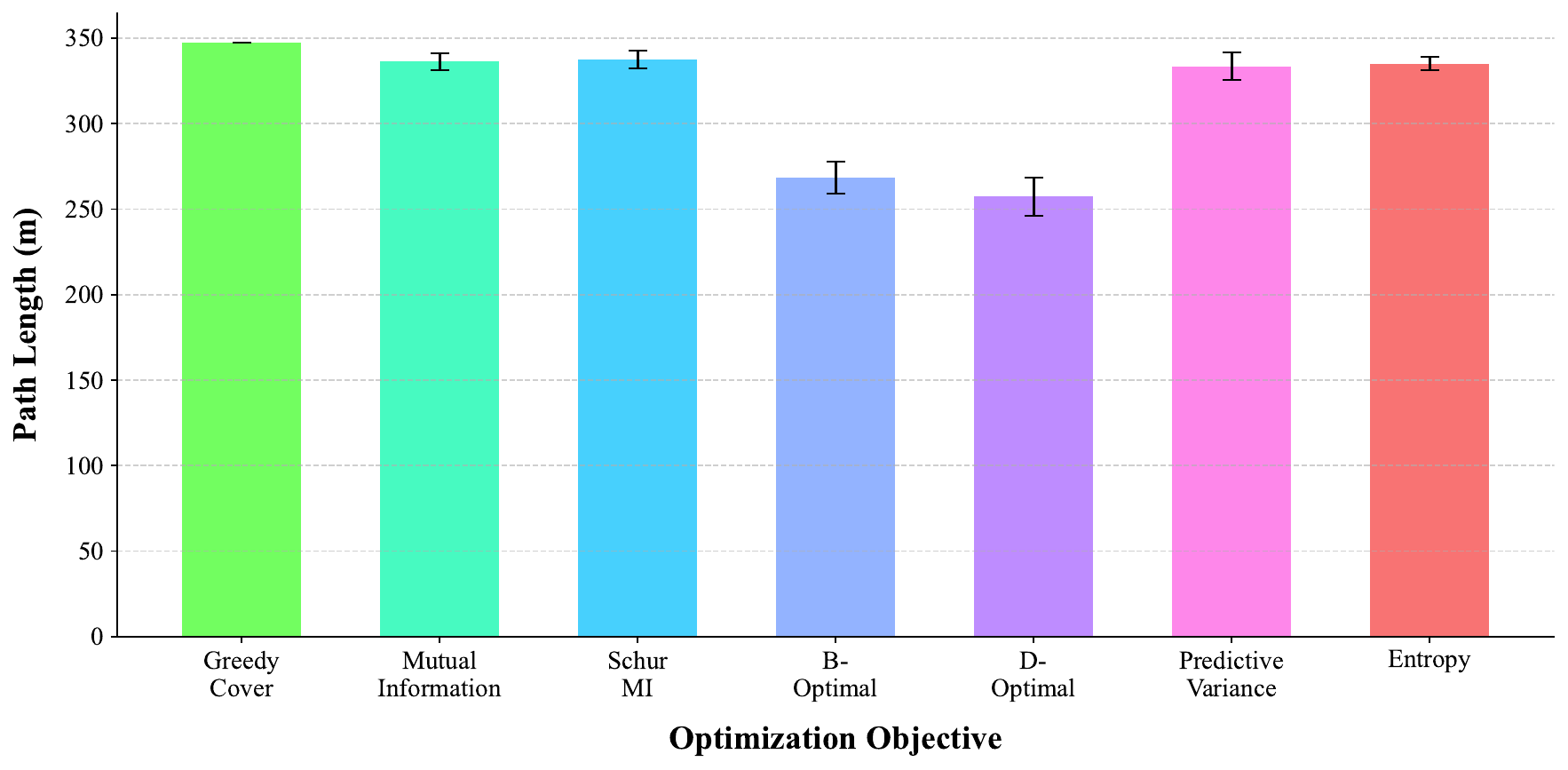}
    \end{subfigure}
    \caption{\((02^\circ\mathrm{N},\,021^\circ\mathrm{W})\) Estimation error (MSE and path length) across prominent information-theoretic objectives.}  
    \label{fig:objective_metrics_N02E021}
\end{figure*}

\begin{figure*}[!ht]
   \centering
    \begin{subfigure}{0.49\linewidth}
        \includegraphics[width=\linewidth]{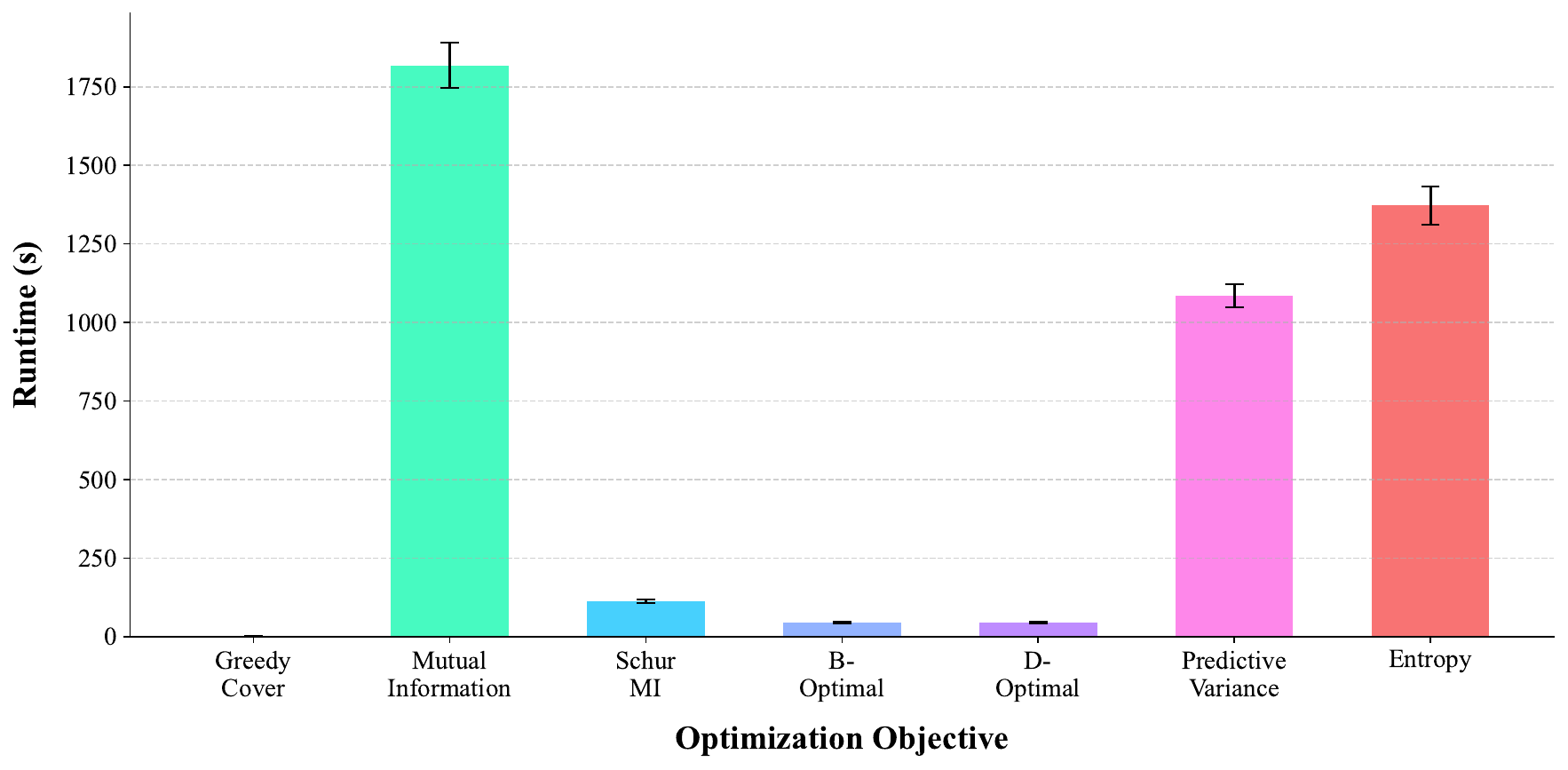}
    \end{subfigure}
    \hfill
    \begin{subfigure}{0.49\linewidth}
        \includegraphics[width=\linewidth]{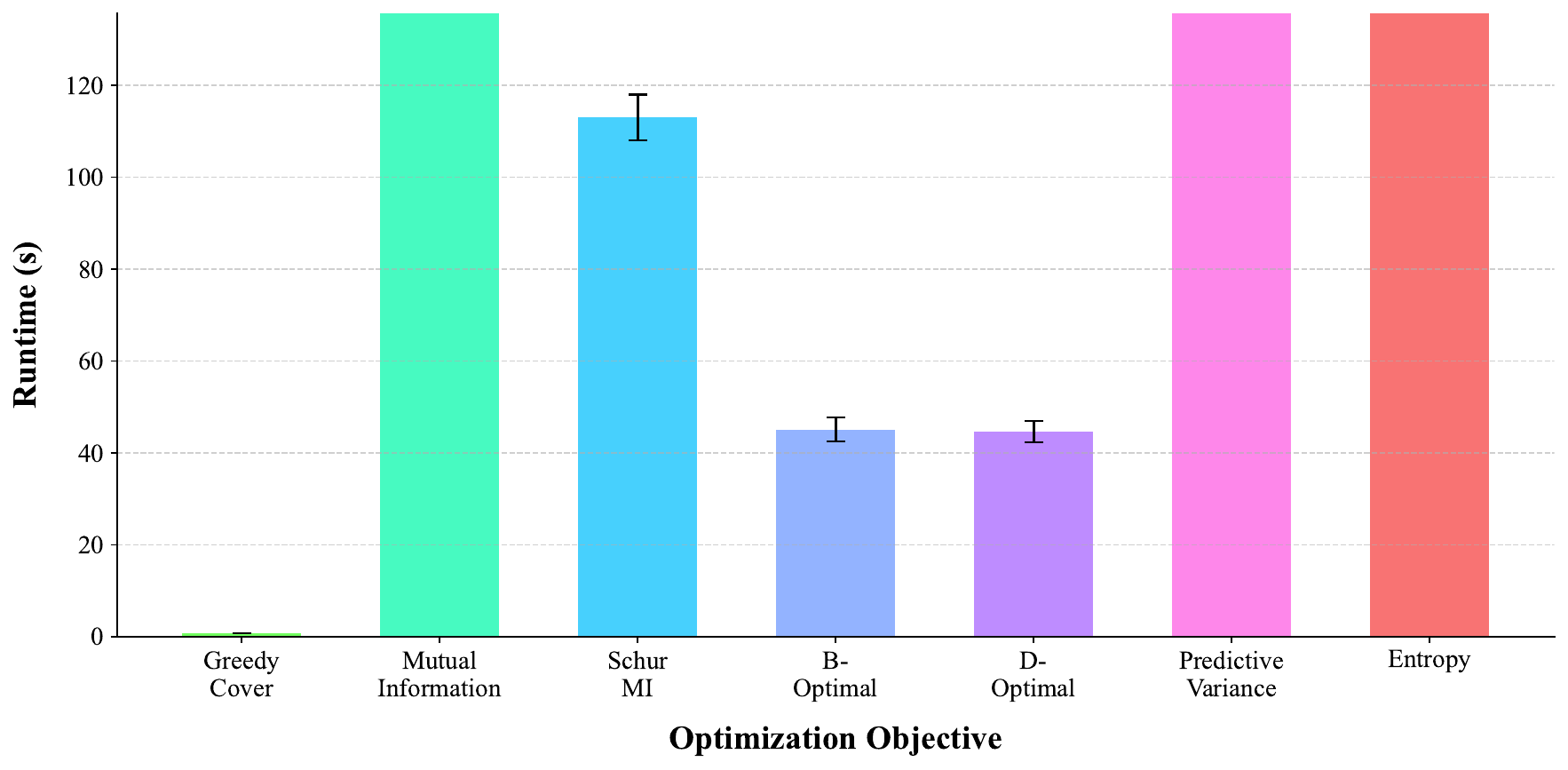}
    \end{subfigure}
    \caption{\((02^\circ\mathrm{N},\,021^\circ\mathrm{W})\) Runtime evaluation across information-theoretic objectives (right: zoomed version of the left).} 
    \label{fig:objective_runtime_N02E021}
\end{figure*}

\begin{figure*}[!ht]
   \centering
    \begin{subfigure}{0.49\linewidth}
        \includegraphics[width=\linewidth]{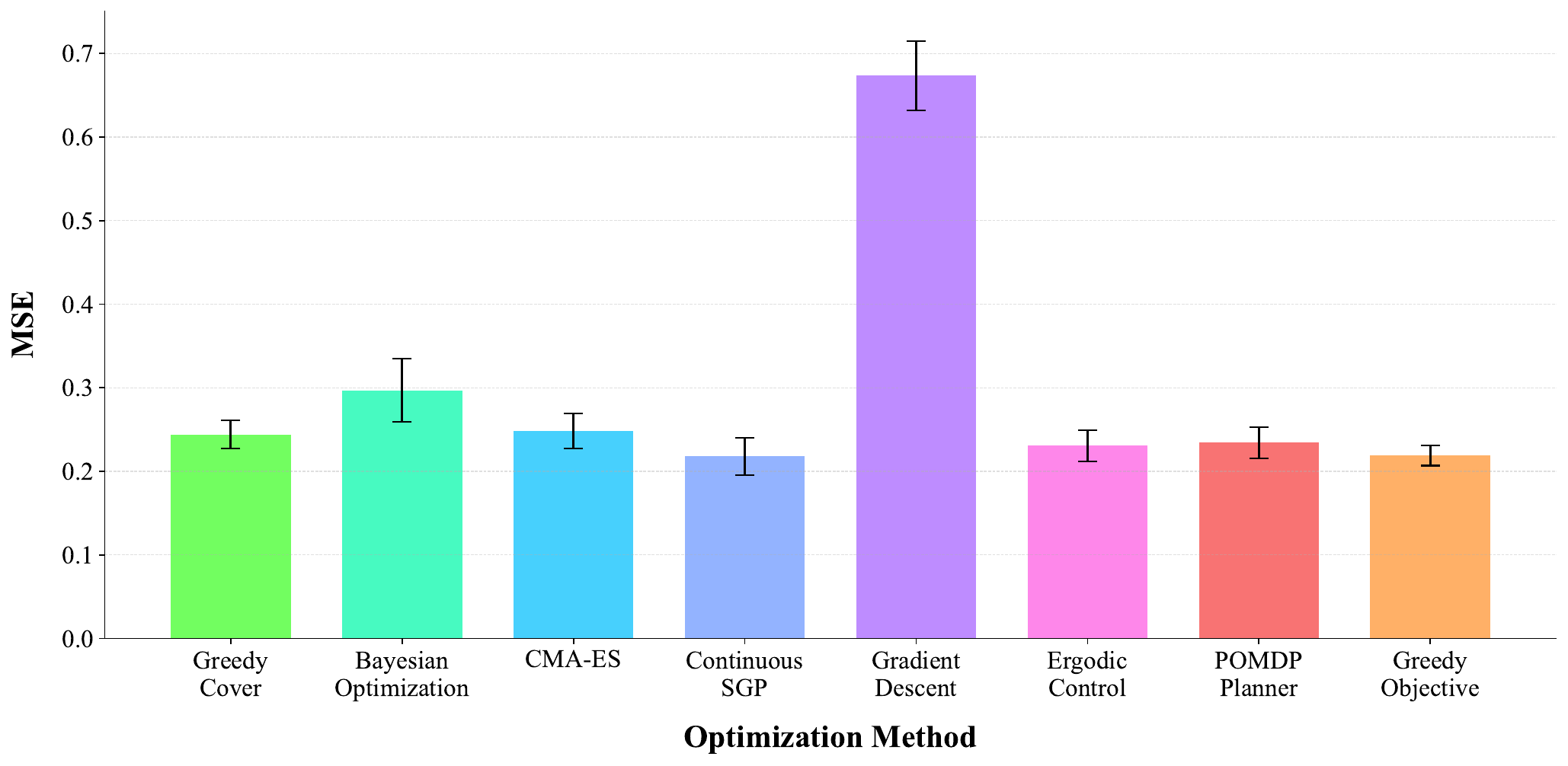}
    \end{subfigure}
    \hfill
    \begin{subfigure}{0.49\linewidth}
        \includegraphics[width=\linewidth]{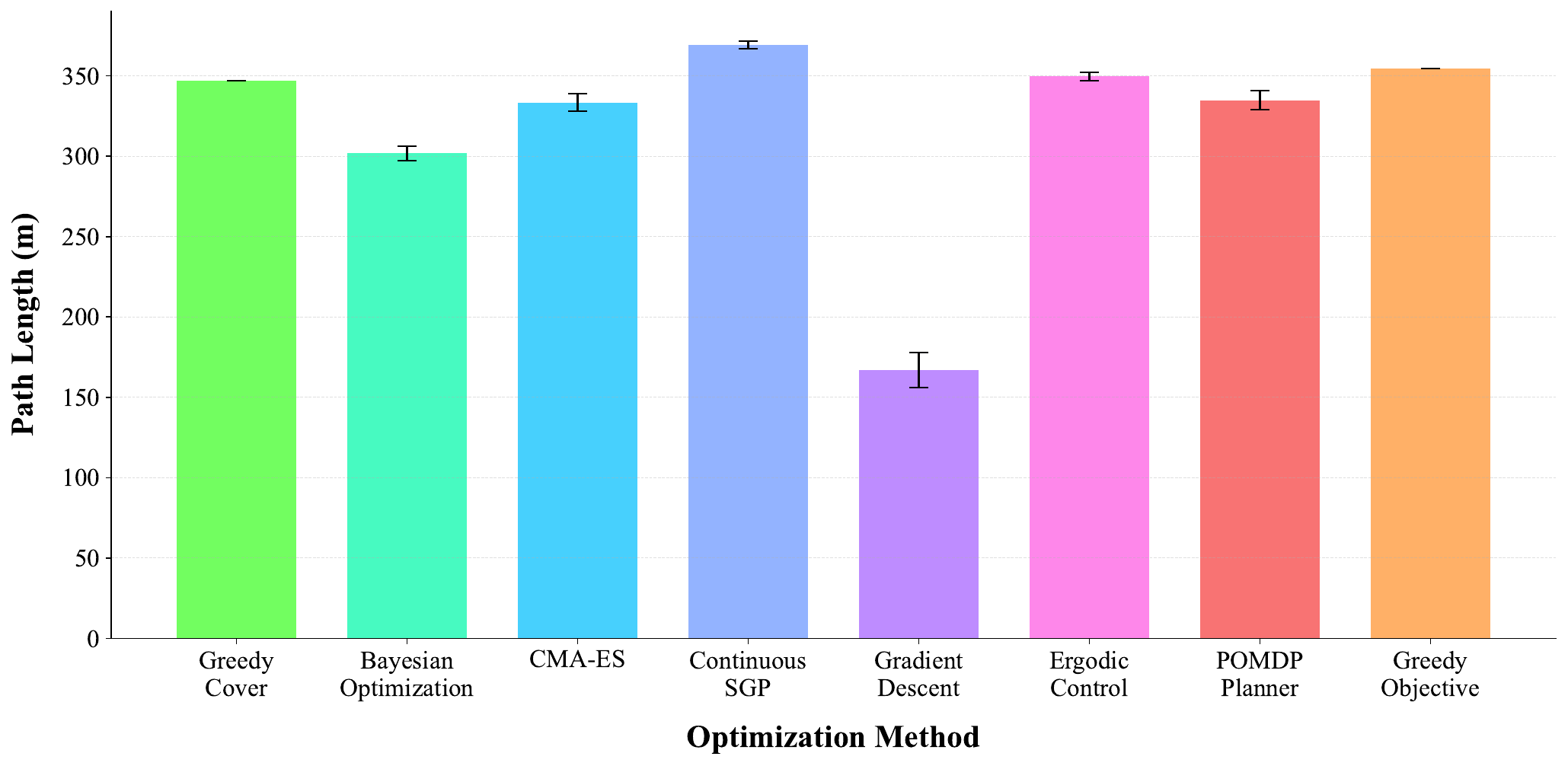}
    \end{subfigure}
    \caption{\((02^\circ\mathrm{N},\,021^\circ\mathrm{W})\) Estimation error (MSE and path length) across prominent optimization methods.}
    \label{fig:optimizers_metrics_N02E021}
\end{figure*}

\begin{figure*}[!ht]
   \centering
    \begin{subfigure}{0.49\linewidth}
        \includegraphics[width=\linewidth]{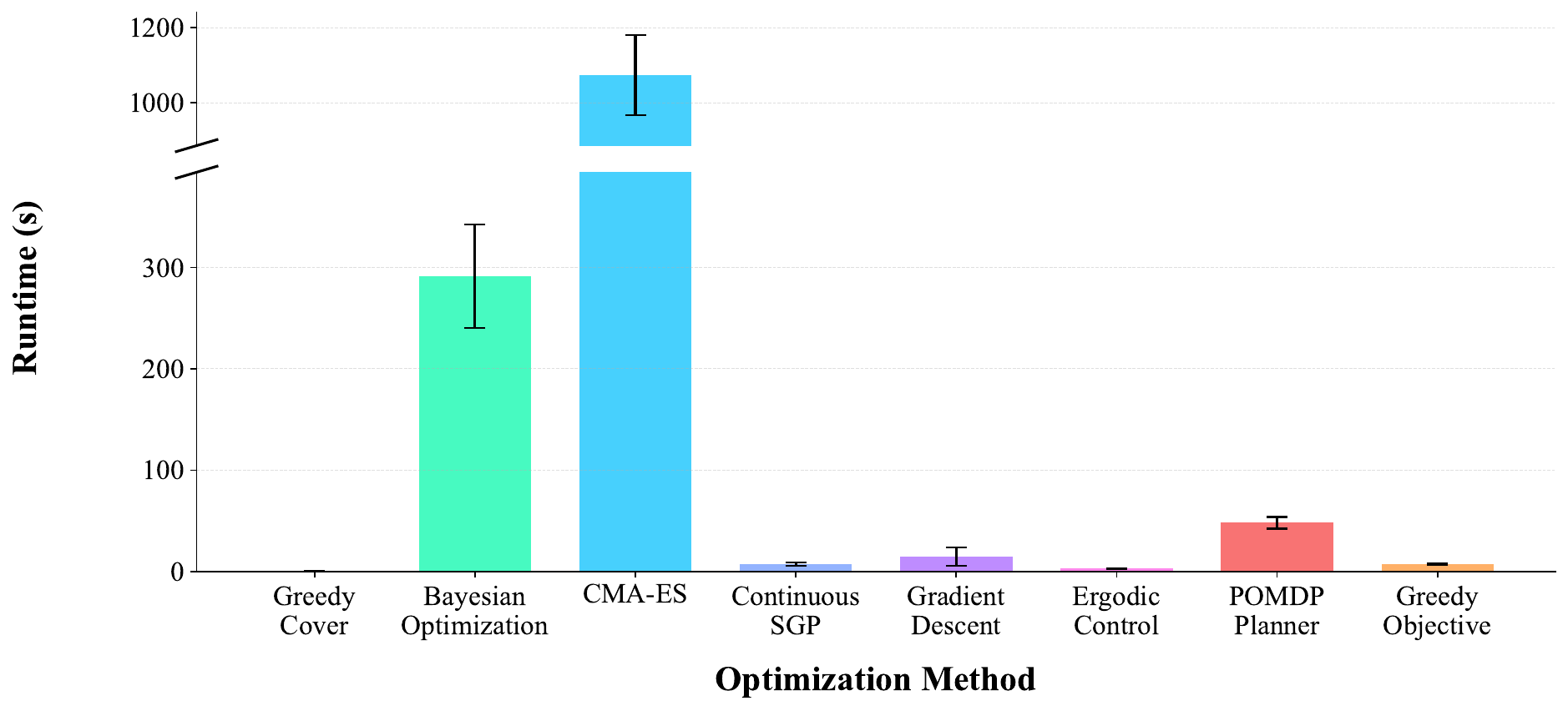}
    \end{subfigure}
    \hfill
    \begin{subfigure}{0.49\linewidth}
        \includegraphics[width=\linewidth]{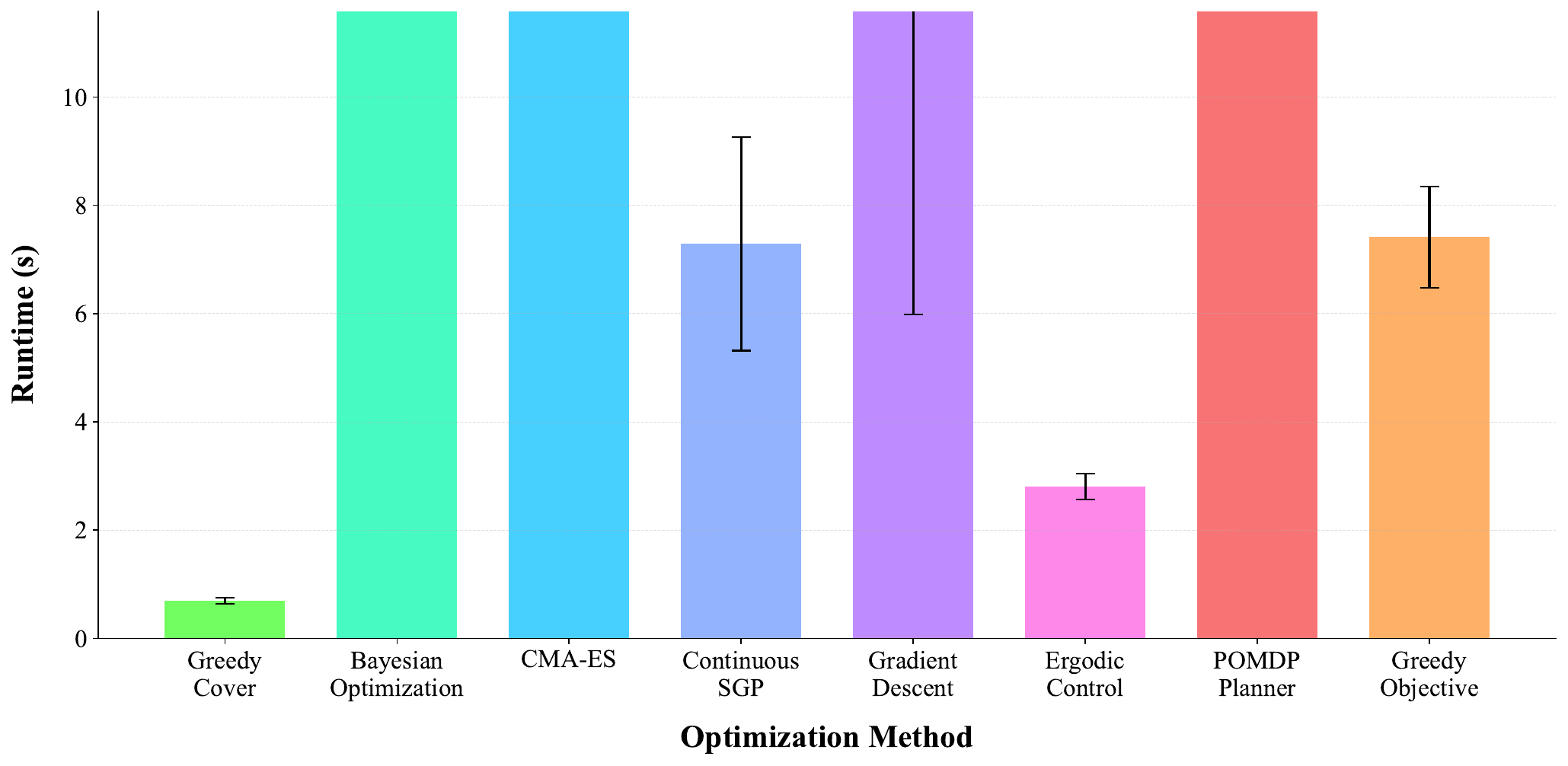}
    \end{subfigure}
    \caption{\((02^\circ\mathrm{N},\,021^\circ\mathrm{W})\) Runtime evaluation across prominent optimization methods (right: zoomed version of the left).}  
    \label{fig:optimizers_runtime_N02E021}
\end{figure*}


\begin{figure*}[!ht]
   \centering
    \begin{subfigure}{0.49\linewidth}
        \includegraphics[width=\linewidth]{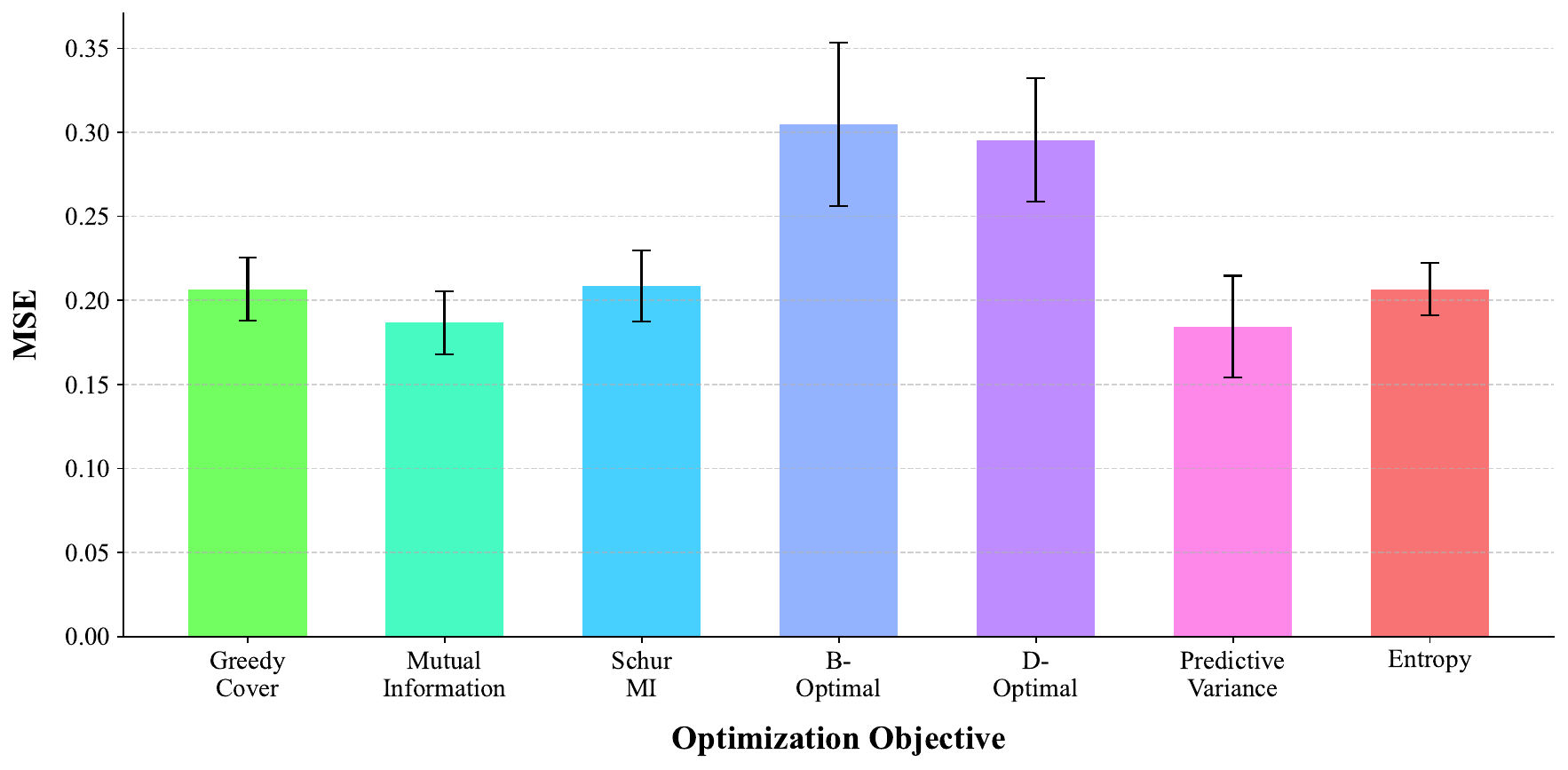}
    \end{subfigure}
    \hfill
    \begin{subfigure}{0.49\linewidth}
        \includegraphics[width=\linewidth]{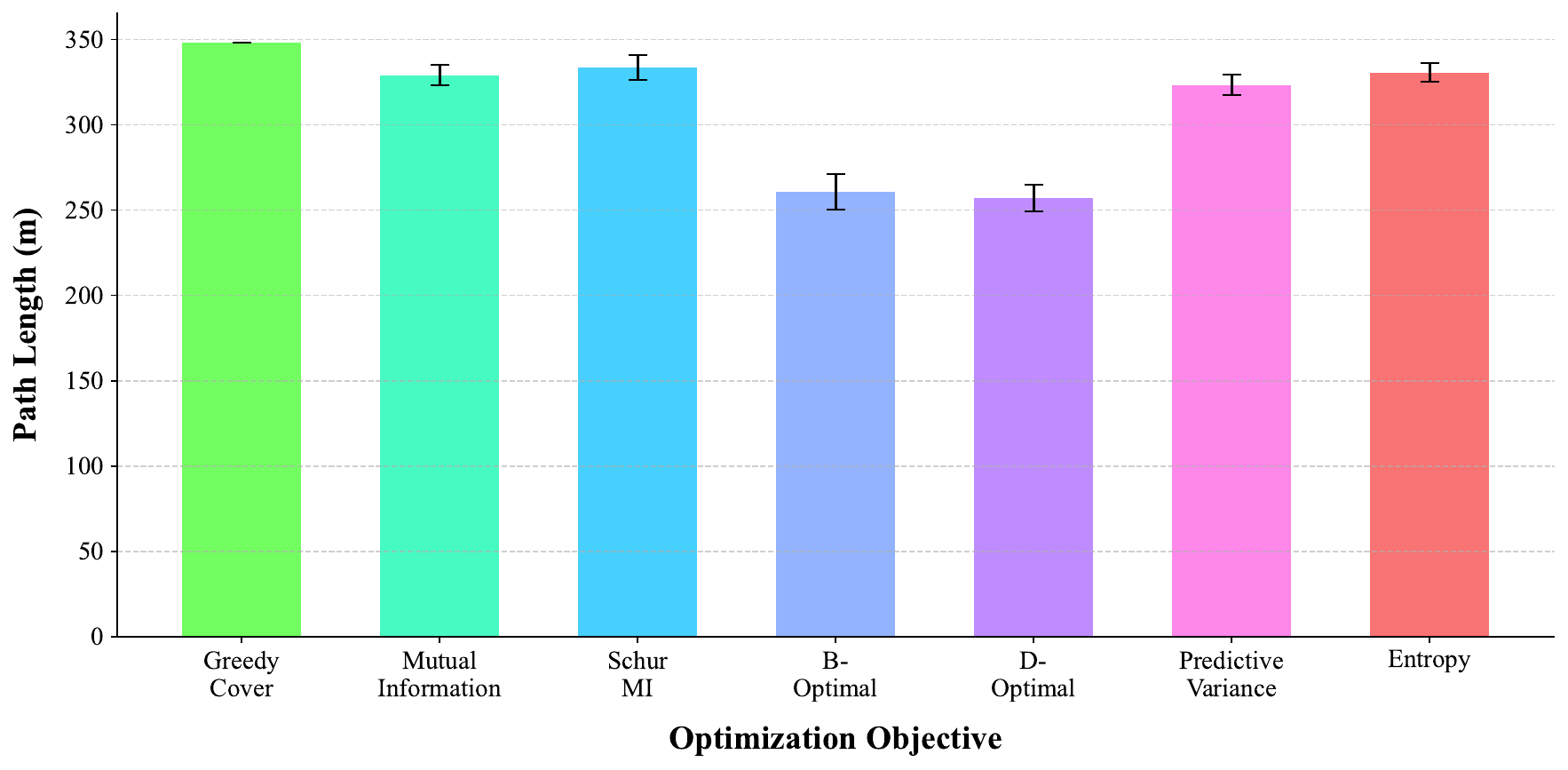}
    \end{subfigure}
    \caption{\((47^\circ\mathrm{N},\,124^\circ\mathrm{W})\) Estimation error (MSE and path length) across prominent information-theoretic objectives.}  
    \label{fig:objective_metrics_N47W124}
\end{figure*}

\begin{figure*}[!ht]
   \centering
    \begin{subfigure}{0.49\linewidth}
        \includegraphics[width=\linewidth]{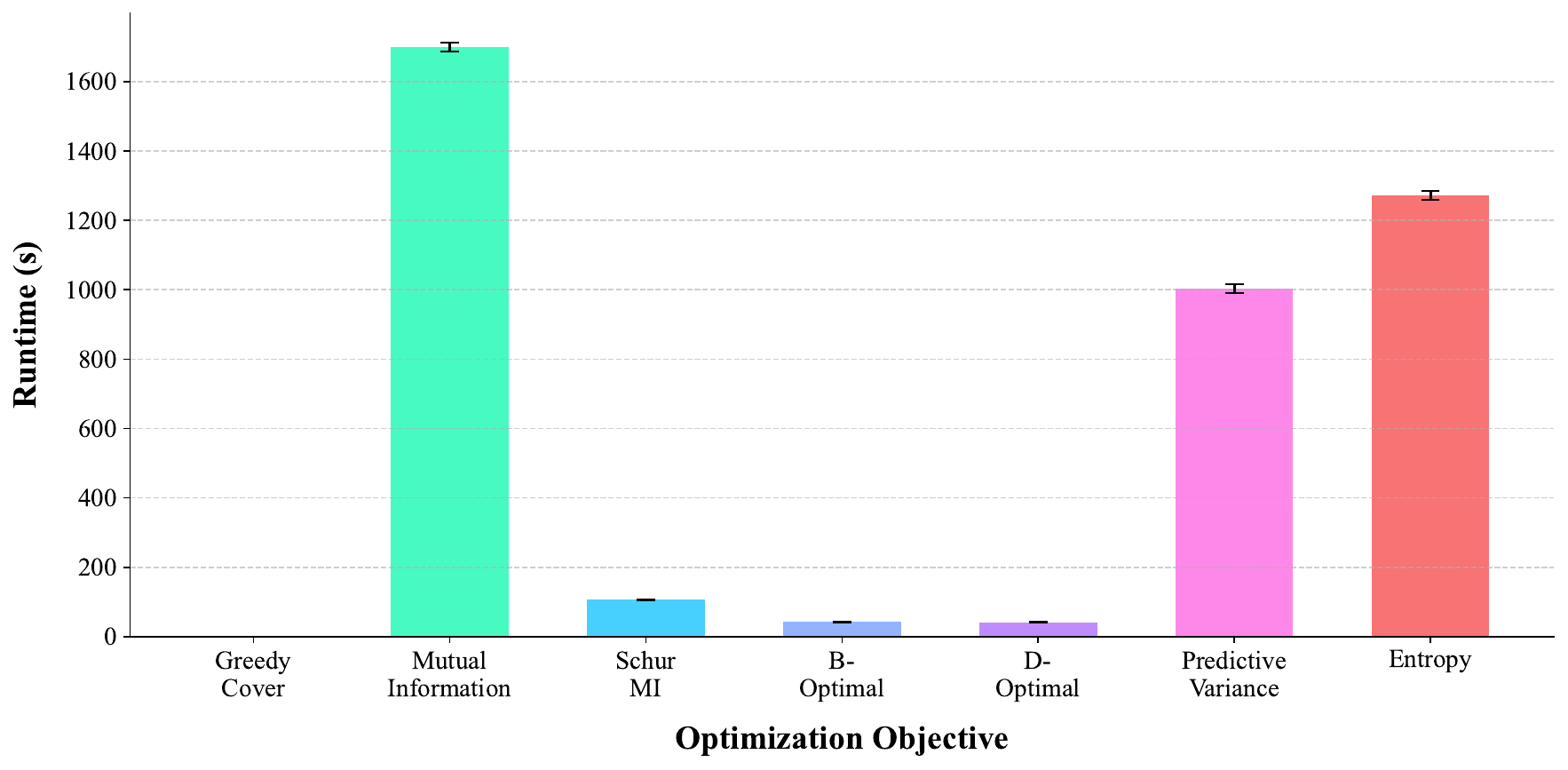}
    \end{subfigure}
    \hfill
    \begin{subfigure}{0.49\linewidth}
        \includegraphics[width=\linewidth]{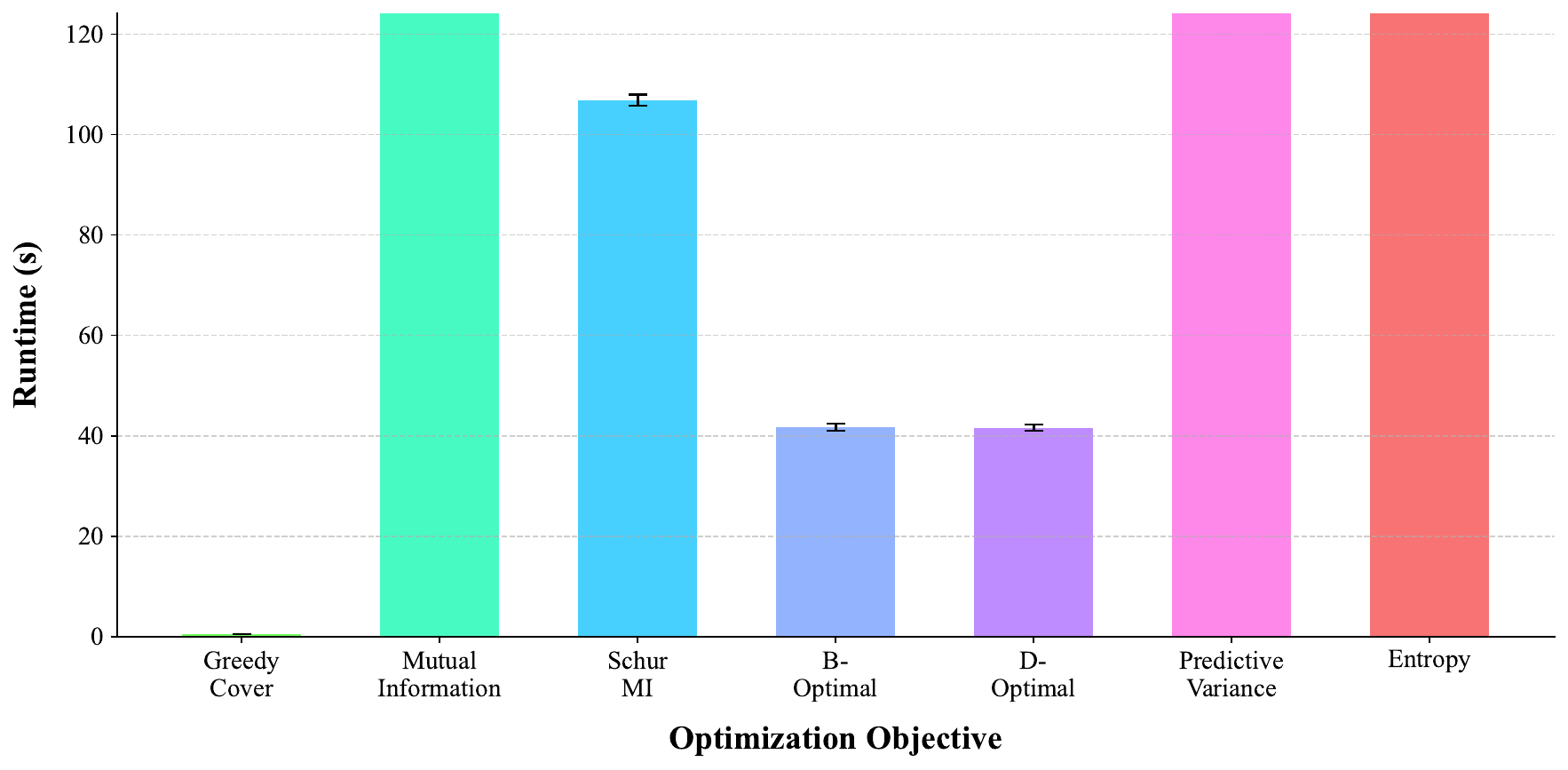}
    \end{subfigure}
    \caption{\((47^\circ\mathrm{N},\,124^\circ\mathrm{W})\) Runtime evaluation across information-theoretic objectives (right: zoomed version of the left).} 
    \label{fig:objective_runtime_N47W124}
\end{figure*}

\begin{figure*}[!ht]
   \centering
    \begin{subfigure}{0.49\linewidth}
        \includegraphics[width=\linewidth]{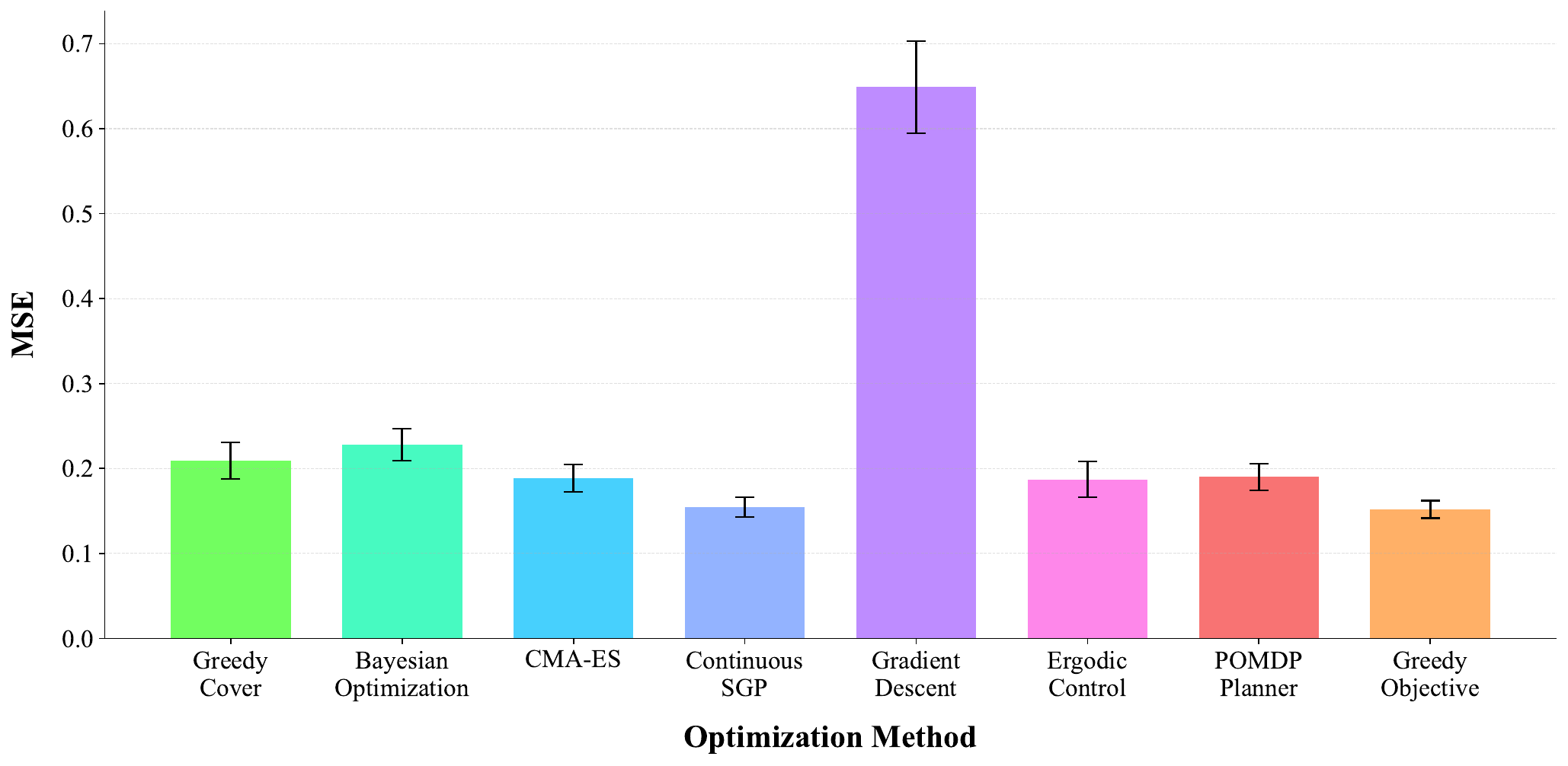}
    \end{subfigure}
    \hfill
    \begin{subfigure}{0.49\linewidth}
        \includegraphics[width=\linewidth]{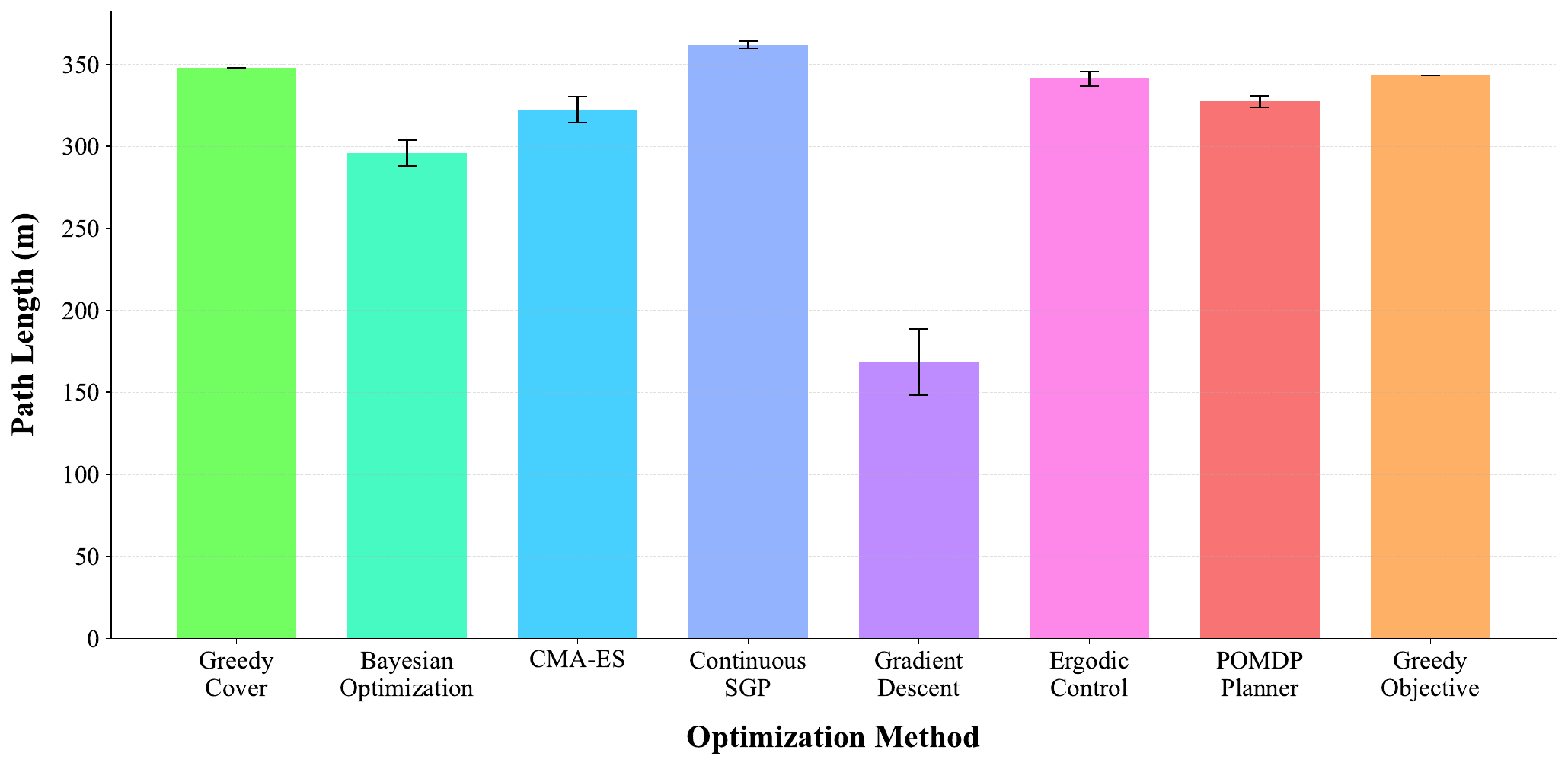}
    \end{subfigure}
    \caption{\((47^\circ\mathrm{N},\,124^\circ\mathrm{W})\) Estimation error (MSE and path length) across prominent optimization methods.}
    \label{fig:optimizers_metrics_N47W124}
\end{figure*}

\begin{figure*}[!ht]
   \centering
    \begin{subfigure}{0.49\linewidth}
        \includegraphics[width=\linewidth]{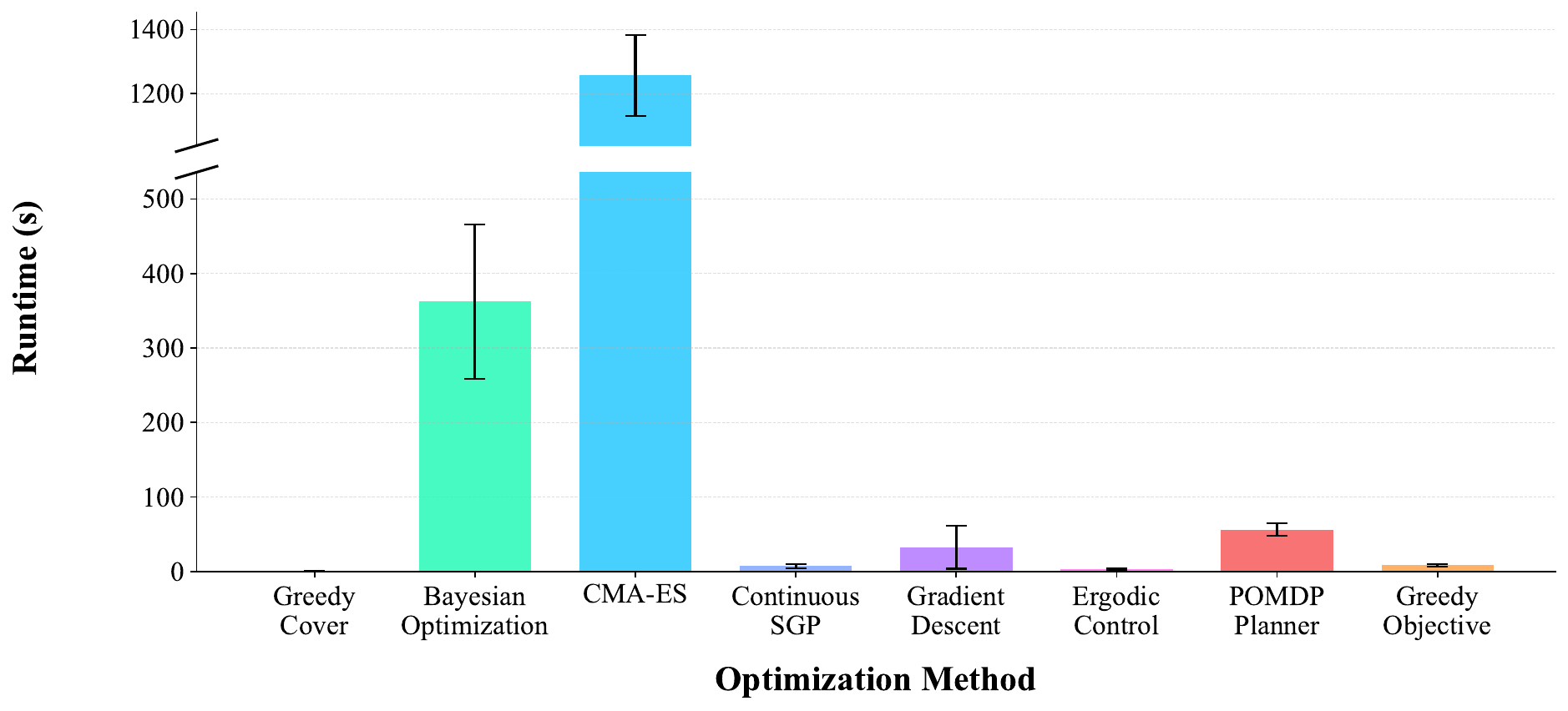}
    \end{subfigure}
    \hfill
    \begin{subfigure}{0.49\linewidth}
        \includegraphics[width=\linewidth]{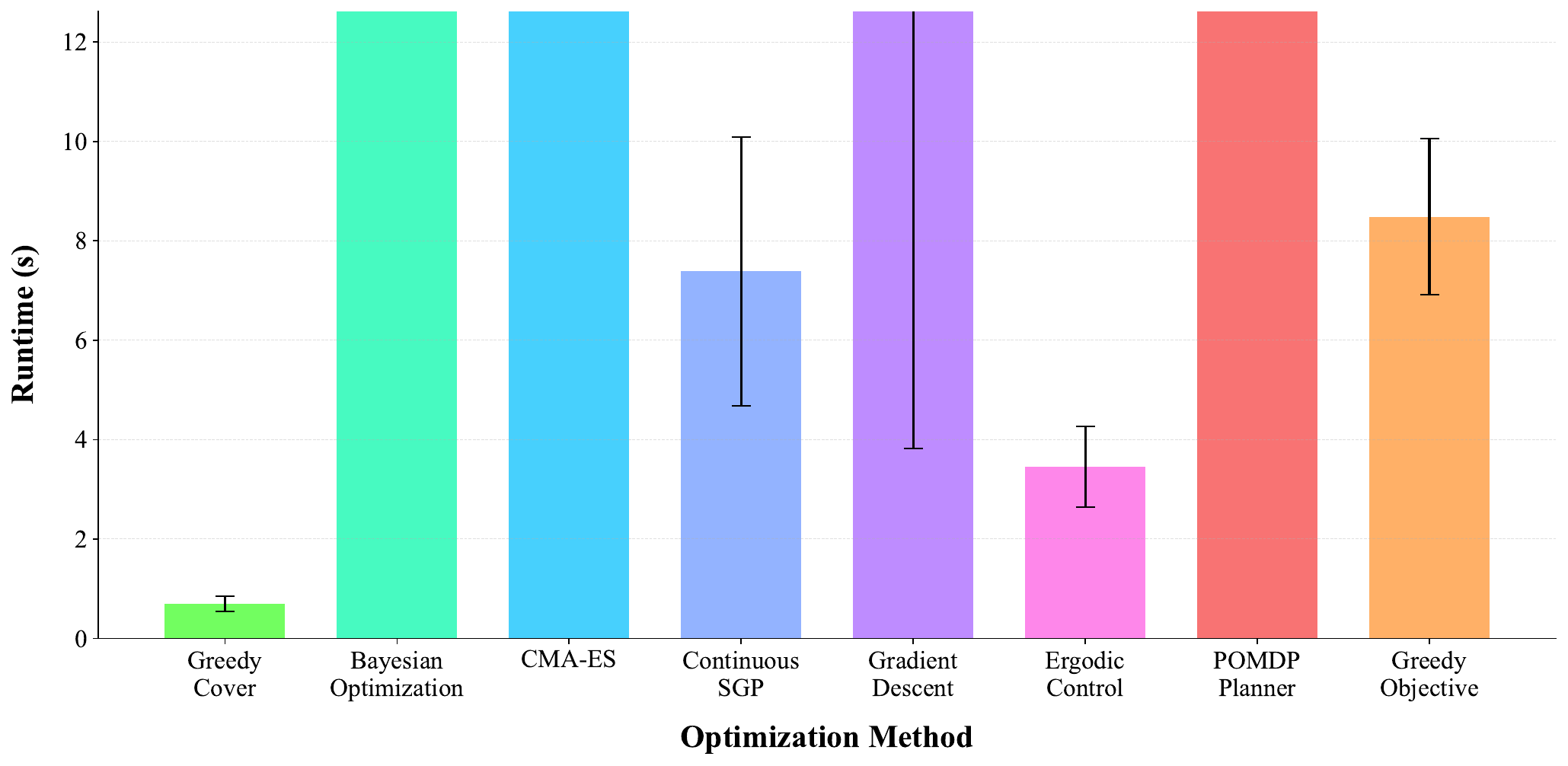}
    \end{subfigure}
    \caption{\((47^\circ\mathrm{N},\,124^\circ\mathrm{W})\) Runtime evaluation across prominent optimization methods (right: zoomed version of the left).}  
    \label{fig:optimizers_runtime_N47W124}
\end{figure*}

\end{document}